%% file: main_neurips.tex
\documentclass{article}

\PassOptionsToPackage{round}{natbib}
     \usepackage[final]{neurips_2019}

\usepackage[utf8]{inputenc} % allow utf-8 input
\usepackage[T1]{fontenc}    % use 8-bit T1 fonts
\usepackage{hyperref}       % hyperlinks
\usepackage{url}            % simple URL typesetting
\usepackage{booktabs}       % professional-quality tables
\usepackage{amsfonts}       % blackboard math symbols
\usepackage{nicefrac}       % compact symbols for 1/2, etc.
\usepackage{microtype}      % microtypography

\usepackage{microtype}
\usepackage{graphicx}
\usepackage{diagbox}  % for diagonal line in table
\usepackage{subfigure}
\usepackage{booktabs} % for professional tables
\usepackage[utf8]{inputenc} % allow utf-8 input
\usepackage[T1]{fontenc}    % use 8-bit T1 fonts
\usepackage{url}            % simple URL typesetting
\usepackage{amsfonts}       % blackboard math symbols
\usepackage{nicefrac}       % compact symbols for 1/2, etc.
\usepackage{color}
\usepackage{amsmath}
\usepackage{xcolor}
\usepackage{paralist}
\usepackage{enumerate}
\usepackage{siunitx}
\usepackage{listings}
\usepackage{courier}
\usepackage{algorithm}
\usepackage{algorithmic}
\usepackage{enumitem}
\usepackage{float}
\usepackage{rotating}  % for hypers table
\usepackage{tablefootnote}
\usepackage{footmisc}
\usepackage{refcount}
\usepackage{upgreek}
\usepackage{wrapfig}
\setlist[itemize]{noitemsep, topsep=0pt}

\hypersetup{
    colorlinks=true,
    linkcolor=black,
    filecolor=black,      
    urlcolor=blue,
    citecolor = black,
}

\title{Generalization of Reinforcement Learners with Working and Episodic Memory}

\author{
Meire Fortunato$^\star$ \qquad
Melissa Tan$^\star$\\
\textbf{
Ryan Faulkner$^\star$ \qquad
Steven Hansen$^\star$ \qquad 
Adri\`a Puigdom\`enech Badia
}\\
\textbf{
Gavin Buttimore \qquad
Charlie Deck \qquad
Joel Z Leibo \qquad
Charles Blundell
}
\\[0.5em]
DeepMind  \\[0.5em]
\texttt{\{meirefortunato, melissatan, rfaulk, stevenhansen, }\\
\texttt{adriap, buttimore, cdeck, jzl, cblundell\}@google.com}\\[0.5em]
($^\star$ Equal Contribution)\\
}

\begin{document}
\maketitle
\begin{abstract}

Memory is an important aspect of intelligence and plays a role in many deep reinforcement learning models. However, little progress has been made in understanding when specific memory systems help more than others and how well they generalize. The field also has yet to see a prevalent consistent and rigorous approach for evaluating agent performance on holdout data.
In this paper, we aim to develop a comprehensive methodology to test different kinds of memory in an agent and assess how well the agent can apply what it learns in training to a holdout set that differs from the training set along dimensions that we suggest are relevant for evaluating memory-specific generalization. To that end, we first construct a diverse set of memory tasks\footnote{\href{https://github.com/deepmind/dm_memorytasks}{https://github.com/deepmind/dm\textunderscore memorytasks}. Videos available at \href{https://sites.google.com/view/memory-tasks-suite/home}{https://sites.google.com/view/memory-tasks-suite}} that allow us to evaluate test-time generalization across multiple dimensions. Second, we develop and perform multiple ablations on an agent architecture that combines multiple memory systems, observe its baseline models, and investigate its performance against the task suite.

\end{abstract}

\input{01_introduction}
\input{02_tasks}

\input{03_model}
\input{04_results}

\input{05_discussion}

 \section*{Acknowledgements}

 We would like to thank Jessica Hamrick, Jean-Baptiste Lespiau, Frederic Besse, Josh Abramson, Oriol Vinyals, Federico Carnevale,
 Charlie Beattie, Piotr Trochim, Piermaria Mendolicchio, Aaron van den Oord, Chloe Hillier, Tom Ward, Ricardo Barreira, Matthew Mauger, Thomas Köppe, Pauline Coquinot
 and many others at DeepMind for insightful discussions, comments and feedback on this work.

%\clearpage
\small
\bibliographystyle{abbrvnat}
\bibliography{main_neurips}

\input{appendix}

\end{document}

%% file: 01_introduction.tex
\section{Introduction}
% Motivation
Humans use memory to reason, imagine, plan, and learn.  Memory is a foundational component of intelligence, and enables information from past events and contexts to inform decision-making in the present and future. Recently, agents that utilize memory systems have advanced the state of the art in various research areas including reasoning, planning, program execution and navigation, among others \citep{dnc, relrl, rmc, banino2018vector, transformer, memnets}.

% Memory: different aspects, what those are
Memory has many aspects, and having access to different kinds allows intelligent organisms to bring the most relevant past information to bear on different sets of circumstances. In cognitive psychology and neuroscience, two commonly studied types of memory are working and episodic memory. Working memory \citep{miyake_working_mem} is a short-term temporary store with limited capacity.

In contrast, episodic memory \citep{tulving1985elements} is typically a larger autobiographical database of experience (e.g. recalling a meal eaten last month) that lets one store information 
over a longer time scale and compile sequences of events into episodes \citep{epmem}.
Episodic memory has been shown to help reinforcement learning agents adapt more quickly and thereby boost data efficiency \citep{mfec, nec, eva}.
More recently, \citet{ritter2018been} shows how episodic memory can be used to provide agents with context-switching abilities in contextual bandit problems.
The transformer \citep{transformer} can be viewed as a hybrid of working memory and episodic memory that has been successfully applied to many supervised learning problems.

In this work, we explore adding such memory systems to agents and propose a consistent and rigorous approach for evaluating whether an agent demonstrates generalization-enabling memory capabilities similar to those seen in animals and humans.

% The Tasks Suite and train-holdout split: motivation, description
One fundamental principle in machine learning is to train on one set of data and test on an unseen holdout set, but it has to date been common in reinforcement learning to evaluate agent performance solely on the training set which is suboptimal for testing generalization \citep{pineau2018repro}. Also, though advances have recently been made on evaluating generalization in reinforcement learning \citep{cobbe2018quantifying} these have not been specific to memory. 

Our approach is to construct a train-holdout split where the holdout set differs from the training set along axes that we propose are relevant specifically to memory, i.e. the scale of the task and precise objects used in the task environments. For instance, if an agent learns in training to travel to an apple placed in a room, altering the room size or apple color as part of a generalization test should ideally not throw it off. 

We propose a set of environments that possess such a split and test different aspects of working and episodic memory, to help us better understand when different kinds of memory systems are most helpful and identify memory architectures in agents with memory abilities that cognitive scientists and psychologists have observed in humans. 

Alongside these tasks, we develop a benchmark memory-based agent, the Memory Recall Agent (MRA), that brings together previously developed systems thought to mimic working memory and episodic memory. This combination of a controller that models working memory, an external episodic memory, and an architecture that encourages long-term representational credit assignment via an auxiliary unsupervised loss and backpropagation through time that can `jump' over several time-steps obtains better performance than baselines across the suite. In particular, episodic memory and learning good representations both prove crucial and in some cases stack synergistically.

% Summary
To summarize, our contribution is to:
\begin{itemize}[noitemsep]
	\item Introduce a suite of tasks that require an agent to utilize fundamental functional properties of memory in order to solve in a way that generalizes to holdout data.
	\item Develop an agent architecture that explicitly models the operation of memory by integrating components that functionally mimic humans' episodic and working memory.
	\item Show that different components of our agent's memory have different effectiveness in training and in generalizing to holdout sets.
	\item Show that none of the models fully generalize outside of the train set on the more challenging tasks, and that the extrapolation incurs a greater level of degradation.
\end{itemize}

%% file: 02_tasks.tex
\section{Task suite overview}
% Task suite overview
We define a suite of 13 tasks designed to test different aspects of memory, with train-test splits that test for generalization across multiple dimensions (\href{https://github.com/deepmind/dm_memorytasks}{https://github.com/deepmind/dm \textunderscore memorytasks}). 
These include cognitive psychology tasks adapted from PsychLab \citep{psychlab} and DMLab \citep{dmlab}, and new tasks built with the Unity 3D game engine \citep{unity} that require the agent to 1) spot the difference between two scenes; 2) remember the location of a goal and navigate to it; or 3) infer an indirect transitive relation between objects. Videos with task descriptions are at \href{https://sites.google.com/view/memory-tasks-suite}{https://sites.google.com/view/memory-tasks-suite}.

\subsection{PsychLab}

Four tasks in the Memory Tasks Suite use the PsychLab environment \citep{psychlab}, which simulates a psychology laboratory in first-person. The agent is presented with a set of one or multiple consecutive images, where each set is called a `trial'. Each episode has multiple trials.

In \textbf{Arbitrary Visuomotor Mapping (AVM)} a series of objects is presented, each with an associated look-direction (e.g. up,left). The agent is rewarded if it looks in the associated direction the next time it sees a given object in the episode (Fig \ref{fig:screenshot_avm} in App. \ref{appendix:results}). \textbf{Continuous Recognition} presents a series of images with rewards given for correctly indicating whether an image has been previously shown in the episode (Fig \ref{fig:screenshot_continuous_recognition} in App. \ref{appendix:results}). In \textbf{Change Detection} the agent sees two consecutive images, separated by a variable-length delay, and has to correctly indicate if the two images differ (Fig \ref{fig:screenshot_change_detection} in App. \ref{appendix:results}). In \textbf{What Then Where} the agent is shown a single `challenge' MNIST digit, then an image of that digit with three other digits, each placed along an edge of the rectangular screen. It next has to correctly indicate the location of the `challenge' digit (Fig \ref{fig:screenshot_wtw} in App. \ref{appendix:results}).

\subsection{3D tasks}

\begin{figure}[!htb]
    \centering
    \vspace*{-0.6cm}
    \subfigure[Spot the Difference basic]{
        \includegraphics[width=0.27\linewidth]{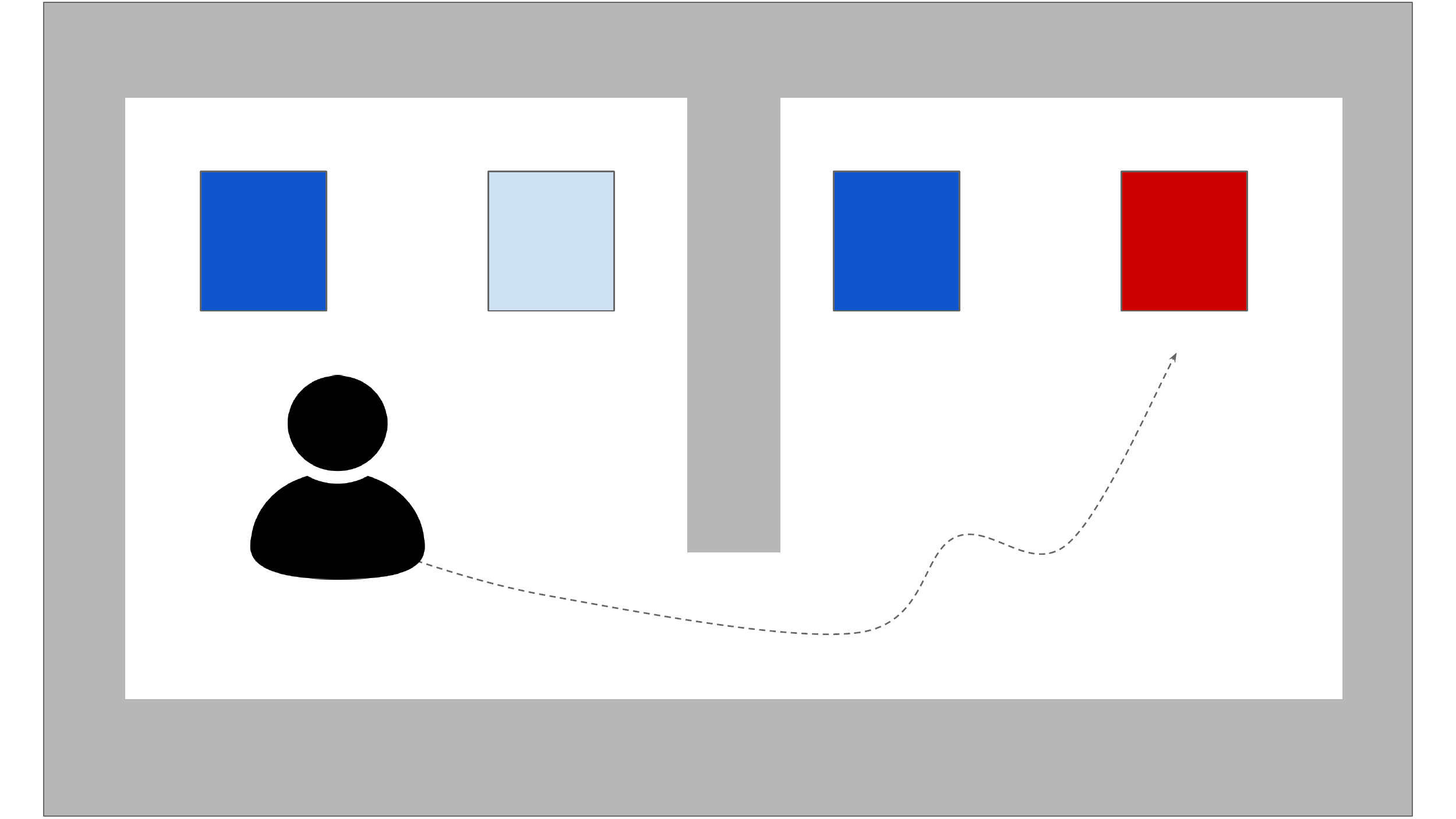}
        \label{diag:spot_diff}}
    \hfill
    \subfigure[Navigate to Goal]{
        \includegraphics[width=0.27\linewidth]{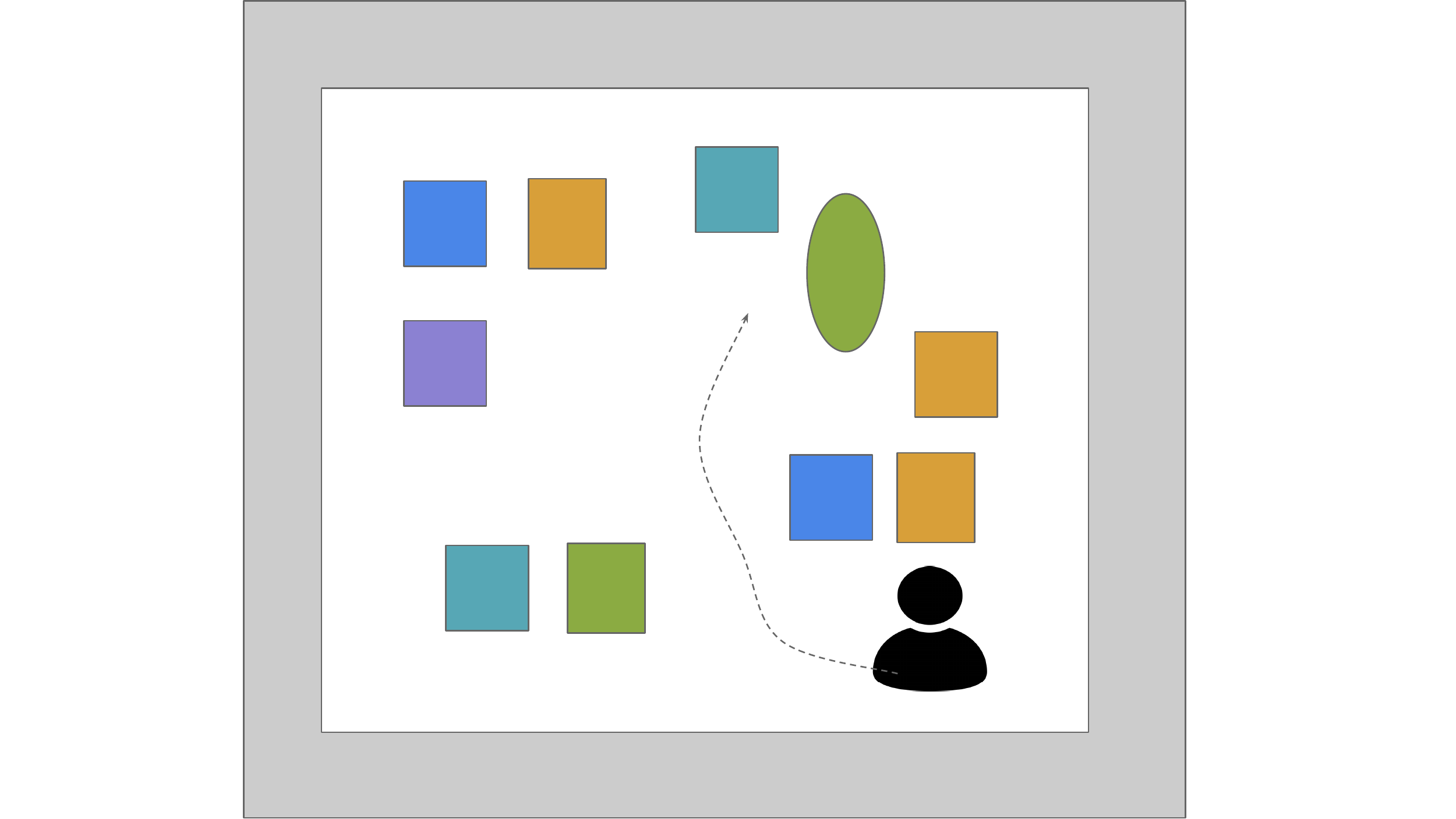}
        \label{diag:goal_navigation}}%
    \hfill    
    \subfigure[Transitive Inference]{
        \includegraphics[width=0.31\linewidth]{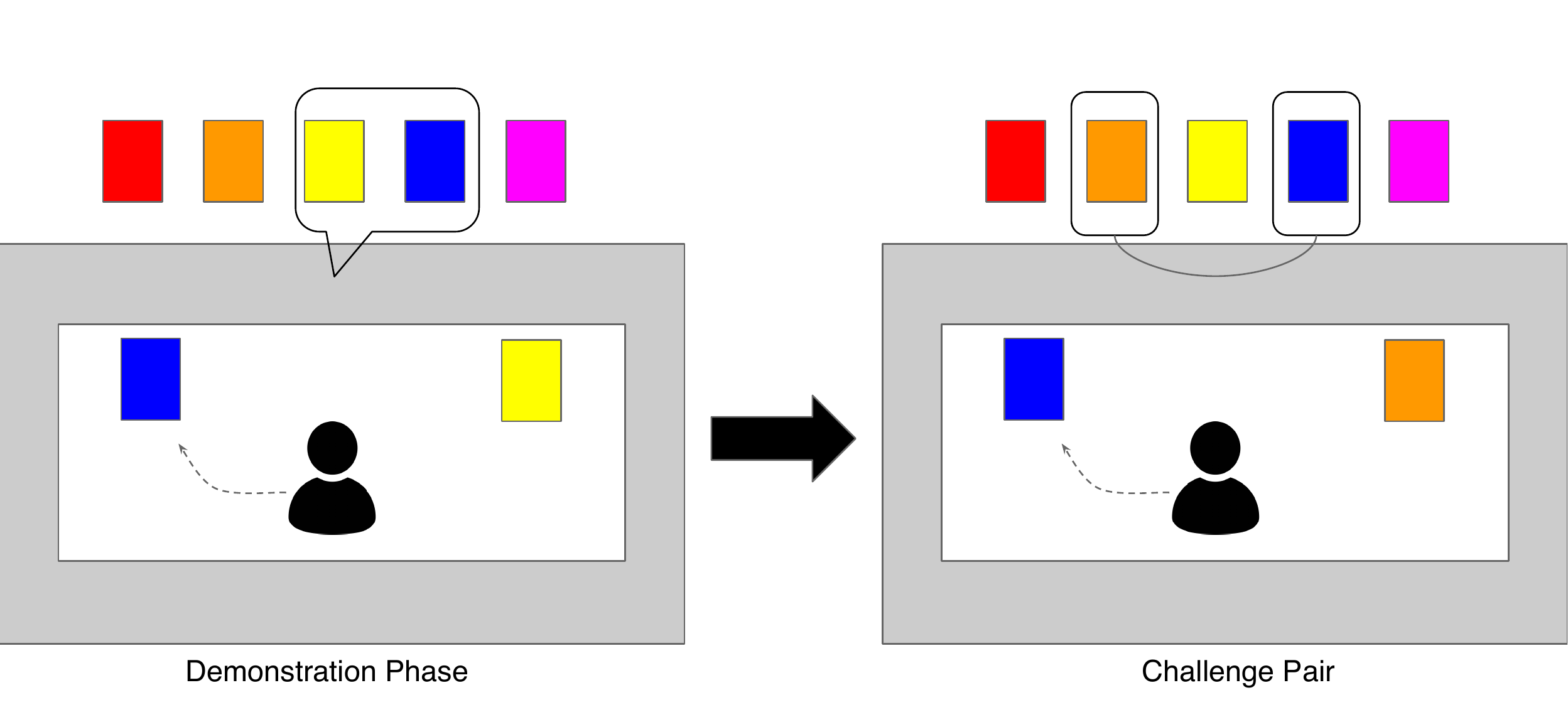}
        \label{diag:ti}}
    \caption{Task layouts for Spot the Difference, Goal Navigation, and Transitive Inference. In (a), the agent has to identify the difference between the two rooms. In (b), the agent has to go to the goal. which is represented by an oval symbol here and may be visible or not to the agent. In (c), the agent has to go to the higher-valued object in each pair. The value order is given by the transitive chain outside the room. It is shown here solely for illustration; the agent cannot see it.}
\end{figure}

\begin{figure}[!htb]
    \centering
    \vspace*{-0.5cm}
    \subfigure[Spot the Difference \it{Basic}]{
        \includegraphics[width=0.22\linewidth,height=0.15\textwidth]{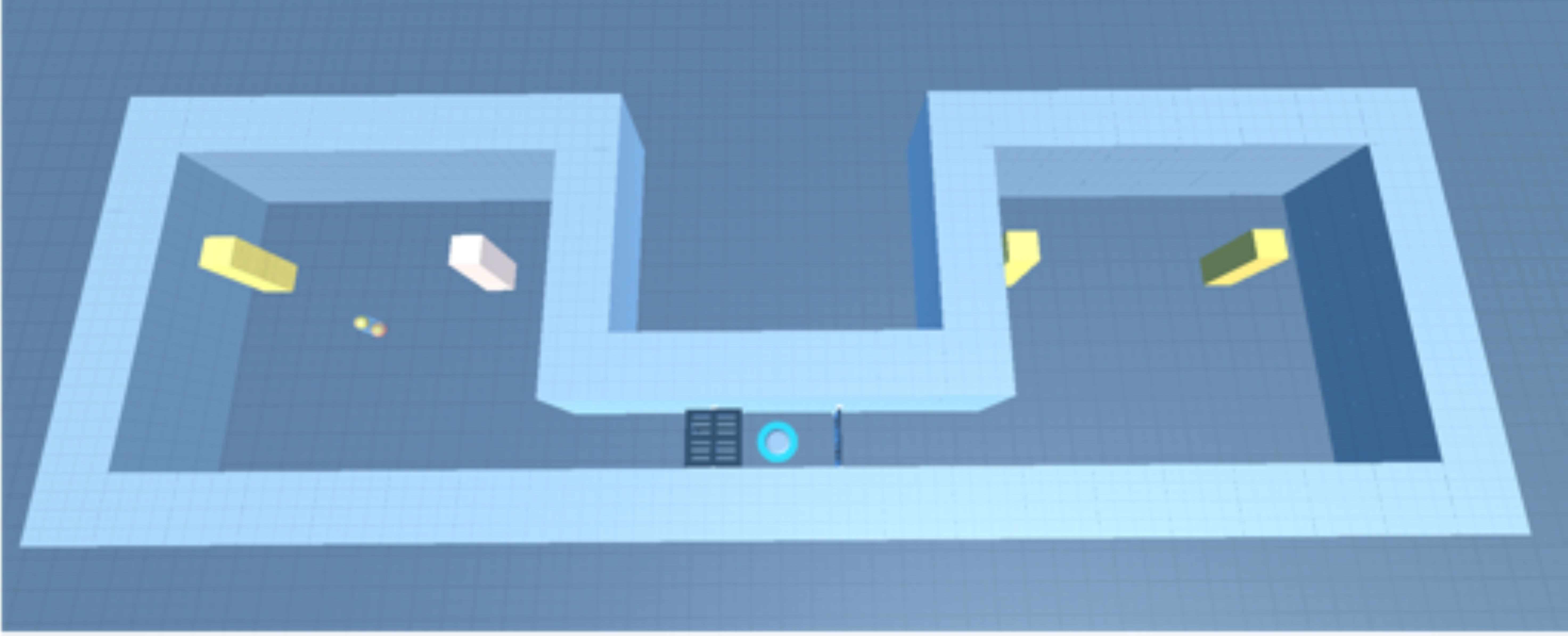}
        \label{fig:screenshot_spot_diff}}
    \hfill
    \subfigure[Spot the Difference \it{Passive}]{
        \includegraphics[width=0.22\linewidth,height=0.15\textwidth]{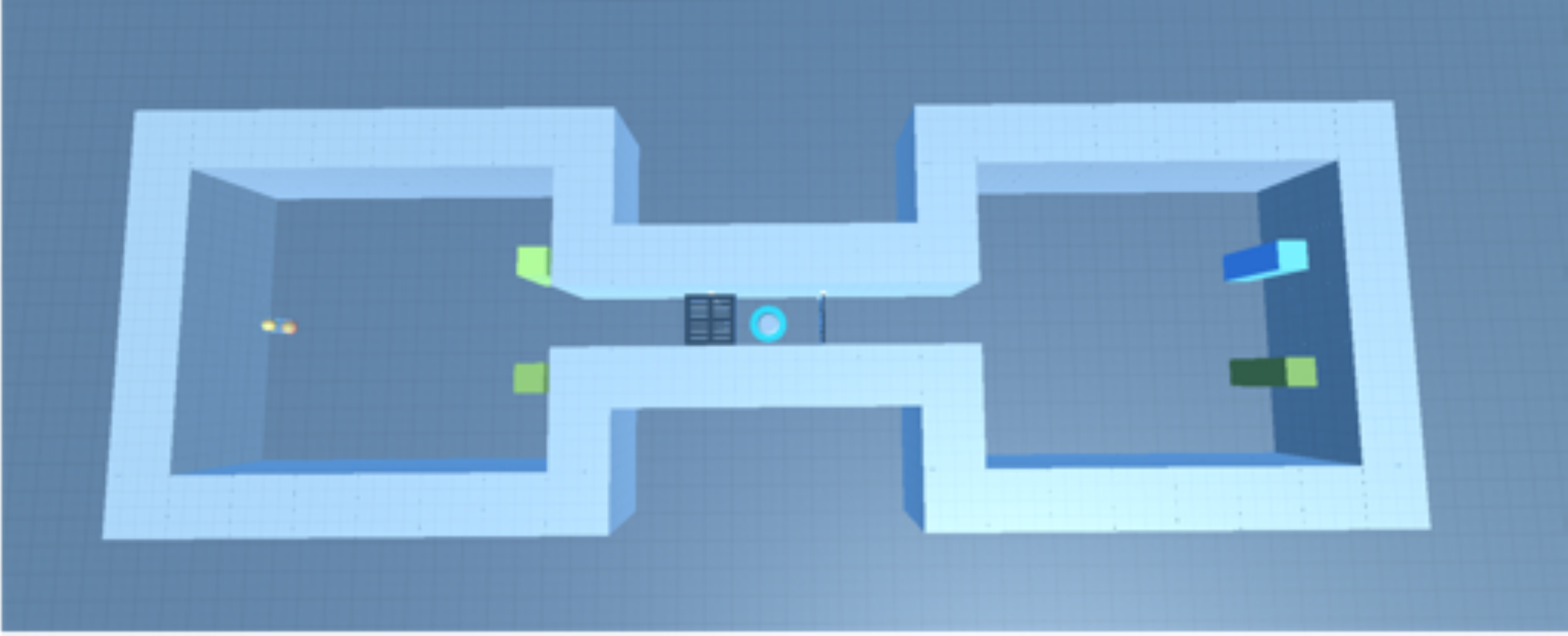}
        \label{fig:screenshot_spot_diff_passive}}%
  \hfill        
   \subfigure[Spot the Difference \it{Multi-Object}]{
        \includegraphics[width=0.22\linewidth,height=0.15\textwidth]{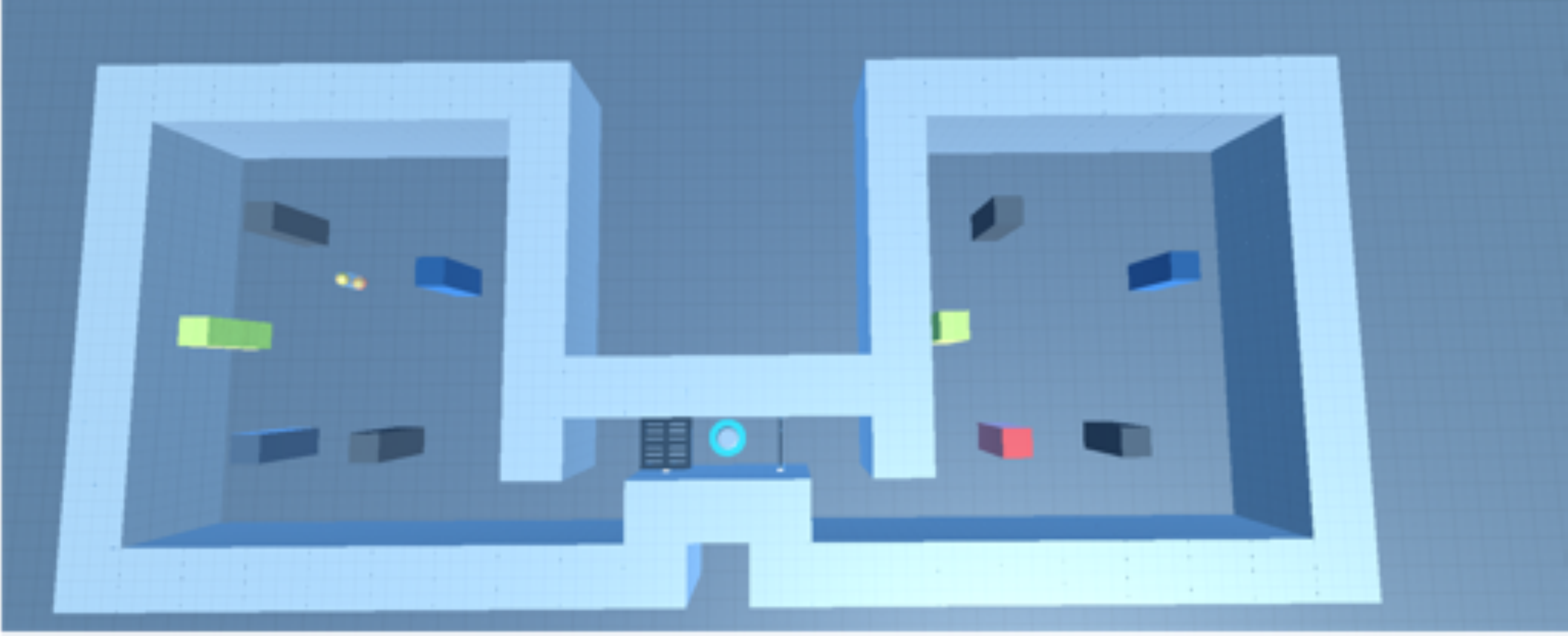}
        \label{fig:screenshot_spot_diff_ring}}
    \hfill
    \subfigure[Spot the Difference \it{Motion}]{
        \includegraphics[width=0.22\linewidth,height=0.15\textwidth]{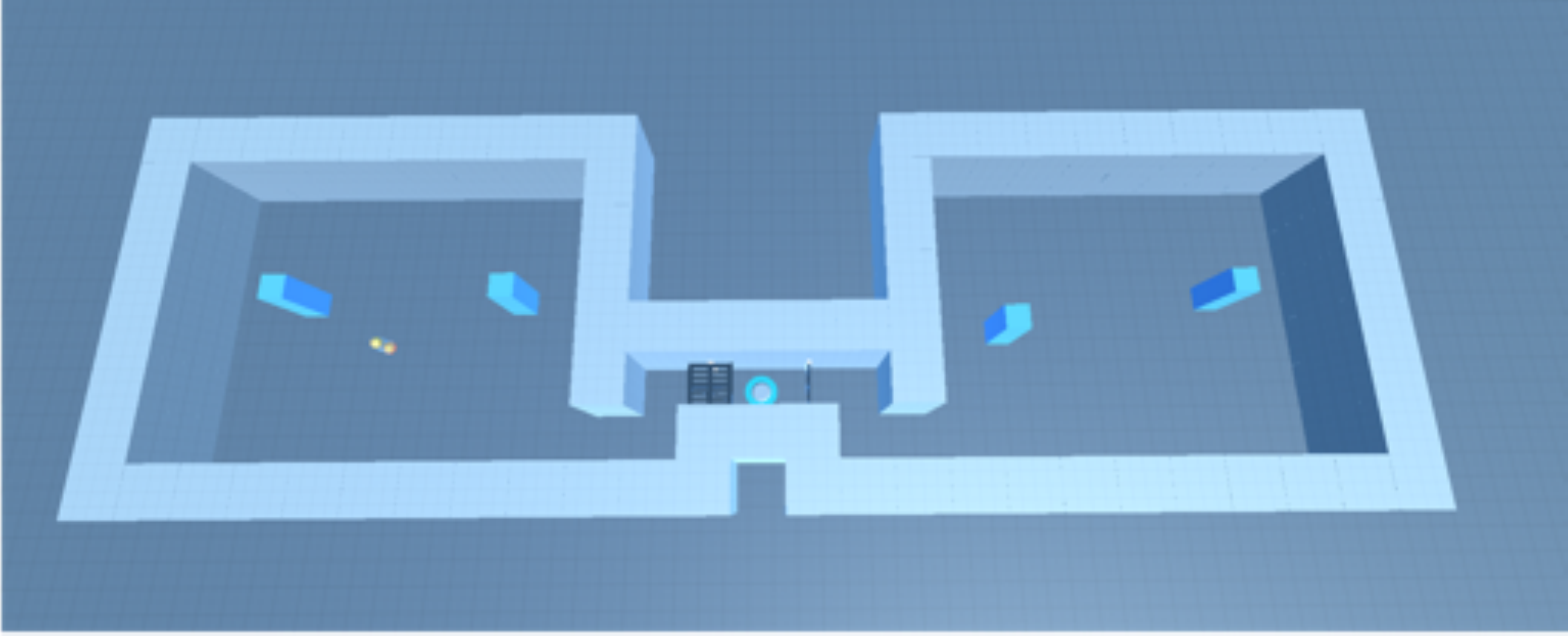}
        \label{fig:screenshot_spot_diff_motion}}%
    \caption{Spot the Difference tasks. (a) All the tasks in this family are variants of this basic setup, where each room contains two blocks. (b) By placing Room 1's blocks right next to the corridor entrance, we guarantee that the agent will always see them. (c) The number of objects varies. (d) Instead of differing in color between rooms, the altered block follows a different motion pattern.}
    \label{fig:screenshot_spot_diff_all}
\end{figure}

\textbf{Spot the Difference}:
This tests whether the agent can correctly identify the difference between two nearly identical scenes (Figure \ref{diag:spot_diff}). The agent has to move from the first to the second room, with a `delay' corridor in between. See Fig. \ref{fig:screenshot_spot_diff_all} for the four different variants.

\begin{figure}[!htb]
    \centering
    \vspace*{-0.5cm}
    \subfigure[Invisible Goal Empty Arena]{
        \includegraphics[width=0.2\linewidth,height=0.21\textwidth]{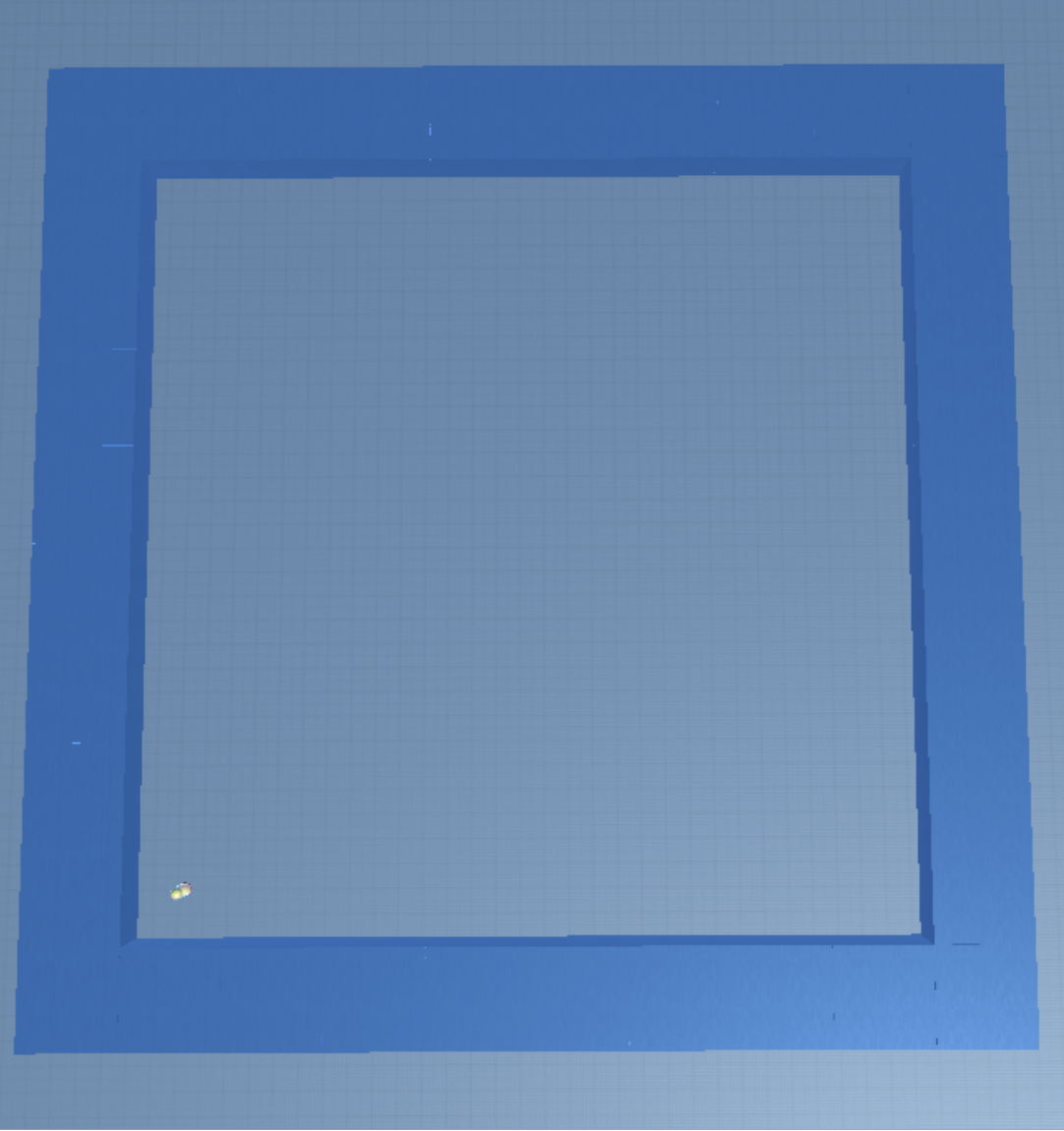}
        \label{fig:screenshot_empty_room_watermaze}}
    \hfill
    \subfigure[Invisible Goal, With Buildings]{
        \includegraphics[width=0.2\linewidth,height=0.21\textwidth]{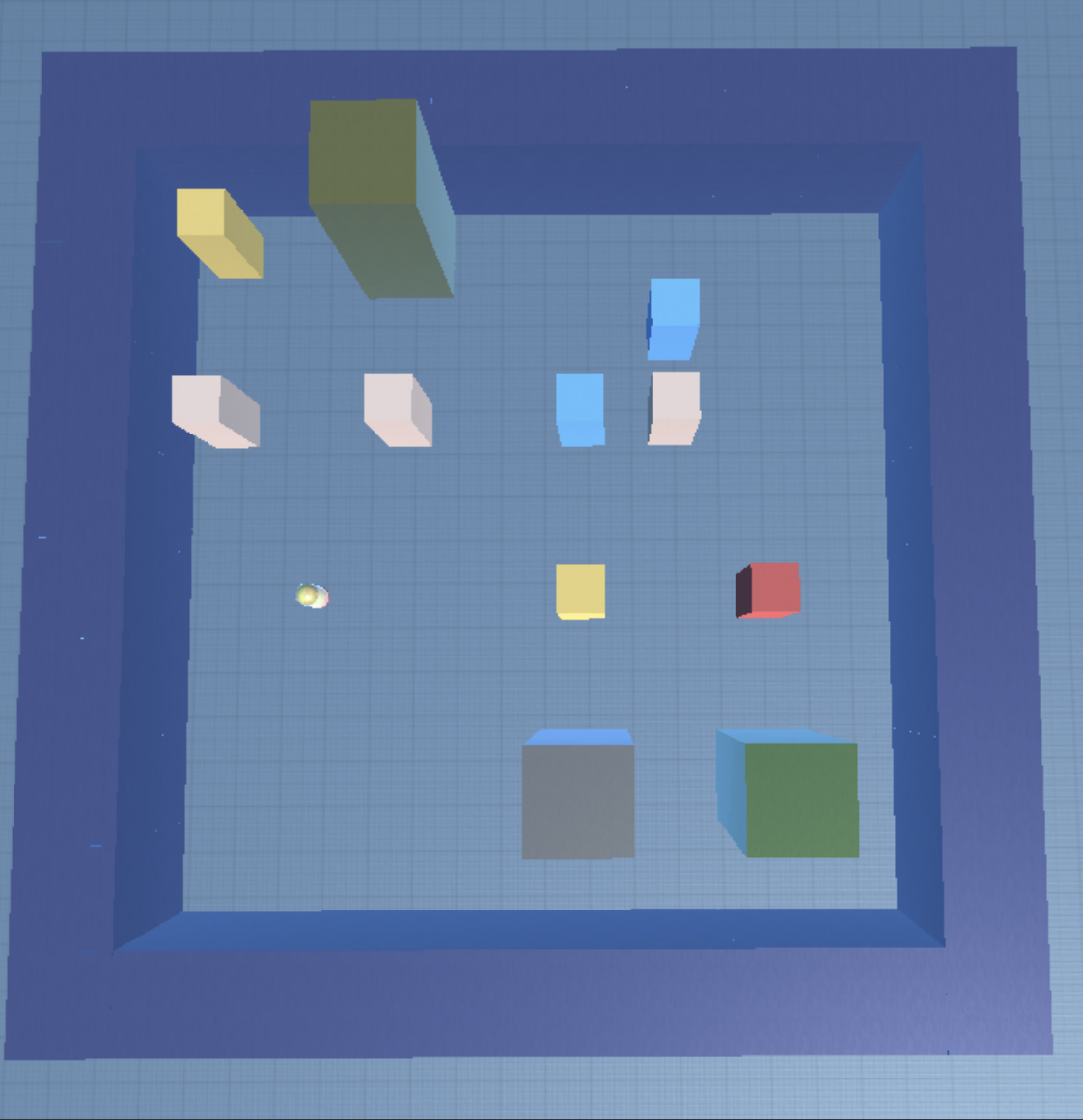}
        \label{fig:screenshot_mini_city_watermaze}}%
   \hfill     
   \subfigure[Visible Goal With Buildings]{
        \includegraphics[width=0.2\linewidth,height=0.21\textwidth]{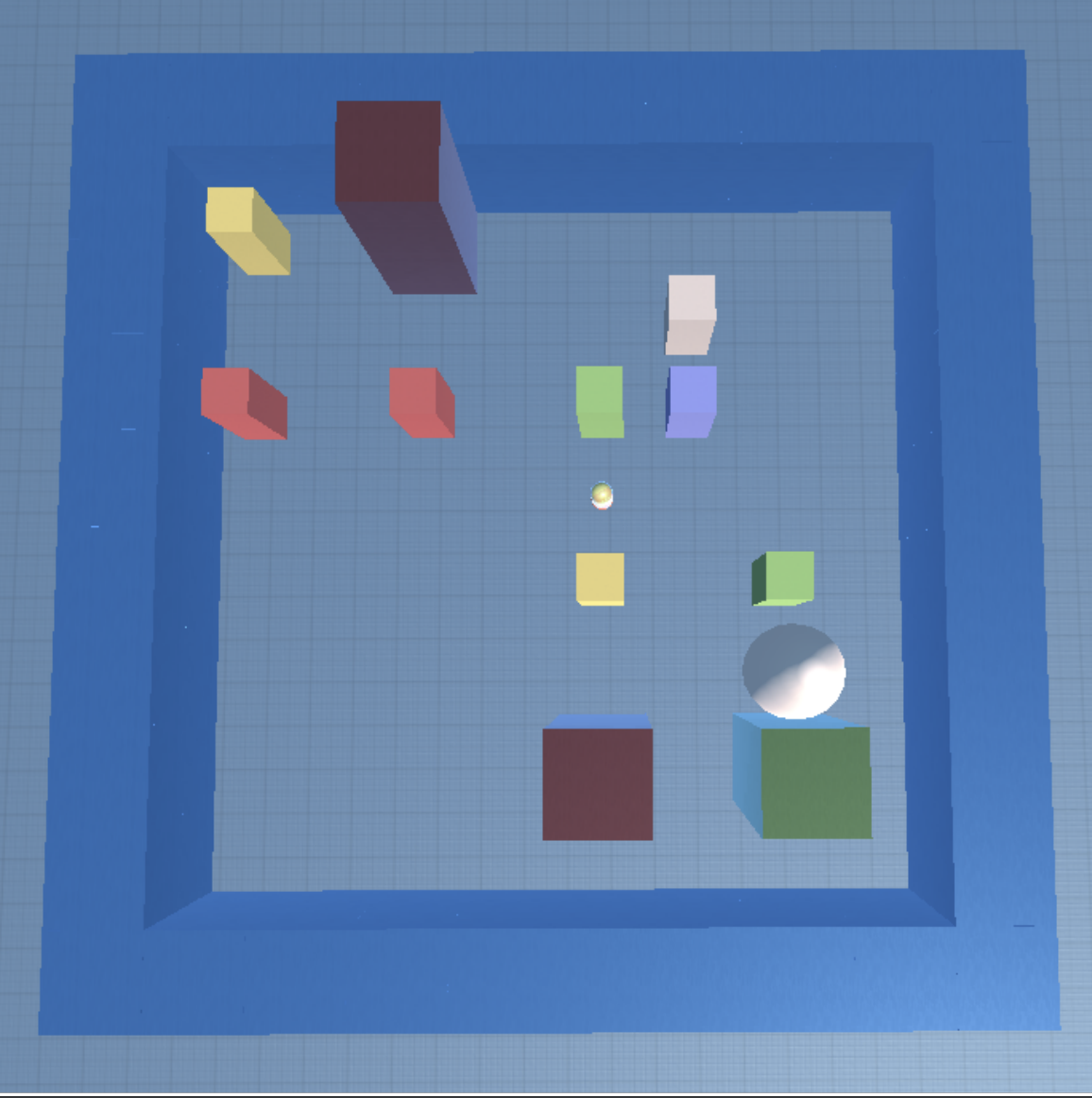}
        \label{fig:screenshot_mini_city_visible_goal}}
    \hfill
    \subfigure[Visible Goal Procedural Maze]{
        \includegraphics[width=0.2\linewidth,height=0.21\textwidth]{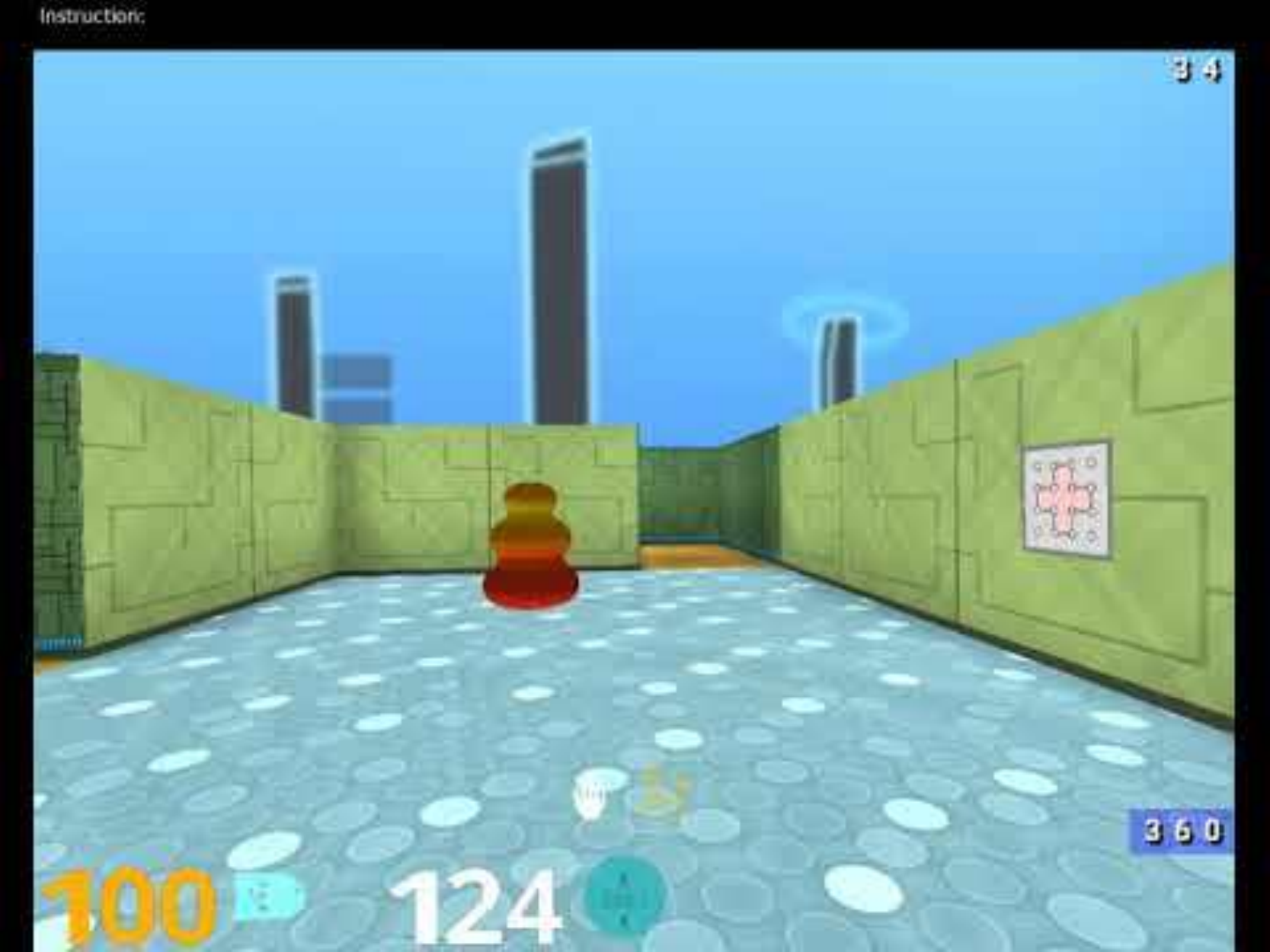}
        \label{fig:screenshot_explore_goal_locations}}%
    \caption{Goal Navigation tasks. (a) The arena has no buildings, agent must navigate by skybox. (b) There are rectangular buildings at fixed, non-randomized locations in the arena. (c) As in (b), but the goal appears as an oval. (d) A visible goal in a procedurally generated maze.}
    \label{fig:navigate_tasks}
\end{figure}

\textbf{Goal Navigation}:
This task family was inspired by the Morris Watermaze \citep{miyake_working_mem} setup used with rodents in behavioral neuroscience. The agent is rewarded every time it successfully reaches the goal; once it gets there it is respawned randomly in the arena and has to find its way back to the goal. The goal location is re-randomized at the start of episode (Fig. \ref{diag:goal_navigation}, Fig. \ref{fig:navigate_tasks}). 

\textbf{Transitive Inference}:
 
This task tests if an agent can learn an overall transitive ordering over a chain of objects, through being presented with ordered pairs of adjacent objects (See Fig. \ref{diag:ti} and App. \ref{appendix:results}).

\subsection{Scale and Stimulus Split} 
\label{scale_stimulus_split}

To test how well the agent can generalize to holdout data after training, we create per-task holdout levels that differ from the training level along a scale and a stimulus dimension. 
The scale dimension is intended to capture something about the memory demand of the task: e.g., a task with a longer time delay between events that must be related should be harder than one with a short delay.
The stimulus dimension is to guard against trivial overfitting to the particular visual input presented to the input: the memory representation should be more abstract than the particular colour of an object. 

The training level comprises a `small' and `large' scale version of the task. When training the agent we uniformly sample between these two scales. As for the holdout levels, one of them -- `holdout-interpolate' -- corresponds to an interpolation between those two scales (call it `medium') and the other, `holdout-extrapolate', corresponds to an extrapolation beyond the `large' scale (call it `extra-large'). 
Alterations made for each task split and their settings are in Table \ref{tab:alterations} in App. \ref{appendix:level_config}.

%% file: 03_model.tex
\section{The Memory Recall Agent}
\label{sec:model}

% The Agent.

\begin{figure}[!htb]
    \centering
    \vspace{-0.5cm}
    \subfigure[Architecture of the MRA.]{
        \includegraphics[width=0.4\linewidth]{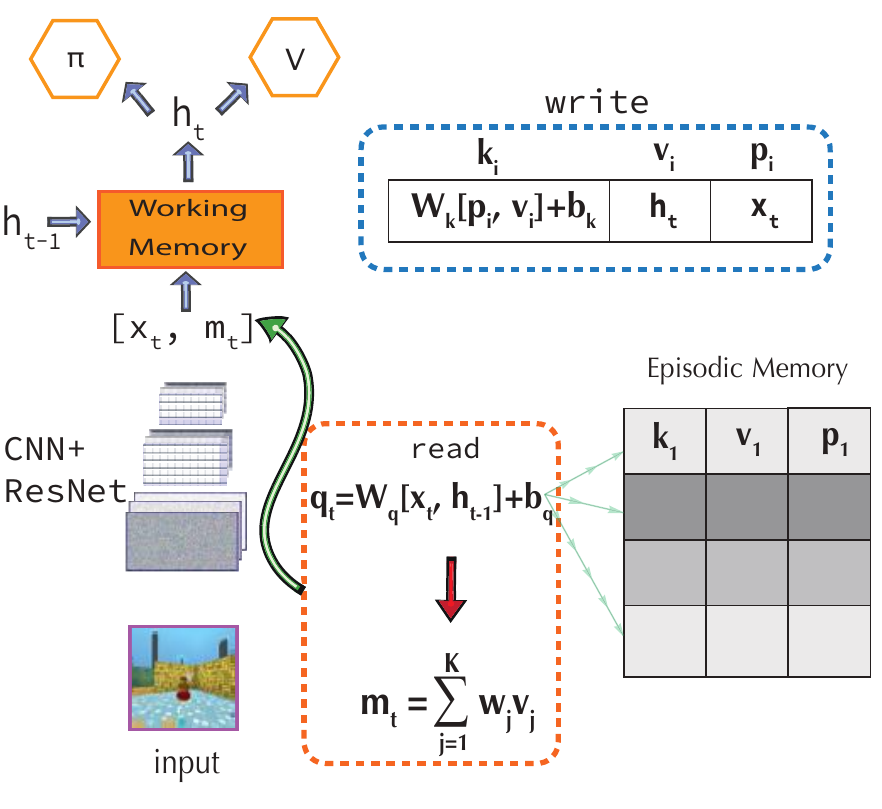}
        \label{fig:episodic_core}}
    \hfill
    \subfigure[Contrastive Predictive Coding loss for MRA.]{
        \includegraphics[width=0.52\linewidth]{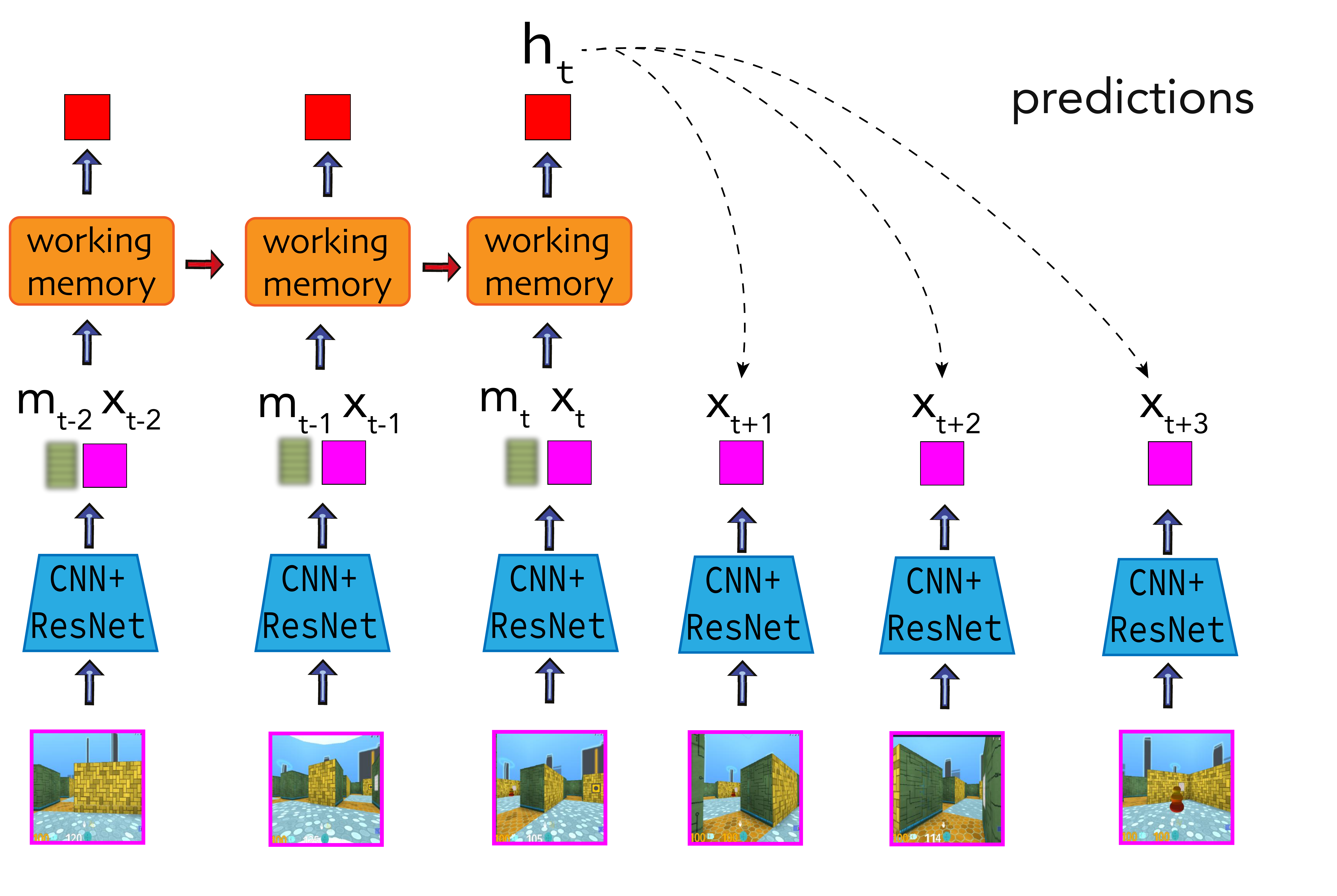}
        \label{fig:cpc}}%
    \caption{The Memory Recall Agent (MRA) architecture. Here $p_i$ is the pixel input embedding $x_t$ from step $t$, and $v_i$ is the LSTM hidden state $h_t$. $k_i$ is the key used for reading; we compute it from $p_i$ and $v_i$. $q_t$ is the query we use to compare against keys to find nearest neighbors.}
\end{figure}

Our agent, the Memory Recall Agent (MRA), incorporates five components: 1) a pixel-input convolutional, residual network, 2) a working memory, 3) a slot-based episodic memory, 4) an auxiliary contrastive loss for representation learning \citep{cpc}, 5) a jumpy backpropagation-through-time training regime.
Our agent architecture is shown in Figure \ref{fig:episodic_core}. The overall agent is built on top of the IMPALA model \citep{dmlab_impala} and is trained in the same way with the exceptions described below. Component descriptions are below.

\paragraph{Pixel Input}

Pixel input is fed to a convolutional neural network, as is common in recent agents, followed by a residual block \citep{resnet}.
The precise hyper-parameters are given in \ref{appendix:resnet}: we use three convolutional layers followed by two residual layers.
The output of this process is $x_t$ in Figure \ref{fig:episodic_core} and serves as input to three other parts of the network: 1) part of the input to the working memory module, 2) in the formation of keys and queries for the episodic memory, 3) as part of the target for the contrastive predictive coding. 
\paragraph{Working Memory}

Working memory is often realized through latent recurrent neural networks (RNNs) with some form of gating, such as LSTMs and Relational Memory architectures \citep{lstm, rmc}. These working memory models calculate the next set of hidden units using the current input and the previous hidden units. Although models which rely on working memory can perform well on a variety of problems, their ability to tackle dependencies and represent variables over long time periods is limited. The short-term nature of working memory is pragmatically, and perhaps unintentionally, reflected in the use of truncated backprop through time and the tendency for gradients through these RNNs to explode or vanish.
Our agent uses an LSTM as a model of working memory.
As we shall see in experiments, this module is able to perform working memory--like operations on tasks: i.e., learn calculations involving short-term memory.
As depicted in Figure \ref{fig:episodic_core}, the LSTM takes as input $x_t$ from the pixel input network and $m_t$ from the episodic memory module. 
As in \citet{dmlab_impala}, the LSTM has two heads as output, producing the policy $\pi$ and the baseline value function $V$.
In our architecture these are derived from the output from the LSTM, $h_t$. $h_t$ is also used to form episodic memories, as described below.

\paragraph{Episodic Memory (MEM)}

If our agent only consisted of the working memory and pixel input described above, it would be almost identical to the model in IMPALA \citep{dmlab_impala}, an already powerful RL agent.
But MRA also includes a slot-based episodic memory module as that can store values more reliably and longer-term than an LSTM, is less susceptible to the intricacies of gradient propagation, and its fundamental operations afford the agent different abilities (as observed in our experiments). The MEM in MRA has a key-value structure which the agent reads from and writes to at every time-step (see Fig. \ref{fig:episodic_core}). MRA implements a mechanism to learn how to store summaries of past experiences and retrieve relevant information when it encounters similar contexts. The reads from memory are used as additional inputs to the neural network (controller), which produces the model predictions. This effectively augments the controller's working memory capabilities with experiences from different time scales retrieved from the MEM, which facilitate learning long-term dependencies, a difficult task when relying entirely on backpropagation in recurrent architectures \citep{lstm,dnc,transformer}.

The MEM has a number of slots, indexed by $i$.
Each slot stores activations from the pixel input network and LSTM from previous times $t_i$ in the past. The MEM acts as a fixed-size circular (first-in-first-out) buffer: New keys and values are added, overwriting the least recently added entry if there are no unused slots available. The contents of the episodic memory buffer is wiped at the end of each episode.

    \subparagraph{Memory Writing} Crucially, writing to episodic memory is done without gradients.
    At each step a free slot is chosen for writing, denoted $i$.
    Next, the following is stored:
    \begin{align}
    p_i \leftarrow x_t; \quad&  
    v_i \leftarrow  h_t; \quad
    \label{eq:key} k_i \leftarrow  W_k [p_i, v_i] + b_k
    \end{align}
    where $p_i$ is the pixel input embedding from step $t$ and $v_i$ is the LSTM hidden state (if the working memory is something else, e.g. a feedforward, this would be the output activations).
    $k_i$ is the key, used for reading (described below), computed as a simple linear function of the other two values stored. Caching the key speeds up memory reads significantly. However, the key can become stale as the weights and biases, $W_k$ and $b_k$ are learnt (the procedure for learning them is described below under Jumpy Backpropagation).
    In our experiments we did not see an adverse effect of this staleness.

\subparagraph{Memory Reading}
The agent uses a form of dot-product attention \citep{neur_mach_trans} over its MEM, to select the most relevant events to provide as input $m_t$ to the LSTM.
The query $q_t$ is a linear transform of the pixel input embedding $x_t$ and the LSTM hidden state from the previous time-step $h_{t-1}$,
with weight $W_q$ and bias $b_q$.
\begin{align}
\label{eq:query} q_t &= W_q [x_t, h_{t-1}] + b_q
\end{align}

The query $q_t$ is then compared against the keys in MEM as in \citet{nec}: Let $(p_j, v_j, k_j)$, $1\le j \le K$ be the $K$ nearest neighbors to $q_t$ from MEM, under an L2 norm between $k_j$ and $q_t$.
\begin{align}
\label{eq:retrieval} m_t &= \sum_{j=1}^K w_j v_j&\quad 
where\quad
w_j &\propto \frac{1}{\epsilon + || q_t - W_k[p_j, v_j] - b_k||^2_2}
\end{align}
We compute a weighted aggregate of the values ($v_j$) of the $K$ nearest neighbors, weighted by the inverse of each neighbor-key's distance to the query. 
Note that the distance is re-calculated from values stored in the MEM, via the linear projection $W_k, b_k$ in (\ref{eq:key}). We concatenate the resulting weighted aggregate memory $m_t$ with the embedded pixel input $x_t$, and pass it as input to the working memory as shown in Figure \ref{fig:episodic_core}.

\paragraph{Jumpy backpropagation}
\label{subsec:jumpy}

We now turn to how gradients flow into memory writes. Full backpropagation can become computationally infeasible as this would require backpropagation into every write that is read from and so on.
Thus as a new $(p_i, v_i, k_i)$-triplet is added to the MEM, there are trade-offs to be made regarding computational complexity versus performance of the agent. To make it more computationally tractable, we place a stop-gradient in the memory write. In particular, the write operation for the key in \eqref{eq:key} becomes:
\begin{align}
\label{eq:keyjumpy} k_i &\leftarrow  W_k [\text{SG}(p_i), \text{SG}(v_i)] + b_k
\end{align}
where $\text{SG}(\cdot)$ denote that the gradients are stopped.
This allows the parameters $W_k$ and $b_k$ to receive gradients from the loss during writing and reading, while at the same time bounding the computational complexity as the gradients do not flow back into the recurrent working memory (or via that back into the MEM).
To re-calculate the distances, we want to use these learnt parameters rather than, say, random projection, so we need to store the arguments $x_t$ and $h_t$ of the key-generating linear transform $W_k, b_k$ for all previous time-steps. Thus in the MEM we store the full $(p_i, v_i, k_i)$-triplet, where $p_i = x_{t_i}$, $v_i = h_{t_i}$ and $t_i$ is the step that write $i$ was made.
We call this technique `jumpy backpropagation' because the intermediate steps between the current time-step $t$ and the memory write step $t_i$ are not taken into account in the gradient updates. 

This approach is similar to Sparse Attentive Backtracking \citep[SAB]{sab} which uses sparse replay by passing gradients only through memories selected as relevant at each step.  Our model differs in that it does not have a fixed chunking scheme and does not do full backpropagation through the architecture (which in our case becomes quickly intractable). Our approach has minimal computational overhead as we only recompute the keys for the nearest neighbors.

\paragraph{Auxiliary Unsupervised Losses}
\label{subsec:unsupervised_losses}

An agent with good memory provides a good basis for forming a rich representation of the environment, as it captures a history of the states visited by the agent. This is the primary basis for many rich probabilistic state representations in reinforcement learning such as belief states and predictive state representations \citep{littman2002predictive}.
Auxiliary unsupervised losses can significantly improve agent performance \citep{unreal}.
Recently it has been shown that agents augmented with one-step contrastive predictive coding \citep[CPC]{cpc} can learn belief state representations of the environment \citep{belief_repres}.
Thus in MRA we combine the working and episodic memory mechanisms listed above with a CPC unsupervised loss to imbue the agent with a rich state representation. The CPC auxiliary loss is added to the usual RL losses, and is of the following form:
\begin{align}
    \sum_{\tau=1}^N \text{CPCLoss}\left[h_t ; x_{t+1}, x_{t+2}, \dots, x_{t+\tau} \right]
\end{align}
where $\text{CPCLoss}$ is from \citet{cpc}, $h_t$ is the working memory hidden state, and $x_{t+\tau}$ is the encoding pixel input at $\tau$ steps in the future. $N$ is the number of CPC steps (typically $10$ or $50$ in our experiments). See Figure \ref{fig:cpc} for an illustration and further details and equations elaborating on this loss in App. \ref{appendix:cpc}.

Reconstruction losses have also been used as an auxiliary task \citep{unreal,merlin} and we include this as a baseline in our experiments.
Our reconstruction baseline minimizes the L2 distance between the predicted reward and predicted pixel input and the true reward and pixel input, using the working memory state $h_t$ as input. Details of this baseline are given in App. \ref{appendix:recon}.

%% file: 04_results.tex
\section{Experiments}

\paragraph{Setup}

We ran 10 ablations on the MRA architecture, on the training and the two holdout levels:
\begin{itemize}[noitemsep]
	\item Working Memory component: Either feedforward neural network (`FF' for short) or LSTM. The LSTM-only baseline corresponds to IMPALA \citep{dmlab_impala}.
	\item With or without using episodic memory module (`MEM').
	\item With or without auxiliary unsupervised loss (either CPC or reconstruction loss (`REC')).
	\item With or without jumpy backpropagation, for MRA (i.e. LSTM + MEM + CPC)
\end{itemize}

Given that the experiments are computationally demanding, we only performed small variations within as part of our hyper-parameter tuning process for each task (see App. \ref{hypers}). 

We hypothesize that in general the agent should perform the best in training, somewhat worse on the holdout-interpolation level and the worst on the holdout-extrapolation level. That is, we expect to see a \textit{generalization gap}. Our results validated this hypothesis for the tasks that were much harder for agents than for humans.

\subsection{Full comparison}

\begin{figure}[h!]
 \vspace*{-0.5em}
  \includegraphics[width=\textwidth]{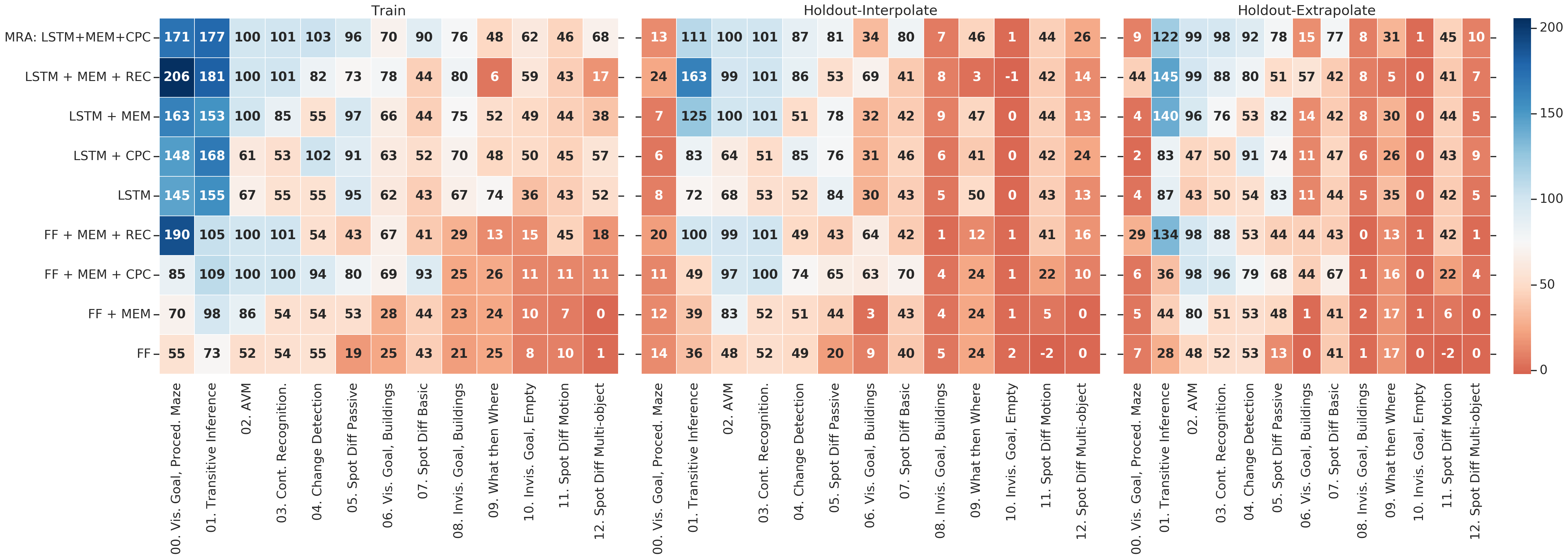}
  \vspace*{-2em}
  \caption{Heatmap of ablations per task sorted by normalized score for Train, Holdout-Interpolate, Holdout-Extrapolate. The same plot with standard errors is in App. \ref{appendix:results} Fig. \ref{appendix:heatmap}.}
  \label{fig:heatmap}
\end{figure}

We computed human-normalized scores (details in App. \ref{appendix:results}) and plotted them into a heatmap (Fig \ref{fig:heatmap}) sorted such that the model with the highest train scores on average is the top row and the task with highest train scores on average is the leftmost column.
The heatmap suggests that the MRA architecture, LSTM + MEM + CPC, broadly outperforms the other models (App. \ref{appendix:results} Table \ref{tab:ranked_ablations}). 
This ranking was almost always maintained across train and holdout levels, despite MRA performing worse than the LSTM-only baseline on \textit{What Then Where}. \textit{What Then Where} was one of the tasks where all models did poorly, along with \textit{Spot the Difference: Multi-Object}, \textit{Spot the Difference: Multi-Object}, \textit{Spot the Difference: Multi-Object} (rightmost columns in heatmap). At the other end of the difficulty spectrum, LSTM + MEM had superhuman scores on \textit{Visible Goal Procedural Maze} in training and on \textit{Transitive Inference} in training and holdout, and further adding CPC or REC boosted the scores even higher.

\begin{wrapfigure}[15]{r}{0.5\textwidth}
%   \begin{center}
  \centering
  \vspace*{-1.2cm}
  \includegraphics[width=0.5\textwidth]{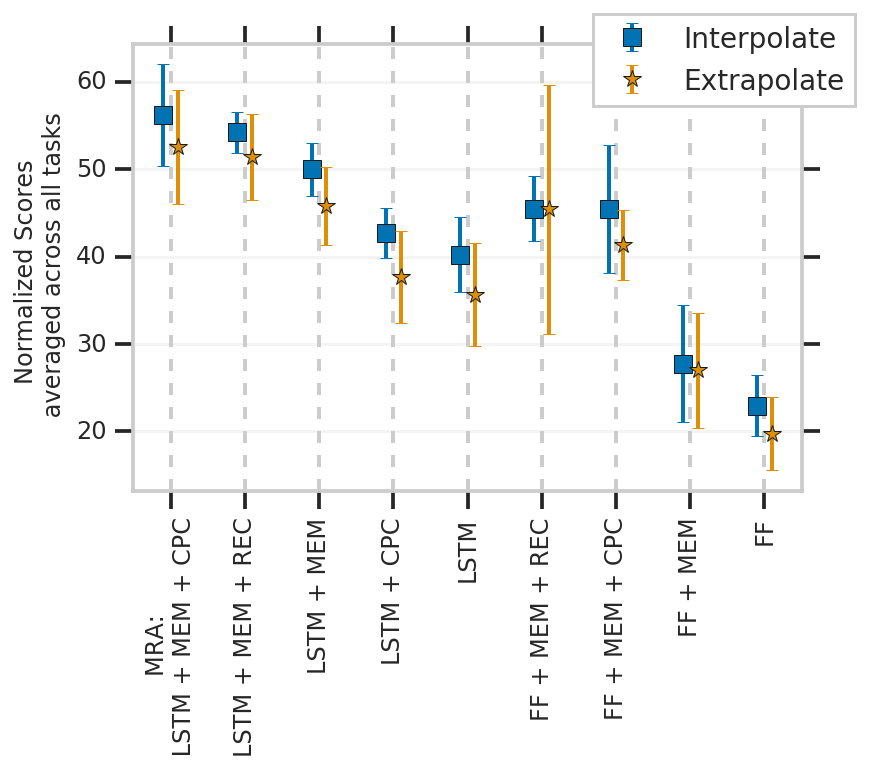}
  \vspace*{-0.8cm}
%   \end{center}
  \caption{Normalized scores averaged across tasks.}
  \label{fig:rebuttal_generalization}
\end{wrapfigure}

\subsection{Results}

Different memory systems worked best for different kinds of tasks, but the MRA architecture's combination of LSTM + MEM + CPC did the best overall on training and holdout (Fig. \ref{fig:rebuttal_generalization}). Removing jumpy backpropagation from MRA hurt performance in five Memory Suite tasks (App. \ref{appendix:results} Fig. \ref{fig:JB_ablation_curves_better}), while performance was the same in the remaining ones (App. \ref{appendix:results} Fig. \ref{fig:JB_ablation_curves_same1} and \ref{fig:JB_ablation_curves_same2}).

\begin{figure}[!htb]
    \centering
    \vspace*{-0.6cm}
    \subfigure{
        \includegraphics[width=0.38\linewidth,height=0.25\textwidth]{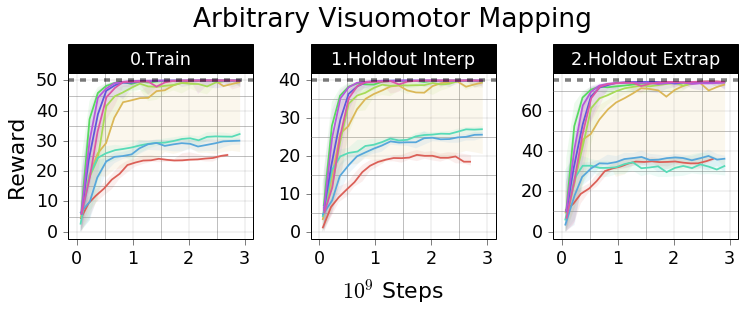}
        }
    \hfill
    \subfigure{
        \includegraphics[width=0.58\linewidth,height=0.25\textwidth]{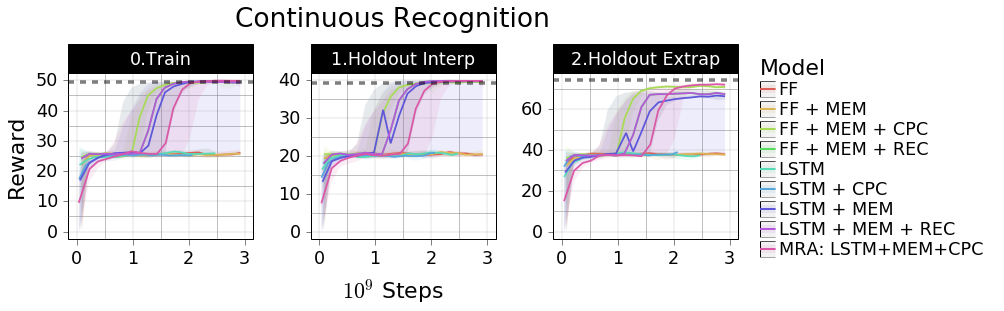}
       }%
   \vspace*{-2em}
    \subfigure{
        \includegraphics[width=0.38\linewidth, height=0.25\textwidth]{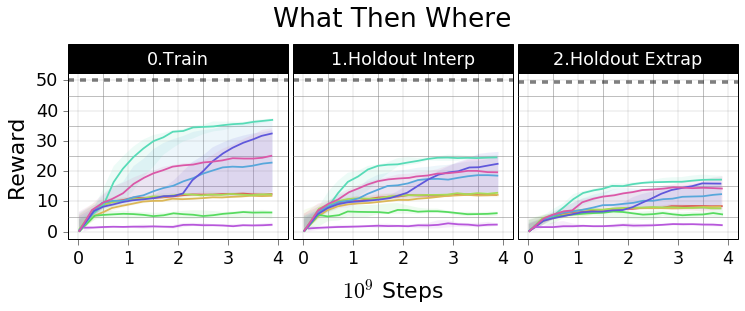}
       }
    \hfill
    \subfigure{
        \includegraphics[width=0.58\linewidth, height=0.25\textwidth]{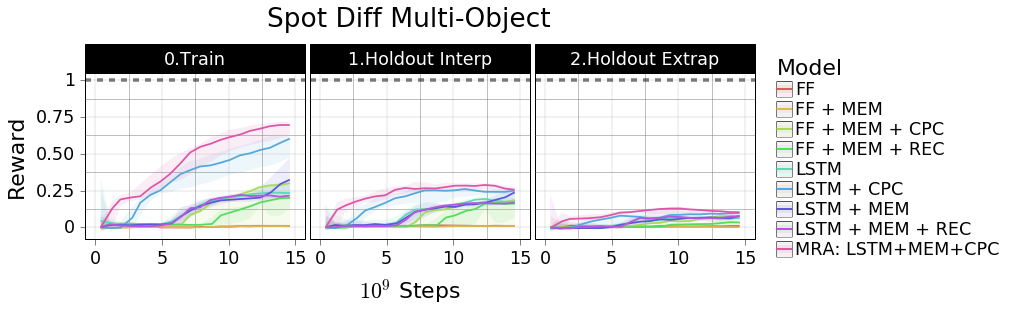}
        }%
    \vspace*{-0.5cm}
    \caption{Generalization gap is smaller for \textit{AVM} and \textit{Continuous Recognition}, larger for \textit{What Then Where} and \textit{Spot the Difference: Multi-Object}. Dotted lines indicate human baseline scores. See other curves in App. \ref{appendix:results} Fig. \ref{fig:curve_all}.}
    \label{plot:decay}
    \setlength{\belowcaptionskip}{-20pt}
\end{figure}

\paragraph{Generalization gap widens as task difficulty increases}
The hypothesized generalization gap was minimal for some tasks e.g. \textit{AVM} and \textit{Continuous Recognition} but significant for others e.g. \textit{What Then Where} and \textit{Spot the Difference: Multi-Object} (Fig \ref{plot:decay}). We observed that the gap tended to be wider as the task difficulty went up, and that in PsychLab, the two tasks where the scale was the number of trials seemed to be easier than the other two tasks where the scale was the delay duration.

\paragraph{MEM critical on some tasks, is enhanced by auxiliary unsupervised loss} Adding MEM improved scores on nine tasks in training, six in holdout-interpolate, and six in holdout-extrapolate. Adding MEM alone, without an auxiliary unsupervised loss, was enough to improve scores on \textit{AVM} and \textit{Continuous Recognition}, all Spot the Difference tasks except \textit{Spot the Difference: Multi-Object}, all Goal Navigation tasks except \textit{Visible Goal Procedural Maze}, and also for \textit{Transitive Inference}. 

Adding MEM helped to significantly boost holdout performance for \textit{Transitive Inference}, \textit{AVM}, and \textit{Continuous Recognition}. For the two PsychLab tasks this finding was in line with our expectations, since they both can be solved by memorizing single images and determining exact matches and thus an external episodic memory would be the most useful. For \textit{Transitive Inference}, in training MEM helped when the working memory was FF but made little difference on an LSTM, but on holdout MEM helped noticeably for both FF and LSTM. 
In \textit{Change Detection} and \textit{Multi-Object}, adding MEM alone had little or no effect but combining it with CPC or REC provided a noticeable boost.

\paragraph{Synergistic effect of MEM + CPC, for LSTM}
On average, adding either the MEM + CPC stack or MEM + REC stack to any working memory appeared to improve the agent's ability to generalize to holdout levels (Fig. \ref{fig:rebuttal_generalization}). Interestingly, on several tasks we found that combining MEM + CPC had a synergistic effect when the working memory was LSTM: The performance boost from adding MEM + CPC was larger than the sum of the boost from adding MEM or CPC alone. We observed this phenomenon in seven tasks in training, six in holdout-interpolate, and six in holdout-extrapolate. Among these, the tasks where there was MEM + CPC synergy across training, holdout-interpolate, and holdout-extrapolate were: the easiest task, \textit{Visible Goal Procedural Maze}; \textit{Visible Goal with Buildings}; \textit{Spot the Difference: Basic}; and the hardest task, \textit{Spot the Difference: Multi-Object}.

\paragraph{CPC vs. REC}
CPC was better than REC on all Spot the Difference tasks, and the two harder PsychLab tasks \textit{Change Detection} and \textit{What Then Where}. On the other two PsychLab tasks there was no difference between CPC and REC. However, REC was better on all Goal Navigation tasks except \textit{Invisible Goal Empty Arena}. When averaged out, REC was more useful when the working memory was FF, but CPC was more useful for an LSTM working memory.

%% file: 05_discussion.tex
\section{Discussion \& Future Work}

We constructed a diverse set of environments \footnote{Available at \href{https://github.com/deepmind/dm_memorytasks}{https://github.com/deepmind/dm\textunderscore memorytasks}.} to test memory-specific generalization, based on tasks designed to identify working memory and episodic memory in humans, and also developed an agent that demonstrates many of these cognitive abilities.
We propose both a testbed and benchmark for further work on agents with memory, and demonstrate how better understanding the memory and generalization abilities of reinforcement learning agents can point to new avenues of research to improve agent performance and data efficiency.
There is still room for improvement on the trickiest tasks in the suite where the agent fared relatively poorly. In particular, solving \textit{Spot the Difference: Motion} might need a generative model that enables forward planning to imagine how future motion unrolls (e.g., \citep{racaniere2017imagination}).
Our results indicate that adding an auxiliary loss such as CPC or reconstruction loss to an architecture that already has an external episodic memory improves generalization performance on holdout sets, sometimes synergistically. This suggests that existing agents that use episodic memory, such as DNC and NEC, could potentially boost performance by implementing an additional auxiliary unsupervised loss.

%% file: appendix.tex
\newpage
\appendix

\input{appendix_task_descriptions_and_results}

\input{appendix_scores}

\newpage

\section{Model}

\subsection{Importance Weighted Actor-Learner Architecture}
\label{subsec:impala}

We use the Importance Weighted Actor-Learner Architecture (IMPALA) \citep{dmlab_impala} in our work.  IMPALA uses an off-policy actor-critic approach where decoupled actors communicate experience to a learner.  The actor-to-learner relationship is many-to-one.  Each actor generates a batched trajectory, or episode, of experience and sends the state-action-reward traces ($s_0,a_0,r_0, ... , s_T,a_T,r_T$) to its respective learner.  The learner gathers trajectories from each actor and computes gradients to update the model parameters continuously.  As actors finish processing a trajectory they receive parameter updates from the learner then continue to generate trajectories.

Under this scheme the actors and learner policies fall out of sync between parameter updates.  The actor's \textit{behaviour policy}, $\mu$, is said to have \textit{policy lag} with respect to the \textit{target policy} of the learner, $\pi$.  To correct for this effect importance weighting with \textit{V-trace} targets are computed for each step:

\begin{equation}
    v_s \buildrel d\over = V(x_s) + \sum_{t=s}^{s+T-1} \gamma^{t-s} (\pi_{i=s}^{t-1}c_i)
    \rho_t  (r_t + \\ \gamma V(x_{t+1}) - V(x_t))
\end{equation}

where $\gamma \in [0,1)$ is a discount factor, $x_t$ and $r_t$ are the state reward at time-step $t$,  $\rho_t = min(\bar{\rho}, \frac{\pi(a_t|x_t)}{\mu(a_t|x_t)})$ and $c_i = min(\bar{c}, \frac{\pi(a_i|x_i)}{\mu(a_i|x_i)})$ are truncated importance sampling weights. These V-trace targets are used to compute gradients for the policy approximation in the learner.  This enables observations and parameters to each flow in a single direction, allowing for high data efficiency and resource allocation in comparison to other other approaches, such as asynchronous advantageous actor critic (A3C) \citep{a3c}.

\subsection{Residual Network Architecture}
\label{appendix:resnet}

To process the pixel input, the Memory Recall Agent and the other baselines reported in this work use a residual network \citep{resnet} with a similar architecture found in \citep{dmlab_impala}. This consists of three convolutional blocks with feature map counts of size 16, 32, and 32; each block has a convolutional layer with kernel size 3x3 followed max pooling with kernel size 3x3 and stride 2x2, followed by two residual subblocks. The output from the top residual block is followed by a 256-unit MLP to generate latent representations $x_{t}$ to be passed to the working memory and query network $f_{k}$.

\subsection{Contrastive Predictive Coding}
\label{appendix:cpc}

We use the encoder already present in the agent's architecture, the convolution neural network that takes the input frame ($i_{t}$) and converts it to the embedded visual input $x_t$. The auto-regressive component is the working memory itself, which takes $x_t$ as input and outputs $h_t$ which can be used to predict future steps in latent space: $x_{t+1}, \dots, x_{t+N}$, where $N$ said to be the number of CPC steps. Figure \ref{fig:cpc} illustrates the CPC approach (\citet{cpc}).

To introduce a noise-contrastive loss the mutual information (Eq. \ref{eq:cpc_mi}) between the target encoded representations $x_{t}$, and the contexts ($c_t$) -- which in our case are the memory states $h_t$.  For each sample, a positive real score is then generated via $f_{k}$, a log-bilinear density function (Eq. \ref{eq:cpc_lbi}) by taking the current output from the working memory $h_t$ and the latent vector of the $k^{th}$ step, $x_{t+k}$.

\begin{equation}
    \label{eq:cpc_mi}
    I(x, c) = \sum_{x, c}p(x, c)\log\frac{p(x|c)}{p(x)}
\end{equation}

\begin{equation}
    \label{eq:cpc_lbi}
    f_{k}(x_{t+k}, h_{t}) =  \exp\left(x_{t+k}^TW_{k}h_{t}\right)
\end{equation}

Given a sample trajectory of length $T$ and a fixed number of maximum CPC steps $N \leq T-1$, predictions are computed for each of the $k$-step predictive models ($1 \leq k \leq N$). For timestep $t$ ($1 \leq t \leq T-k$) and predictive model $k$ let $\upchi_{t, k}$ denote a set of samples from which a contrastive noise estimate is derived.  Each set $\upchi_{t, k}$ may be split into two subsets, a single positive sample and $T-k-1$ negative samples: $\upchi_{t, k}^+ = \{x_{t+k}\}$, $\upchi_{t, k}^- = \{x_{k+1}, ..., x_{t+k-1}, x_{t+k+1}, ..., x_{T-k}\}$ such that $\upchi_{t, k} = \upchi_{t, k}^+ + \upchi_{t, k}^-$ ($|\upchi_{t, k}| = T-k$). The noise contrastive loss is then determined by computing the categorical cross-entropy over the $(t, k)$-trajectory sample set $\forall (t, k)$.  Details can be seen in Eq. \ref{eq:cpc_sce_loss}.

\begin{equation}
    \label{eq:cpc_sce_loss}
    \mathcal{L}_{CPC(\upchi_{t, k})} =\mathbb{E}_{\upchi_{t, k}^+}\left[\log{\frac{f_{k}(x_{t+k}, h_{t})}{\sum_{i_{j} \epsilon \upchi_{t, k}} f_{k}(x_{j}, h_{t})}}\right] + \mathbb{E}_{\upchi_{t, k}^-}\left[\log{\left(1-\frac{f_{k}(x_{t+k}, h_{t})}{\sum_{i_{j} \epsilon \upchi_{t, k}} f_{k}(x_{j}, h_{t})}\right)}\right]
\end{equation}

\subsection{Reconstruction}
\label{appendix:recon}

Action and reward reconstructions are linear projections $f_{r}(r_{t}) = r_{t}W_{r} + b_{r}$ and $f_{a}(a_{t}) = a_{t}W_{a} + b_{a}$ while reconstructions of the image input $i_{t}$ are generated via the transpose residual network $f_{RN}^{T}$.  Sum of squared error losses are used for prior step reward and prior step actions while sigmoid cross-entropy is used for the image reconstruction. The losses are summed and scaled by a cost hyper-parameter for each to produce a full reconstruction loss for the model, $\mathcal{L}_{\text{REC}}$.  See equations  \ref{eq:rec_loss_r} to \ref{eq:rec_loss_tot} below for more details ($\sigma$ is the sigmoid function).

\begin{equation}
    \label{eq:rec_loss_r}
    \mathcal{L}_{\text{reward}} = \frac{\sum_{i=1}^{T}{\left(r_{t-1, i} - f(r_{t-1, i})\right)^{2}}}{2}
\end{equation}

\begin{equation}
    \label{eq:rec_loss_a}
    \mathcal{L}_{\text{action}} = \frac{\sum_{i=1}^{T}{\left(a_{t-1, i} - f(a_{t-1, i})\right)^{2}}}{2}
\end{equation}

\begin{equation}
   \label{eq:rec_loss_i}
   \mathcal{L}_{\text{image}} = -i_{t}\log(\sigma(f_{RN}^{T}(h_t))) - \left(1 - i_{t}\right)\log\left(1 - \sigma(f_{RN}^{T}(h_t))\right)
\end{equation}

\begin{equation}
   \label{eq:rec_loss_tot}
   \mathcal{L}_{\text{REC}} = c_{\text{image}} \mathcal{L}_{\text{image}} + c_{\text{action}} \mathcal{L}_{\text{action}} + c_{\text{reward}} \mathcal{L}_{\text{reward}}
\end{equation}

In our experiments we set $c_{\text{image}}$=$c_{\text{action}}$=$c_{\text{reward}}$=1.0 for all tasks, except in AVM, Continuous Recognition and Change Detection, where $c_{\text{image}}=30, 1.5, $ and 3, respectively. We did not tune for this hyper-parameter, we used first guess or previous work (such as \citep{merlin}) for choosing it.  
\input{hypers}

%% file: appendix_task_descriptions_and_results.tex
\section{Level descriptions and further experimental findings}
\label{appendix:level_config}

As described in Section \ref{scale_stimulus_split}, for each task in the Suite we construct a small training level, a large training level, a `holdout-interpolation' level and a `holdout-extrapolation' level. 

During training the environment uniformly samples from the small and large training levels. The interpolation level has a scale somewhere in between `small' and `large' while the extrapolation level corresponds to `extra-large' (Table \ref{tab:overall_structure}). A summary of the alterations made for each task split is in Table \ref{tab:alterations}. The settings used in each level per task are described below.

\begin{table}[H]
\caption[]{\textbf{Overall structure for scale and stimulus split.}}
\label{tab:overall_structure}%
  \centering
    \begin{tabular}{l|c|c}
    \toprule
    \diagbox{Scale}{Stimuli} & Training set & Holdout set\\
    \midrule
    Small & Used for training & ---  \\
    \hline
    Medium & --- & Used for interpolation \\
    \hline
    Large & Used for training & --- \\
    \hline
    Extra-large & --- & Used for extrapolation \\
    \bottomrule
    \end{tabular}%
\end{table}%
The dashed (`---') settings in Table \ref{tab:overall_structure} are not reported nor used, since they lack a clear interpretation in terms of generalization.

\begin{table}[H]
    \caption{\textbf{Scale and stimulus alterations across task families}}
    \label{tab:alterations}
    \centering
    \begin{tabular}{l|c|c}
    \hline
    Task & Scale & Stimulus \\ 
    \hline
    AVM & Number of trials & Image \\ 
    Continuous Recognition & Number of trials & Image \\ 
    Change Detection & Delay study/test & Color \\ 
    What Then Where & Delay study/query & Digit image \\ 
    \hline
    Spot Diff Basic & Corridor delay & Color \\ 
    Spot Diff Passive & Corridor delay duration & Color \\ 
    Spot Diff Multi-object & Number of objects & Color \\ 
    Spot Diff Motion & Corridor delay & Motion pattern \\
    \hline
    All Goal Navigation tasks & Arena size & Goal spawn \\ 
    \hline
    Transitive Inference & Length of transitive chain & Object color \\ 
    \bottomrule
    \end{tabular}
\end{table}

\subsection{PsychLab}
\label{appendix:psychlab}

\begin{figure}[!htb]
    \centering
    \subfigure[Arbitrary Visuomotor Mapping (AVM)]{
        \includegraphics[width=0.48\linewidth,height=0.25\textwidth]{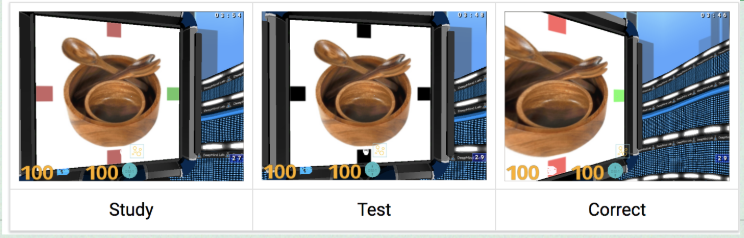}
        \label{fig:screenshot_avm}}
    \hfill
    \subfigure[Continuous Recognition]{
        \includegraphics[width=0.48\linewidth,height=0.25\textwidth]{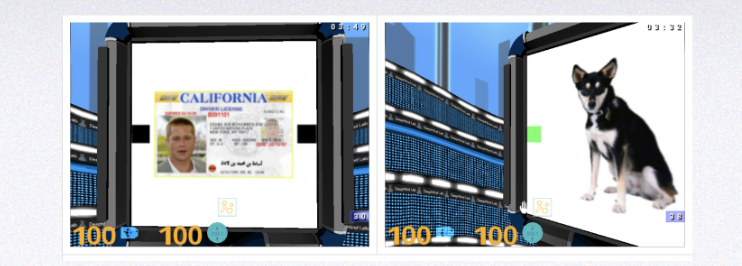}
        \label{fig:screenshot_continuous_recognition}}%
        
  \subfigure[Change Detection]{
         \includegraphics[width=0.48\linewidth,height=0.2\textwidth]{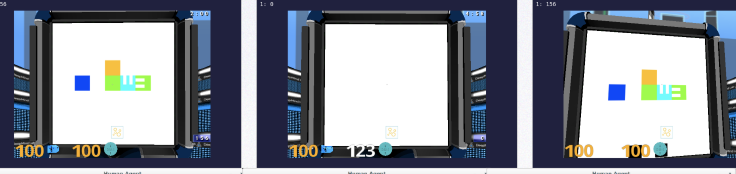}
        \label{fig:screenshot_change_detection}}
    \hfill
    \subfigure[What Then Where]{
        \includegraphics[width=0.48\linewidth,height=0.2\textwidth]{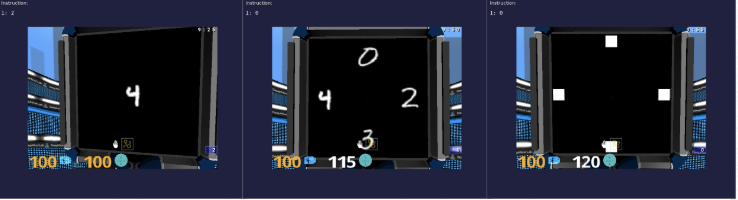}
        \label{fig:screenshot_wtw}}%
    \caption{All PsychLab tasks have multiple trials within an episode. Each trial consists of a single image being displayed on the panel. In (a), when the agent sees an image for the first time, the associated direction is indicated on the screen (green box on the left). By executing the indicated pattern, the agent receives a reward. When the agent is presented with an image it has already seen during the episode, the associated direction is no longer indicated (middle), and the agent must remember it from its previous experience in order to get a reward (right). In (b), the agent is shown a pattern (left), and after a delay (middle), a second pattern is shown (right). The agent has to indicate if there was a change between the two patterns or not  by looking right or left, respectively. The delay period separating the two patterns varies in length. In (c), the agent indicates if it has seen the image in the current episode by looking left or right, respectively. In (d), in the `what' study phase, an MNIST digit is displayed (left). In the `where' study period, four distinct MNIST digits are displayed including the one from the `what' period (middle). In the test phase (right), the agent must remember what digit was displayed in the `what' period, see where it is located during the study where period, and then respond by looking to that location. In this example it has to look left.}
    \label{fig:psychlab}
\end{figure}

Our Memory Tasks Suite has four PsychLab tasks: Arbitrary Visuomotor Mapping (AVM), Continuous Recognition, Change Detection and What Then Where. The description of each task is found in Figure \ref{fig:psychlab}. Videos with agent play and the \href{https://sites.google.com/view/memory-tasks-suite}{https://sites.google.com/view/memory-tasks-suite}.

\paragraph{Scale} Either number of trials per episode or delay duration.

For \textit{Arbitrary Visuomotor Mapping} and \textit{Continuous Recognition}, every episode lasts at most 300 seconds, except for the Extrapolate level where the cap is set to 450 seconds to accommodate the larger number of trials. In \textit{Change Detection} an episode lasts at most 300 seconds, while for \textit{What Then Where} it is 600 seconds.
\begin{table}[H]
    \centering
    \begin{tabular}{l|ccc}
    \hline
    \diagbox{Scale}{Task} &  \shortstack{AVM and Cont. Recog.:\\Trials per episode} & \shortstack{Change Detection:\\delay (seconds)} & \shortstack{What Then Where:\\delay (seconds)}\\
    \hline 
    Small & 50  & 2, 4, 8   & 4, 8\\ 
    Interpolate & 40  & 16, 32 & 16, 64 \\ 
    Large & 50 & 64, 128 & 32, 128\\
    Extrapolate & 75 & 130, 150, 200, 250 & 132, 156, 200, 256 \\ 
    \bottomrule
    \end{tabular}
\end{table}

\paragraph{Stimulus} Either color set or image set.

\begin{table}[H]
    \centering
    \begin{tabular}{l|c|c|c}
    \hline
    Task & AVM and Cont. Recog & Change Detection & What Then Where \\
    \hline
    Stimulus & Different images &  Color set & MNIST digits \\
    \hline
    Training & Images with even ID & Amethyst, Caramel, & 0, 1, 2, 3, 4 \\
    & & Honeydew, Jade, Mallow &  \\
    \hline
    Holdout & Images with odd ID & Yellow, Lime, Pink, Sky, Violet &  5, 6, 7, 8, 9\\
    \bottomrule
    \end{tabular}
\end{table}

\subsubsection{PsychLab: main experimental findings}

\paragraph{AVM:} in this task, the agent must remember associations between images and specific movement patterns (Figure \ref{fig:psychlab} (a)). 

The most useful component turned out to be MEM. This is in line with earlier findings that an external episodic memory is a prerequisite for solving AVM \citep{merlin}. Adding an auxiliary loss helped when the controller was FF but made no difference for an LSTM. Also, choosing between CPC or REC for auxiliary unsupervised loss did not make a major difference for either controller.

\paragraph{Continuous Recognition:} in this task, the agent must remember if it has seen a particular image before by looking left or right (Figure \ref{fig:psychlab} (b)).

MEM was the most useful component when added to an LSTM, but made no difference when added alone to an FF controller. However, adding a stack of MEM plus either CPC or REC provided a substantial performance boost for both FF and LSTM. 

\paragraph{Change Detection:} in this task, agent sees two images separated by a delay and has to correctly indicate if the two images are different (Figure \ref{fig:psychlab} (c)).

CPC brought the largest benefit. Interestingly the addition of MEM to the FF baseline actually hurt performance slightly, and made no difference for LSTM.

\paragraph{What Then Where:} this task consists of a `what' and `where' study phase, followed by a test phase where the agent must remember what image was displayed and where it was located (Figure \ref{fig:psychlab} (d)). 

This was the trickiest task in the Psychlab family. This task was an outlier in the sense that unlike any other task in the suite, the LSTM baseline beat all other models. The worst additional component was REC which dragged down performance to below random.

\subsection{Spot the Difference (SD)}

The tasks were built in Unity, and each episode lasts 120 seconds except for \textit{Spot the Difference: Motion} which has a 240-second timeout.

\paragraph{Scale} Either corridor delay duration or number of objects in room. 

In Spot the Difference Multi-Object, Room 2 has the exact same number of objects as Room 1.

\begin{table}[H]
    \centering
    \begin{tabular}{l|ccc}
    \hline
    \diagbox{Scale}{Task} &  \shortstack{SD Basic, Passive and Motion:\\ Corridor delay (seconds)} & \shortstack{SD Multi-Object:\\ Number of objects in Room 1} \\
    \hline 
    Small & 0 & 2 or 3 \\ 
    Interpolate & 5 & 4 \\ 
    Large & 10 & 5 or 6\\
    Extrapolate & 15 & 7 \\ 
    \bottomrule
    \end{tabular}
\end{table}

\paragraph{Stimulus} Either color set or motion pattern set.

\begin{table}[H]
    \centering
    \begin{tabular}{l|c|c}
    \hline
    Task & SD Basic, Passive and Motion & SD Multi-Object \\
     \hline
    Stimulus & Color Set  & Motion Pattern Set \\
    \hline
    Training & Red, Green, Blue, & Circle, Square, Five-point star, Hexagon\\
    & White, Slate & Linear along X-axis, Linear along Y = X diagonal\\
    \hline
    Holdout & Yellow, Brown, Pink, & No motion, Triangle, Pentagon, Figure-eight\\
    & Orange, Purple &  Linear along Y-axis, Linear along Y = -X diagonal\\
    \bottomrule
    \end{tabular}
\end{table}

\subsubsection{Spot the Diff: main experimental findings}

Every task in this family consists of two rooms connected by a short corridor. There is a set of gates in the middle of the corridor that can trap the agent there for a configurable delay duration.

\paragraph{Basic} In the basic Spot the Difference task, where the agent is not forced to see any of the blocks in Room 1 before it goes to the next room, adding MEM alone to the controller had minimal effect, and using REC with MEM also did not make much difference. Adding CPC to an LSTM helped performance but it turned out that using the combination of MEM + CPC provided the biggest gain and was synergistic.

\paragraph{Passive} In this task the agent is guaranteed to see the two blocks in the first room before it enters the second room. Adding MEM alone to the controller made the biggest positive difference, which makes sense since that hypothetically would make it possible for the agent to solve the task by remembering a single snapshot. CPC helped when added to FF together with MEM, but hurt when added to LSTM alone. REC helped performance when added to FF + MEM, but not as much as CPC did in that case, and actually hurt performance when added to LSTM + MEM.

\paragraph{Motion} Nothing did well on train or holdout sets, and curves took longer to take off in general. 
This is likely due to the highly challenging nature of the task, which requires the agent to memorize 3D motion patterns traced out over some time period by multiple objects and then compare motion patterns against each other.
Results would potentially be improved by hyperparameter tuning or further improvements to agent architecture. 

\paragraph{Multi-object} This was the hardest task in the family, and nothing did well here either. This could be due to there being a variable number of objects in each room, rather than always exactly two objects per room. When added by itself to a controller MEM either had no effect or hurt performance. The combined synergistic stack of MEM + CPC was the most useful addition on this task when the working memory was LSTM. That said, no models fared well on Holdout-Interpolate and Holdout-Extrapolate for this task.

\subsection{Navigate to Goal}

These tasks are in Unity and have an episode timeout of 200 seconds, except \textit{Visible Goal Procedural Maze} which is a modification of DMLab's \textit{Explore Goal Locations} task and has episodes lasting 120 seconds each. 

\paragraph{Scale} Size of square arena, in terms of in-game metric units. 

\begin{table}[H]
    \centering
    \begin{tabular}{l|cc}
     \diagbox{Scale}{Task} &  \shortstack{Visible Goal Procedural Maze: \\ Arena Size} & \shortstack{All but Visible Goal Procedural Maze: \\ Arena Size} \\
    \hline 
    Small & 11 $\times$ 11 & 10 $\times$ 10\\ 
    Interpolate & 15 $\times$ 15 & 15 $\times$ 15 \\ 
    Large & 21 $\times$ 21 & 20 $\times$ 20\\
    Extrapolate & 27 $\times$ 27 & 25 $\times$ 25\\ 
    \bottomrule
    \end{tabular}
\end{table}

\paragraph{Stimulus} Goal spawn region.
\begin{table}[H]
    \centering
    \begin{tabular}{l|cc}
    \hline
    \diagbox{Stimulus}{Task} &  \shortstack{Visible Goal Procedural Maze: \\ Goal spawn region} & \shortstack{All but Visible Goal Procedural Maze: \\ Goal spawn region} \\
    \hline
    Training & North half & Northwest and southeast quadrants \\
    Holdout & South half & The other two quadrants \\
    \bottomrule
    \end{tabular}
\end{table}

\subsubsection{Navigate to Goal: main experimental findings}

Using an auxiliary unsupervised reconstruction loss to learn high-quality representations turned out to be the most useful component for this task family. 

\begin{figure}[H]
    \centering
    \subfigure[Routes taken by MRA agent in one episode]{
        \includegraphics[width=0.9\linewidth]{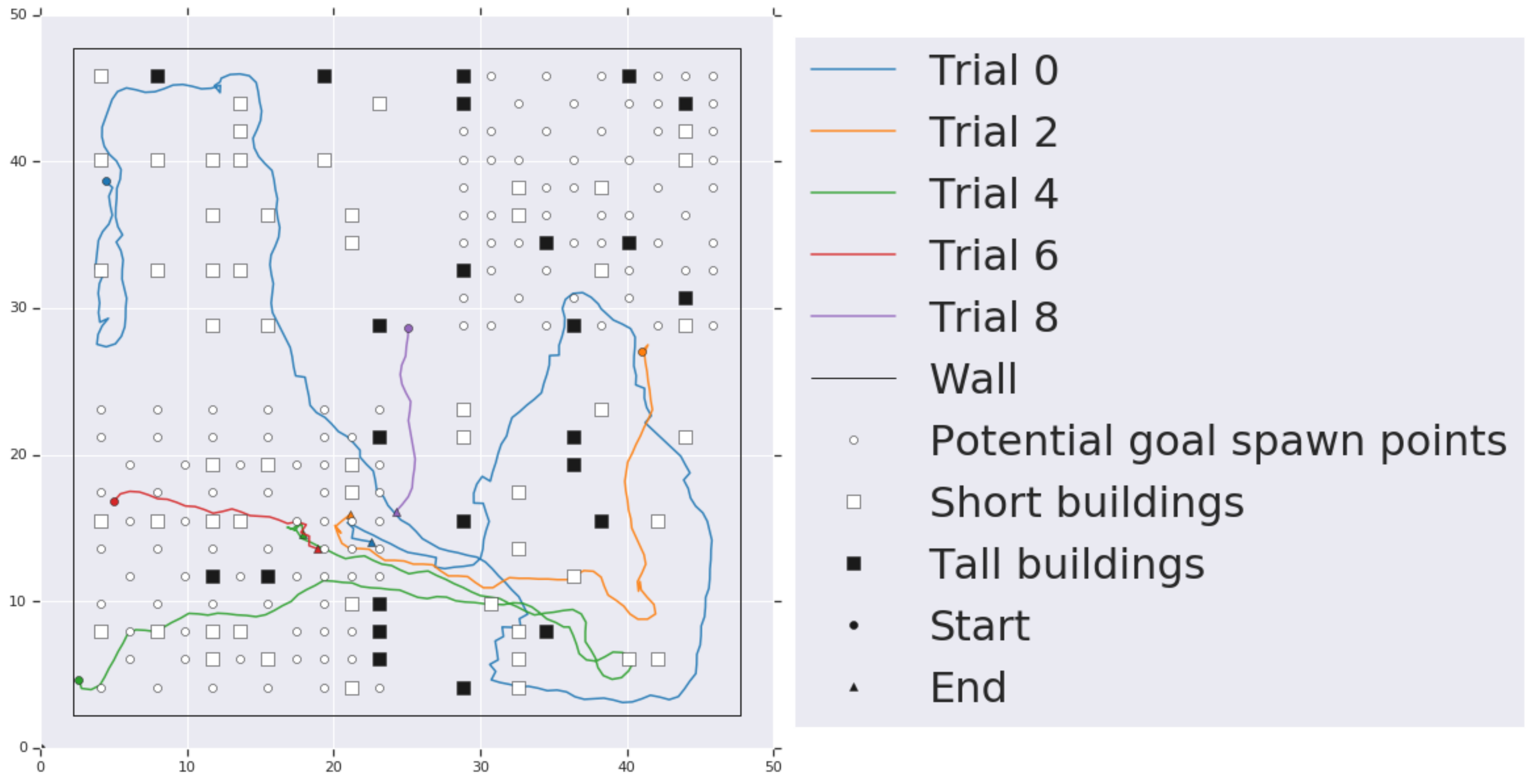}
        \label{diag:watermaze_trajectory}}
    \hfill
    \subfigure[Timesteps taken to reach goal]{
        \includegraphics[width= 0.8\linewidth]{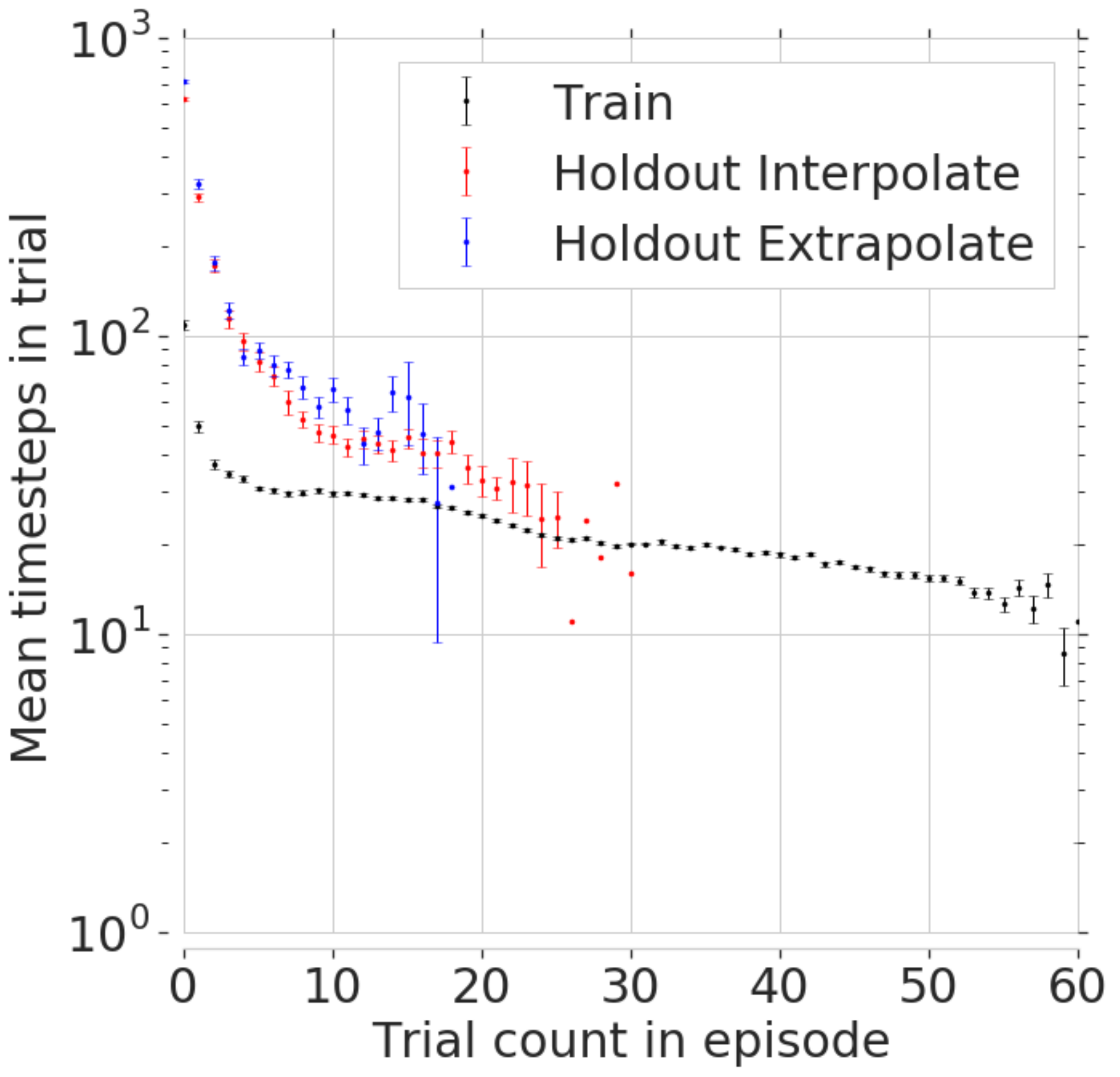}
        \label{diag:watermaze_ttg}}%
    \caption{Trajectories and time-to-goal for \textit{Invisible Goal with Buildings}. In (a), our MRA (LSTM + MEM + CPC) agent learns to take increasingly shorter routes to the goal. Note: The end-points of each trial trajectory appear to be in slightly different locations. This is because the goal is on a map tile rather than a single coordinate, and also due to manual adjustments we made to account for the agent avatar in Unity continuing to move for a small number of frames immediately after reaching the goal but before it is respawned. In (b), the number of time-steps taken per trial is plotted for Train, Holdout-Interpolate, Holdout-Extrapolate, along with standard error bars. Note: Some points at the rightmost end of each curve will have no error bar if there was only one data point.}
\end{figure}

We also observed that in successful models such as LSTM + MEM + CPC, which is the MRA architecture, the agent is able to do better than simply memorizing a route to the invisible goal. Rather, it learns the location of the goal, and the time it takes to reach the goal location grows shorter every time it respawns within an episode (see example trajectory in Fig \ref{diag:watermaze_trajectory} and time-to-goal plot in Fig \ref{diag:watermaze_ttg}).

\paragraph{Visible Goal Procedural Maze} Using REC with LSTM + MEM performed the best here, and FF + MEM + REC was the next best. The MEM + CPC stack was a distant runner-up compared with the MEM + REC stack for both controllers.

\paragraph{Visible Goal With Buildings} Like in the other Visible Goal task, LSTM + MEM + REC was the most successful model. MEM was slightly more helpful than CPC when used in conjunction with an LSTM (we did not have bandwidth to run the FF + CPC ablation). MEM + CPC also had a synergistic effect when stacked with an LSTM.

\paragraph{Invisible Goal With Buildings} Adding MEM + REC was the most useful, for both FF and LSTM.

\paragraph{Invisible Goal Empty Arena} This task can be expected to be the most difficult in the family due to the relative sparsity of visual spatial cues. Adding MEM alone to a controller always helped slightly. REC helped more than CPC did when used with an FF controller but for an LSTM controller CPC had a slight edge. 
\subsection{Transitive Inference}
\label{appendix:ti}

The task was built in Unity and has an episode timeout of 200 seconds.

\textbf{Scale:} Number of objects in transitive chain. \textbf{Stimulus:} Color set.

\begin{tabular}{cc}
 \centering
    \begin{minipage}{.45\linewidth}
        \begin{tabular}{lc}
            \hline
            Scale & Transitive chain length \\ 
            \hline
            Small & 5 \\ 
            Interpolate & 6 \\
            Large & 7 \\
            Extrapolate & 8 \\
            \bottomrule
        \end{tabular}
    \end{minipage}
\begin{minipage}{.5\linewidth} 
    \begin{tabular}{lc}
        \hline
        Stimulus & Color set \\
        \hline
        Training & Red, Green, Blue, White, Black, \\
         & Pink, Orange, Purple, Grey, Tan \\
        \hline
        Holdout & Slate, Yellow, Brown, Lime, Magenta \\ 
         & Mint, Navy, Olive, Teal, Turquoise \\
        \bottomrule
    \end{tabular}
\end{minipage}
\end{tabular}

\subsubsection{Transitive Inference: main experimental findings}

Transitive inference is a form of reasoning where one infers a relation between items that have not been explicitly directly compared to each other. 
In humans, performance on probe pairs and anchor pairs with symbolic distance of greater than one excluding anchor objects tends to correlate with awareness of the implied hierarchy \citep{Smith10138}. 

As an illustrative example: Given a `transitive chain' of five objects \textit{A, B, C, D, E} where we assume \textit{A} is the lowest-valued object and \textit{E} the highest, we begin with a demonstration phase in which we present the agent with pairs of adjacent objects \textit{<A, B>}, \textit{<B, C>}, \textit{<C, D>}, \textit{<D, E>} . 

In this demo phase we scramble the order in which the pairs are presented and also scramble the objects in the pair such that an agent may see \textit{<D, C>} followed by \textit{<A, B>}, etc. The pairs are presented one at a time, and the agent needs to correctly identify the higher-valued object in the current pair in order to proceed to seeing the next pair. 

Once the demo phase is completed, we show the agent a single, possibly-scrambled challenge pair. This challenge pair always consists of the object second from the left and the object second from the right in the transitive chain, in this case \textit{<B, D>}. The agent's task is again to go to the higher-valued object.

In our results, we found that stacking MEM with auxiliary loss was crucial. For an FF controller CPC was more useful than REC, but for LSTM it was the other way round. Also, although both LSTM + MEM + CPC and LSTM + MEM + REC achieved normalized scores that were not too far apart, REC was more data-efficient and took off earlier than the former. We observed a synergistic effect when combining MEM with CPC for an LSTM, but that was still outdone by using MEM + REC.

\subsection{Jumpy Backpropagation (JB) ablation}

\begin{figure}[H]
    \centering
    \subfigure{
        \includegraphics[width=0.9\textwidth]{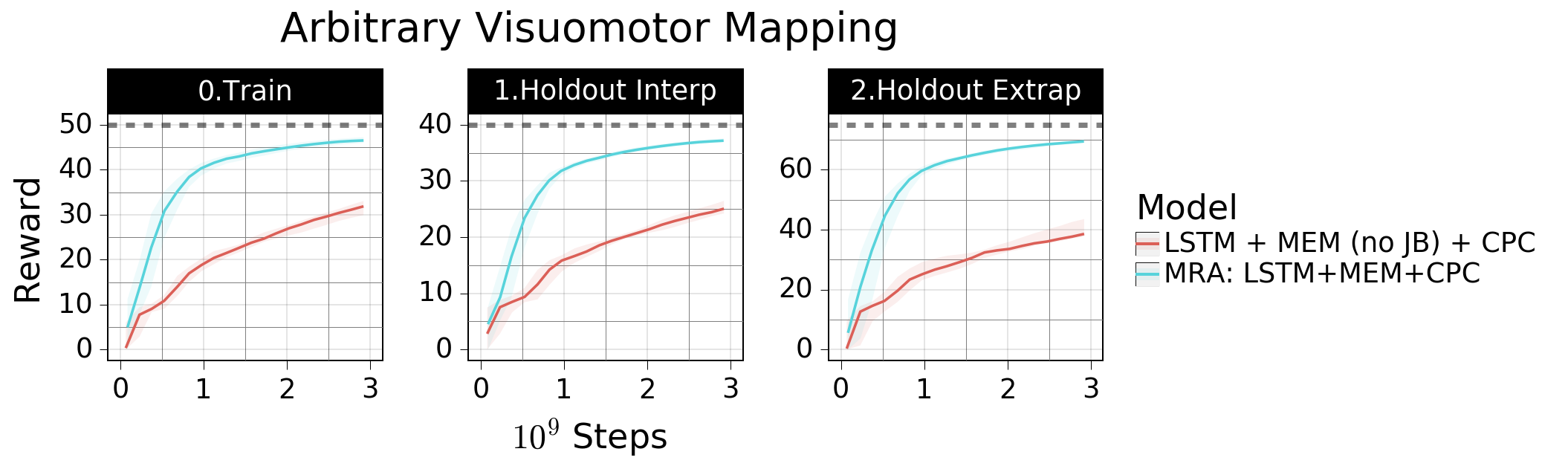}
    }%
    \vspace*{-0.4cm}
    \subfigure{
        \includegraphics[width=0.9\textwidth]{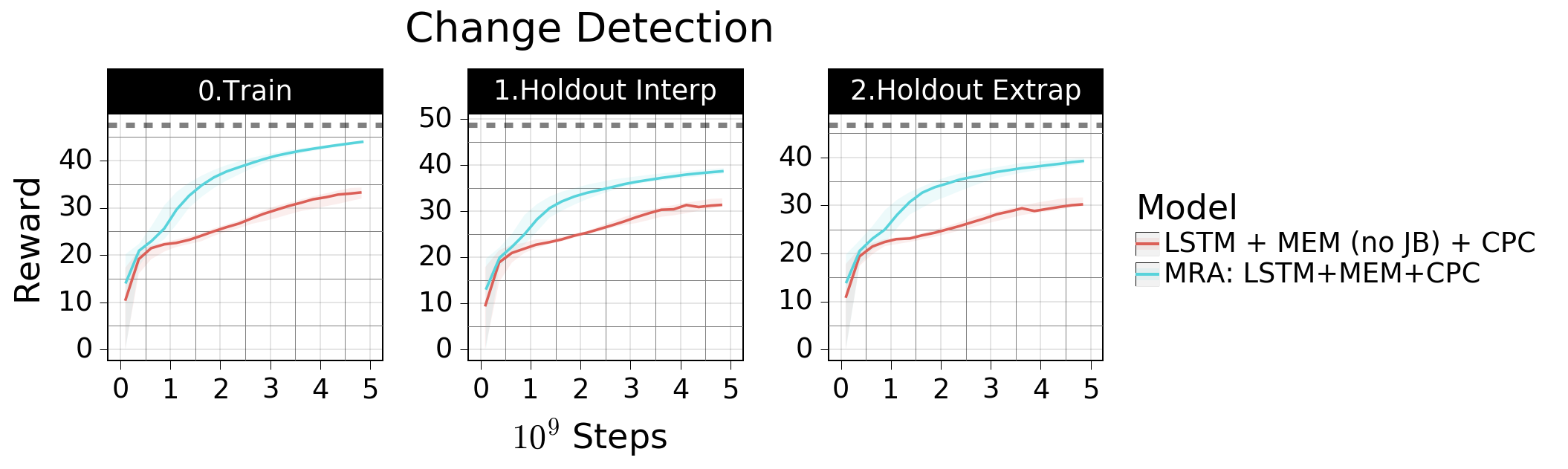}
        }%
   \vspace*{-0.5cm} 
    \subfigure{
        \includegraphics[width=0.9\textwidth]{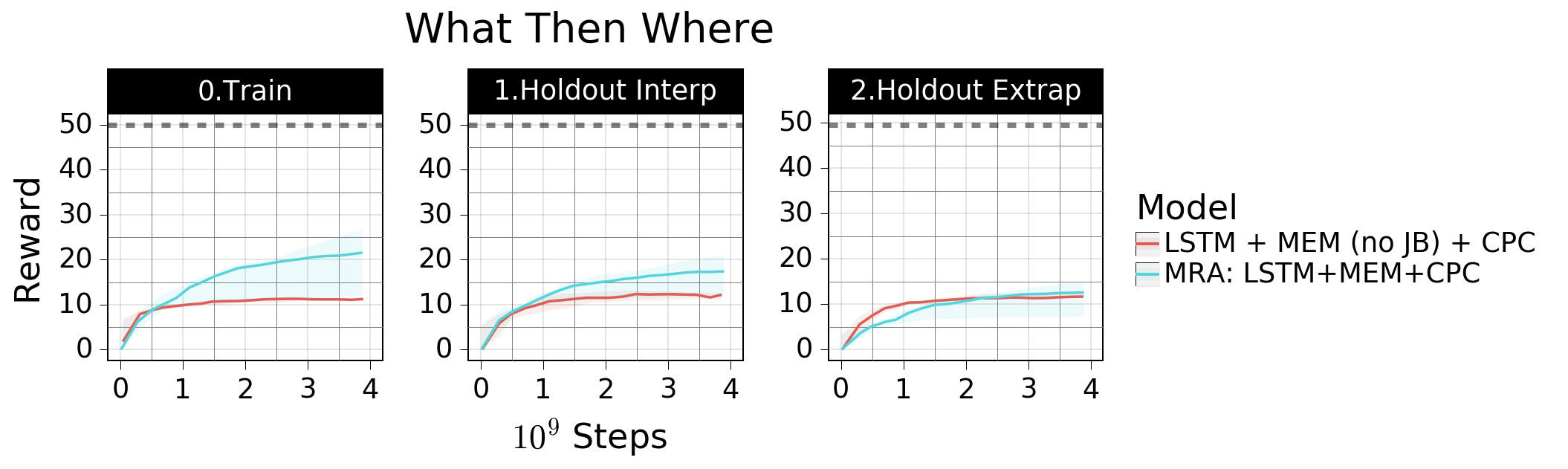}
        }%
  \vspace*{-0.4cm}        
  \subfigure{
         \includegraphics[width=0.9\textwidth]{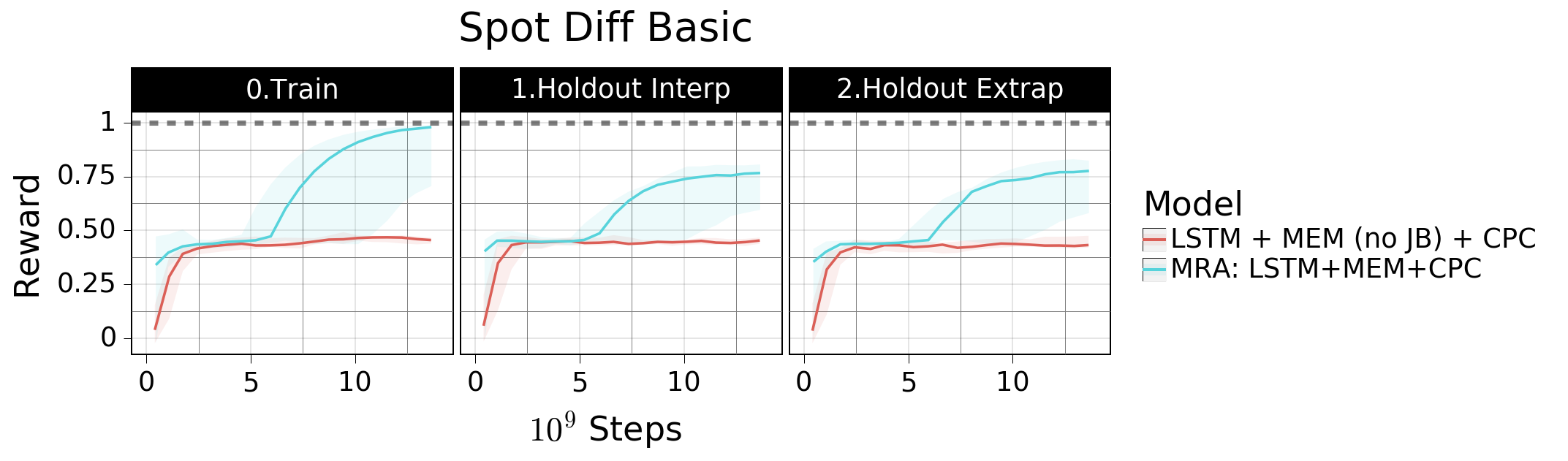}
      }%
    \vspace*{-0.4cm}  
    \subfigure{
        \includegraphics[width=0.9\textwidth]{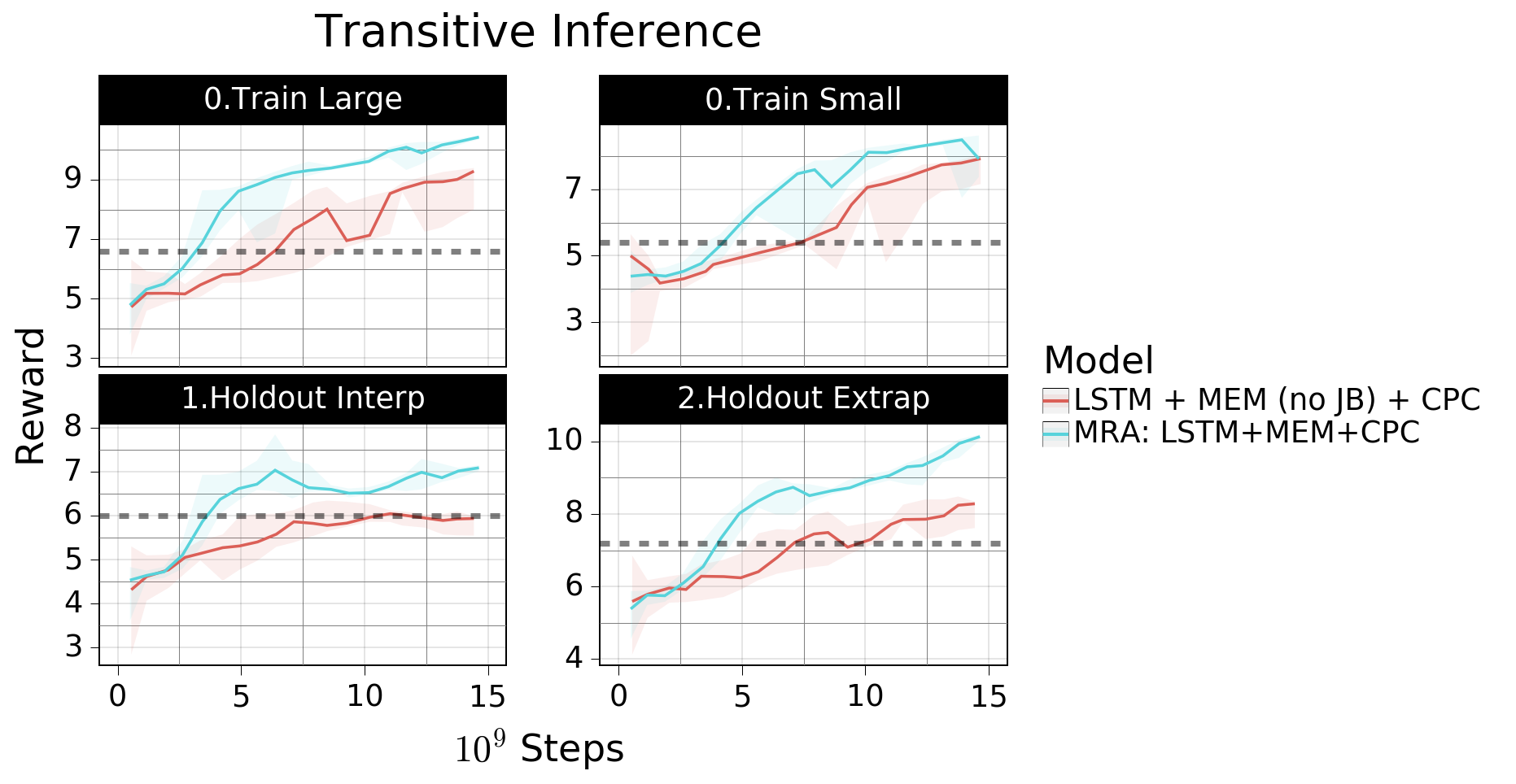}
        }%
        
    \caption{Comparison between MRA (LSTM + MEM +CPC) and its version without the jumpy backpropagation feature on MEM: LSTM + MEM (no JB) + CPC. Here we show the tasks where JB yields improvements on performance both at training time and on the holdout test levels. The dotted lines indicate human baseline scores for each task.}
    \label{fig:JB_ablation_curves_better}
\end{figure}

\begin{figure}[H]
    \centering
    \subfigure{
        \includegraphics[width=\textwidth]{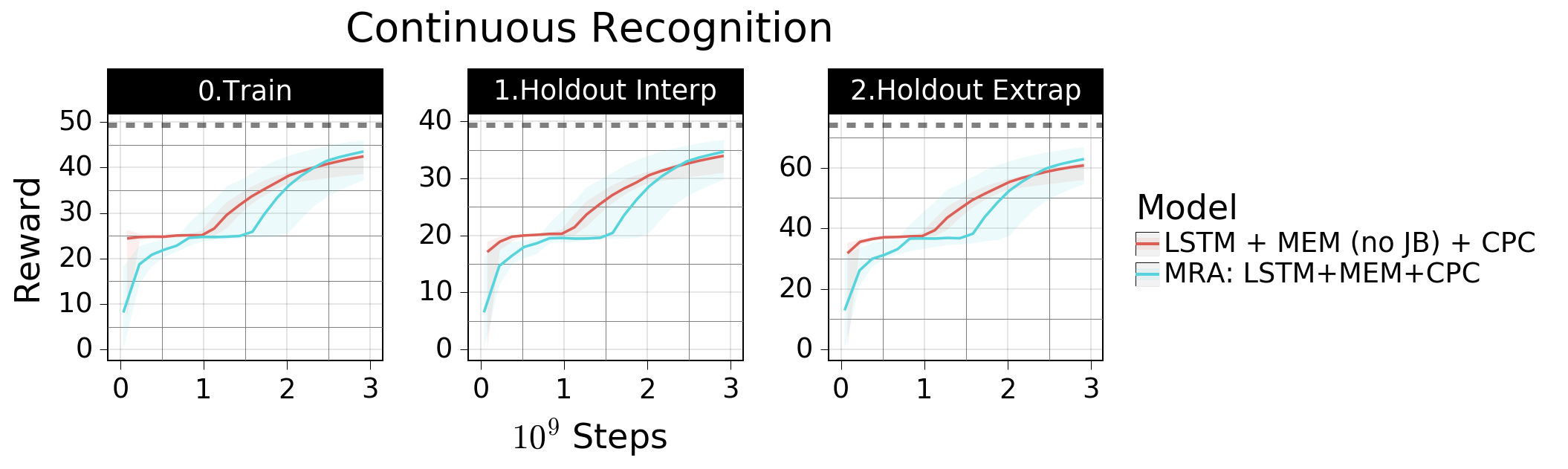}
    }%
  \vspace*{-0.4cm}      
  \subfigure{
         \includegraphics[width=\textwidth]{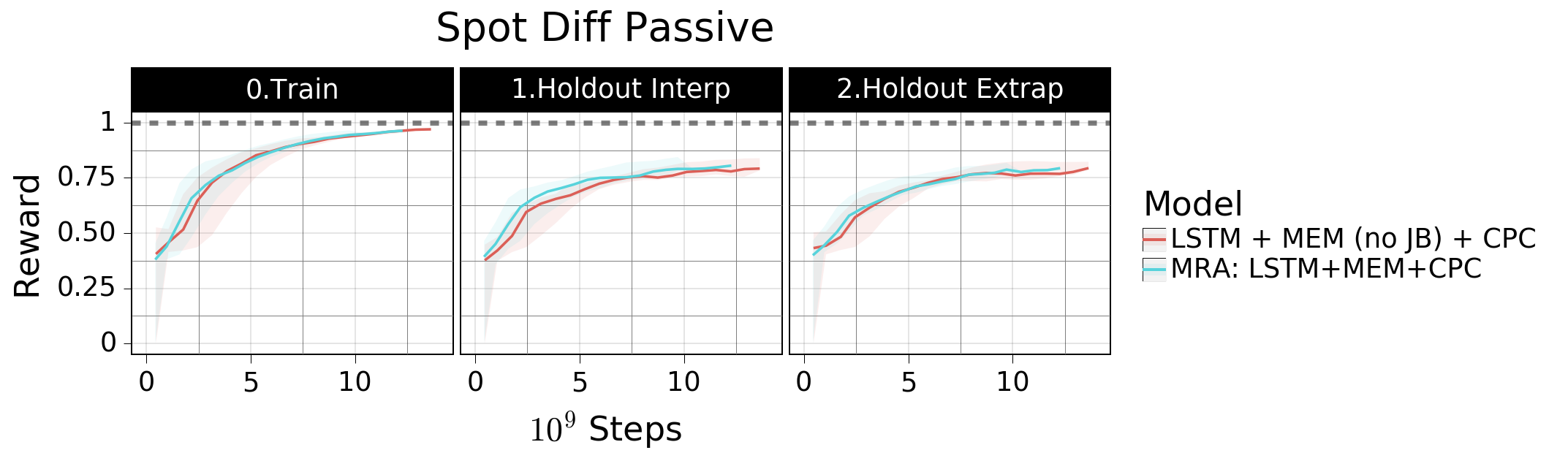}
      }%
    \vspace*{-0.4cm}  
    \subfigure{
        \includegraphics[width=\textwidth]{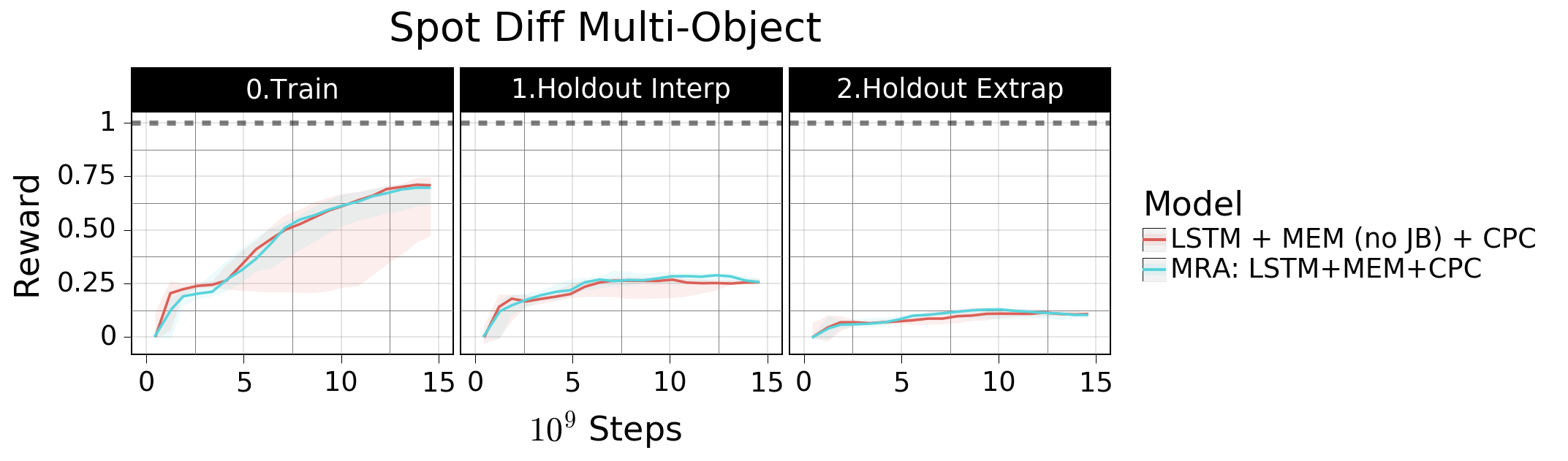}
        }%
    \vspace*{-0.4cm}  
    \subfigure{
        \includegraphics[width=\textwidth]{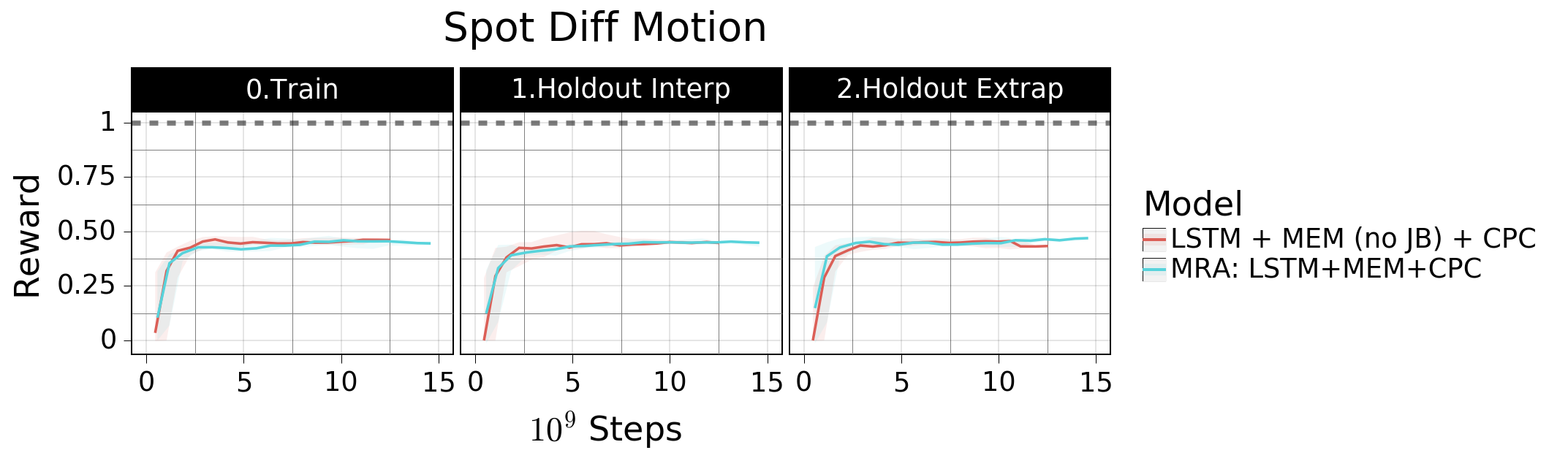}
        }%
    \vspace*{-0.4cm}     
    \subfigure{
        \includegraphics[width=\textwidth]{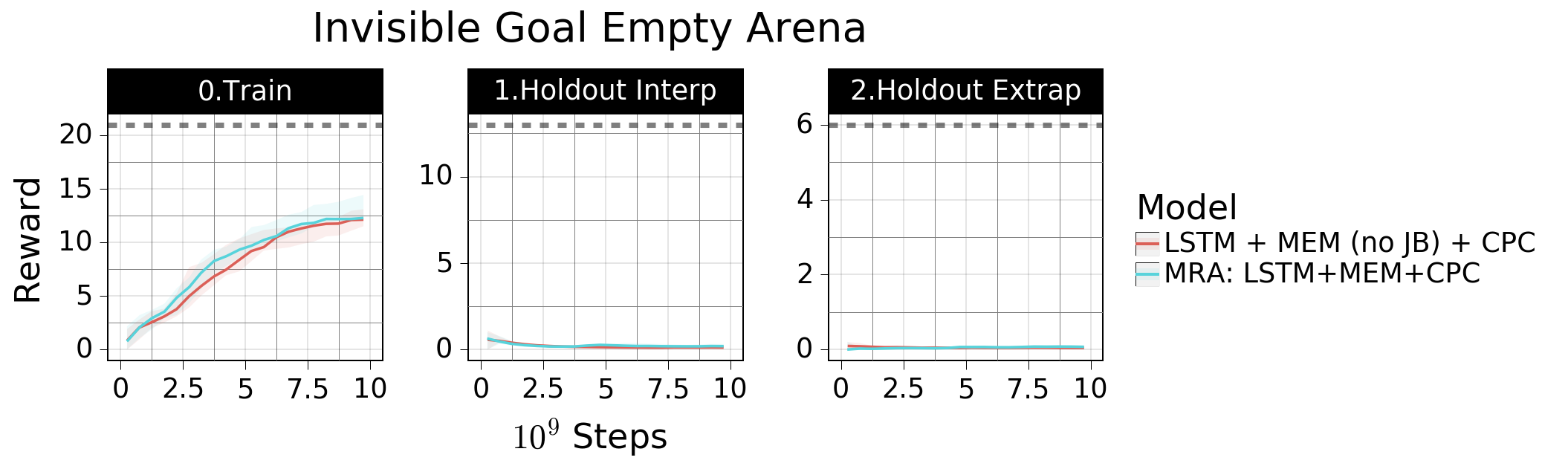}
        }%
    \caption{[1/2] Comparison between MRA and its version without the jumpy backpropagation (JB) feature. Here we show the tasks where JB makes little difference on performance. The dotted lines indicate human baseline scores for each task.}   
    \label{fig:JB_ablation_curves_same1}
\end{figure}

\begin{figure}[H]
    \centering
    \subfigure{
        \includegraphics[width=\textwidth]{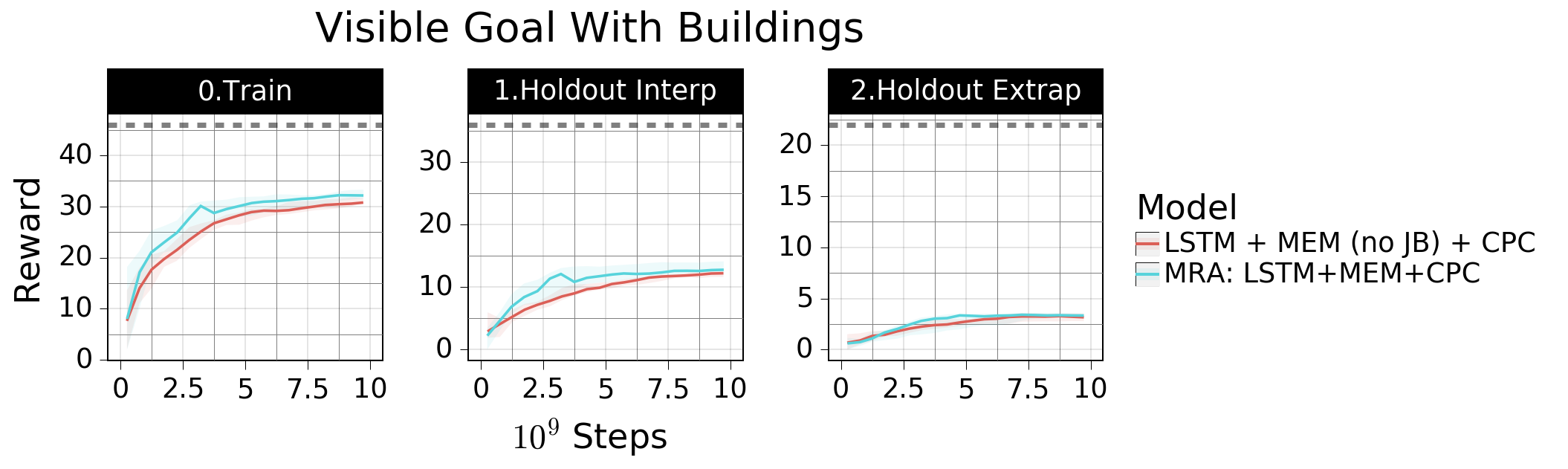}
        }%
    \vspace*{-0.4cm}  
    \subfigure{
        \includegraphics[width=\textwidth]{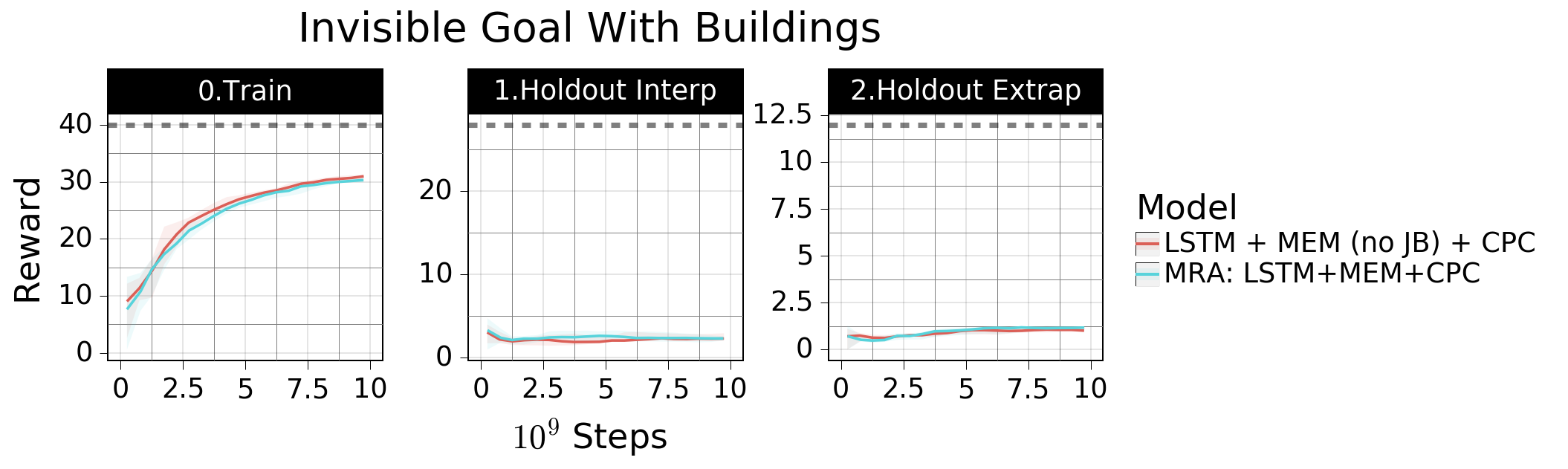}
        }%
        \vspace*{-0.4cm}  
    \subfigure{
        \includegraphics[width=\textwidth]{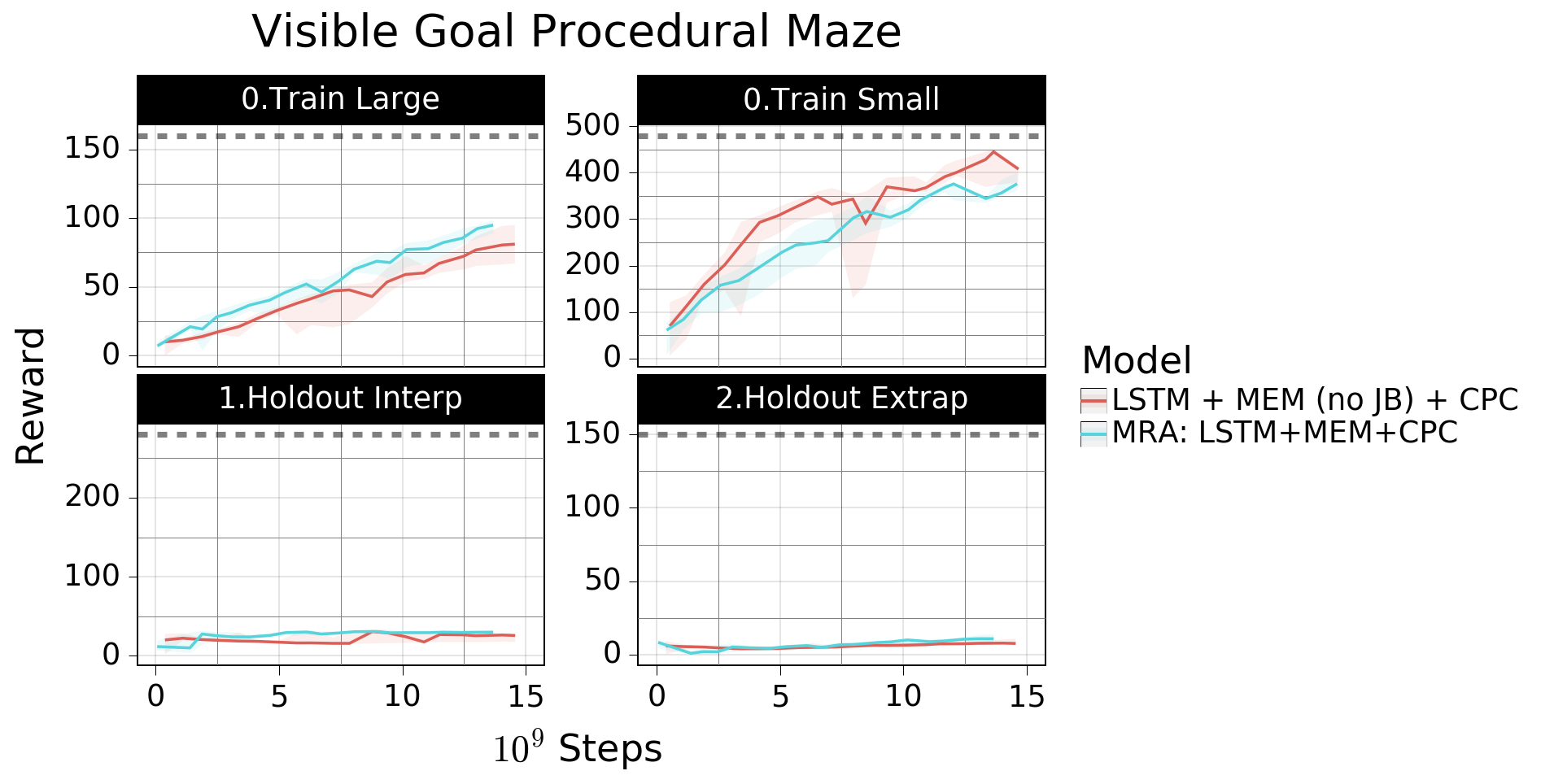}
        }%    
    \caption{[2/2] Comparison between MRA and its version without the jumpy backpropagation (JB) feature. Here we show the tasks where JB makes little difference on performance. The dotted lines indicate human baseline scores for each task.}
    \label{fig:JB_ablation_curves_same2}
\end{figure}

We studied the impact of having Jumpy Backpropagation (JB) as described in Section \ref{sec:model}.  In Fig \ref{fig:JB_ablation_curves_better}, we can see the set of tasks where adding the JB yields improvements on performance both at training time and on the holdout test levels. Figures  \ref{fig:JB_ablation_curves_same1} and \ref{fig:JB_ablation_curves_same2} show the performance on the remaining levels from the Memory Task Suite, where having the JB feature did not hurt performance. We conclude that JB is an important component of the MRA architecture.

\subsection{Agent Performance Curves}
In this session we show training and test curves for all models in all tasks. The dotted lines indicate human baseline scores for each task.

\begin{figure}[H]
    \centering
    \subfigure{
        \includegraphics[width=\textwidth]{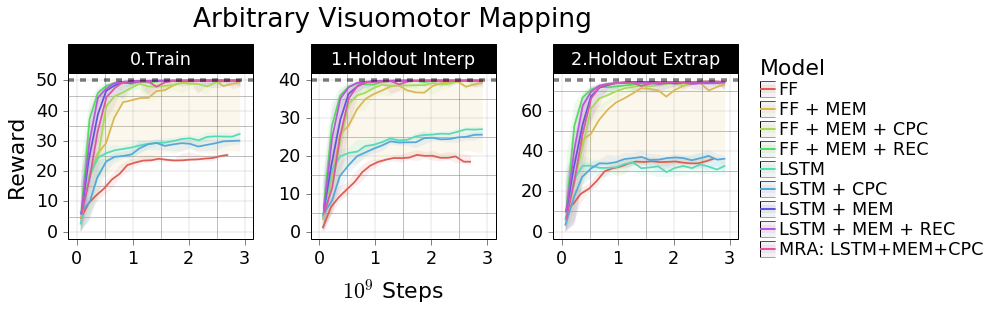}
        }%
    \vspace*{-0.4cm}     
    \subfigure{
        \includegraphics[width=\textwidth]{figures/training_curves/continuous_recognition.png}
        }%
  \vspace*{-0.4cm}       
  \subfigure{
        \includegraphics[width=\textwidth]{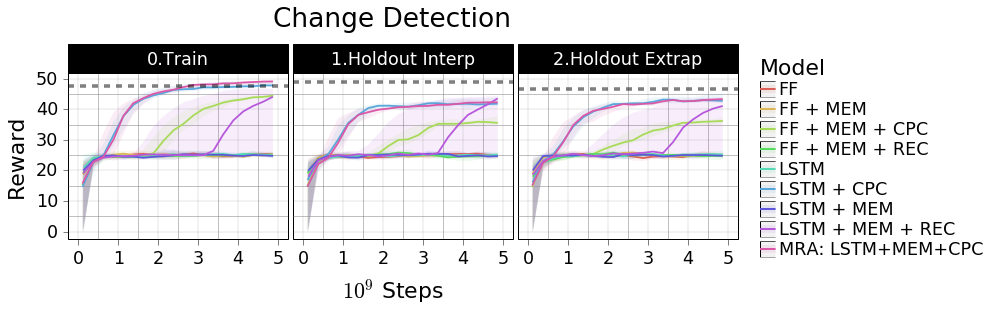}
        }%  
   \vspace*{-0.4cm}      
    \subfigure{
        \includegraphics[width=\textwidth]{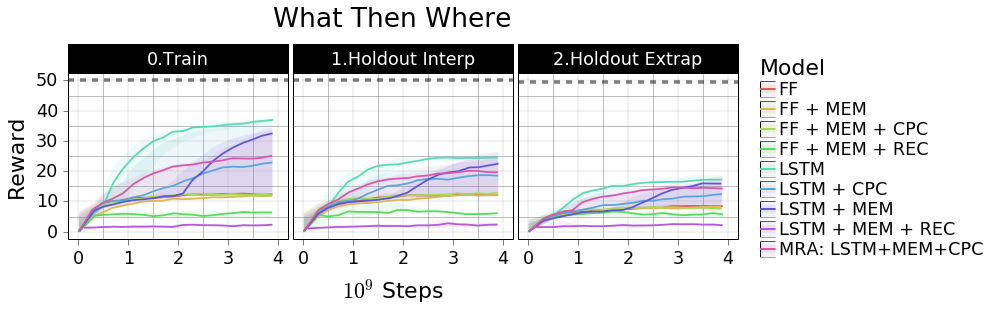}
        }%
    \vspace*{-0.4cm} 
    \subfigure{
        \includegraphics[width=\textwidth]{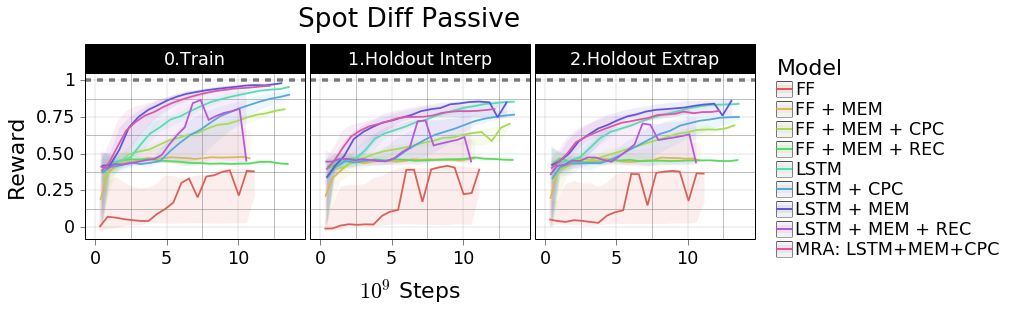}}
\end{figure}        
        
\begin{figure}[H]
    \centering
    \subfigure{
        \includegraphics[width=\textwidth]{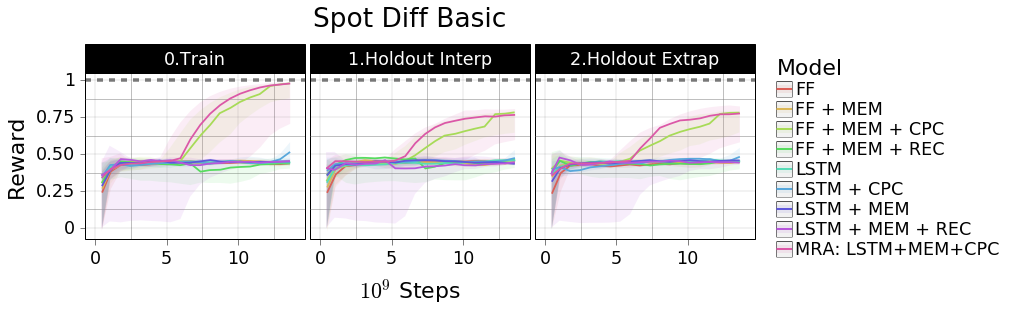}}
  \vspace*{-0.4cm}      
  \subfigure{
        \includegraphics[width=\textwidth]{figures/training_curves/spot_diff_ring_selected.png}}% 
    \vspace*{-0.4cm}    
    \subfigure{
        \includegraphics[width=\textwidth]{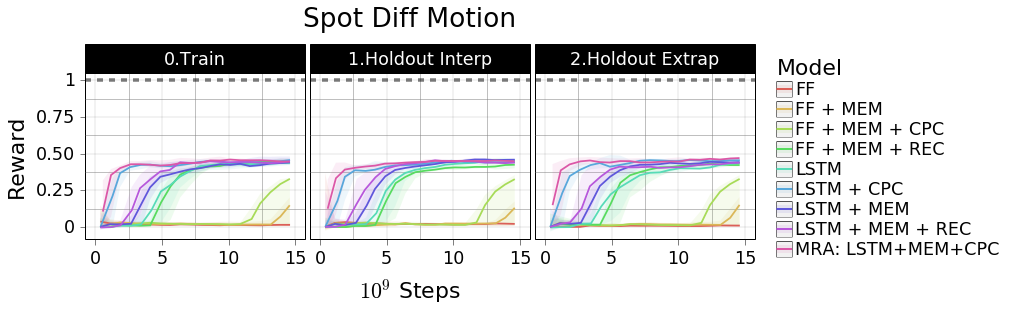}}% 
    \vspace*{-0.4cm}
    \subfigure{
        \includegraphics[width=\textwidth]{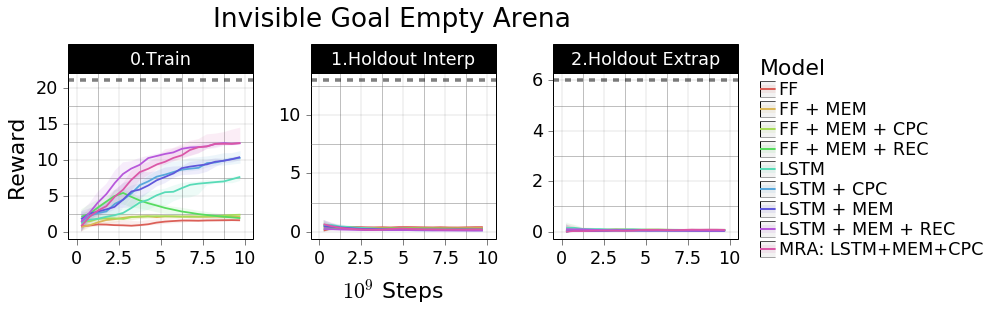}
        }% 
    \vspace*{-0.4cm}     
    \subfigure{
        \includegraphics[width=\textwidth]{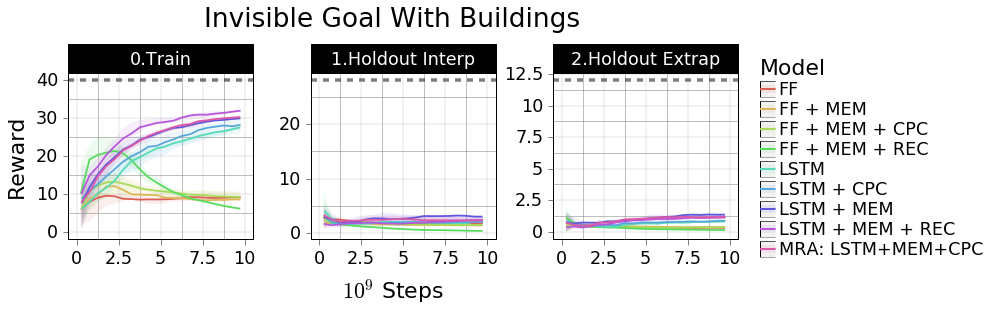}
        }    
\end{figure}

\begin{figure}[H]
    \centering   
    \subfigure{
        \includegraphics[width=\textwidth]{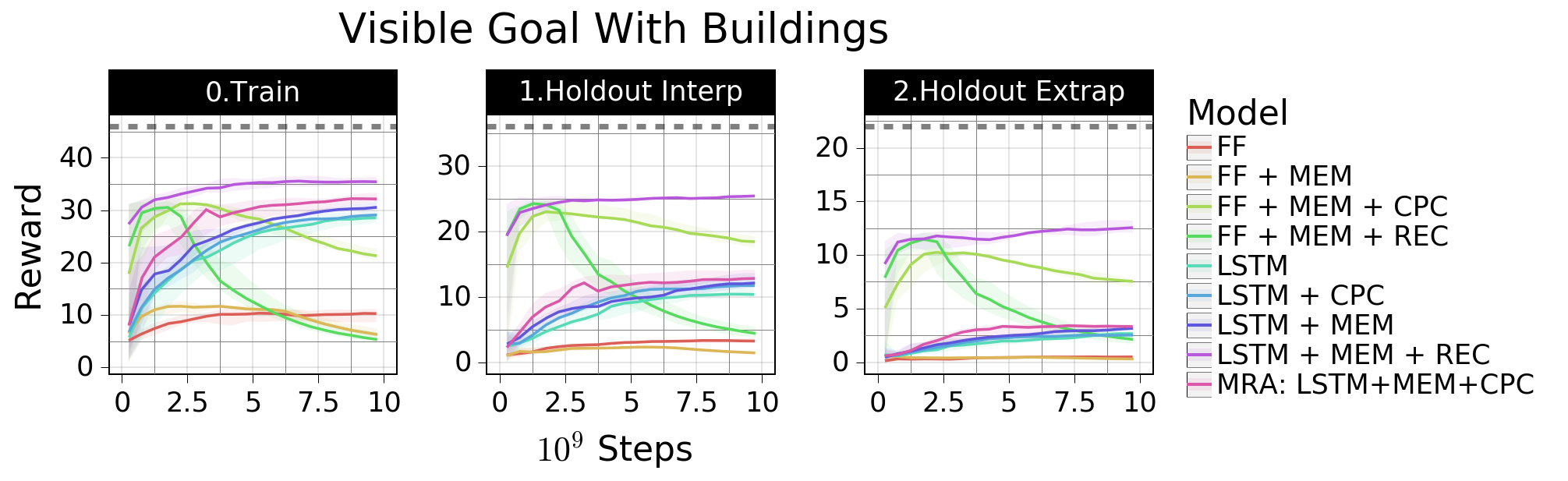}
        }%
    \vspace*{-0.4cm}     
     \subfigure{
        \includegraphics[width=\textwidth]{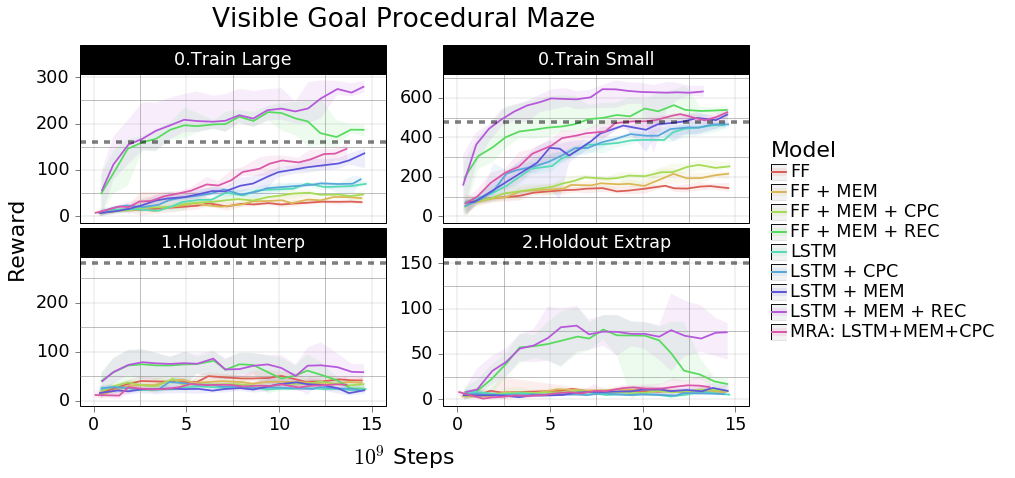}
        }%
    \vspace*{-0.4cm}    
    \subfigure{
        \includegraphics[width=\textwidth]{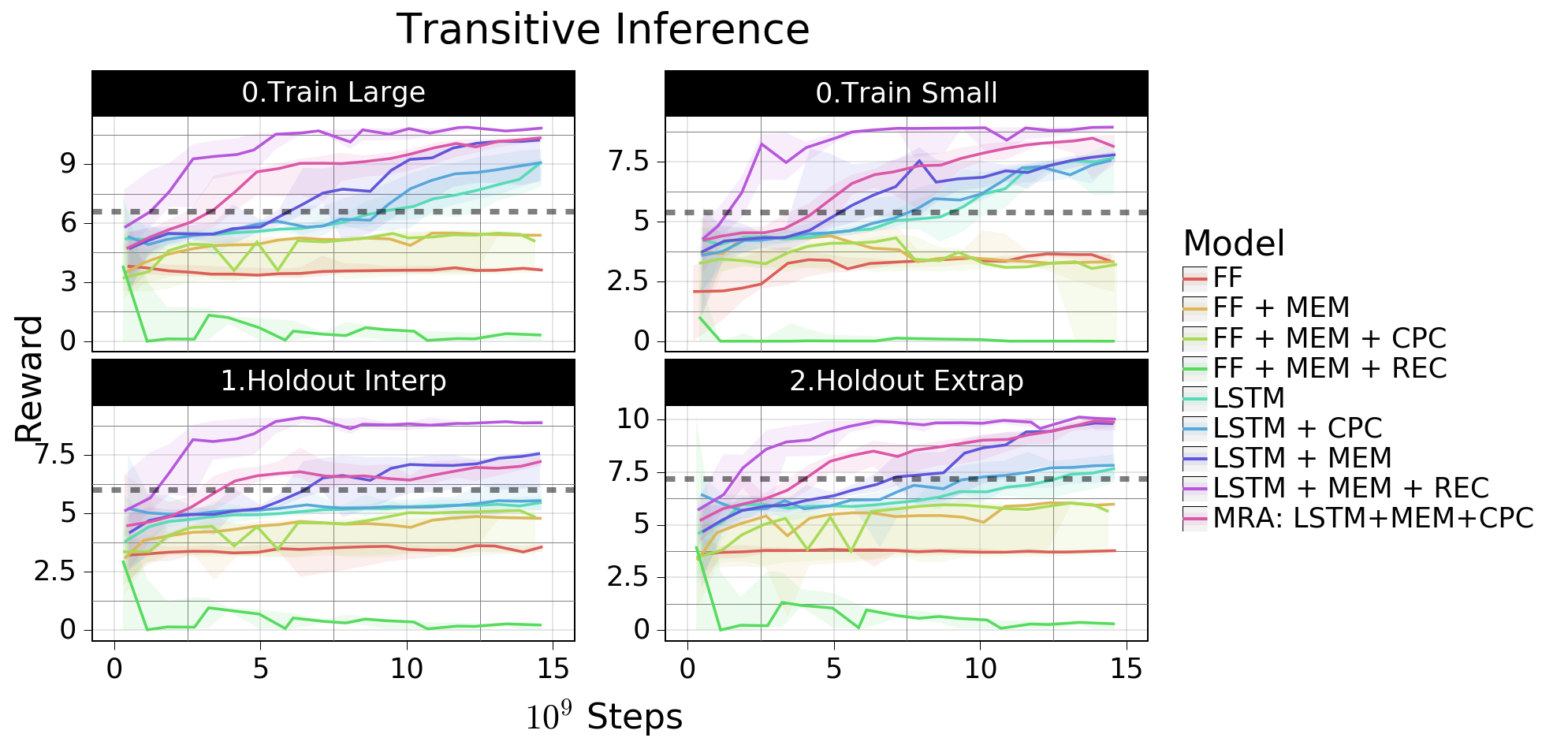}
        }%
    \caption{Training and test curves for all models in all tasks. Dotted lines indicate human baseline scores for each task.}
    \label{fig:curve_all}        
 \end{figure}   
 
 \begin{figure}[H]
    \centering    
    \subfigure{
        \includegraphics[width=0.98\textwidth]{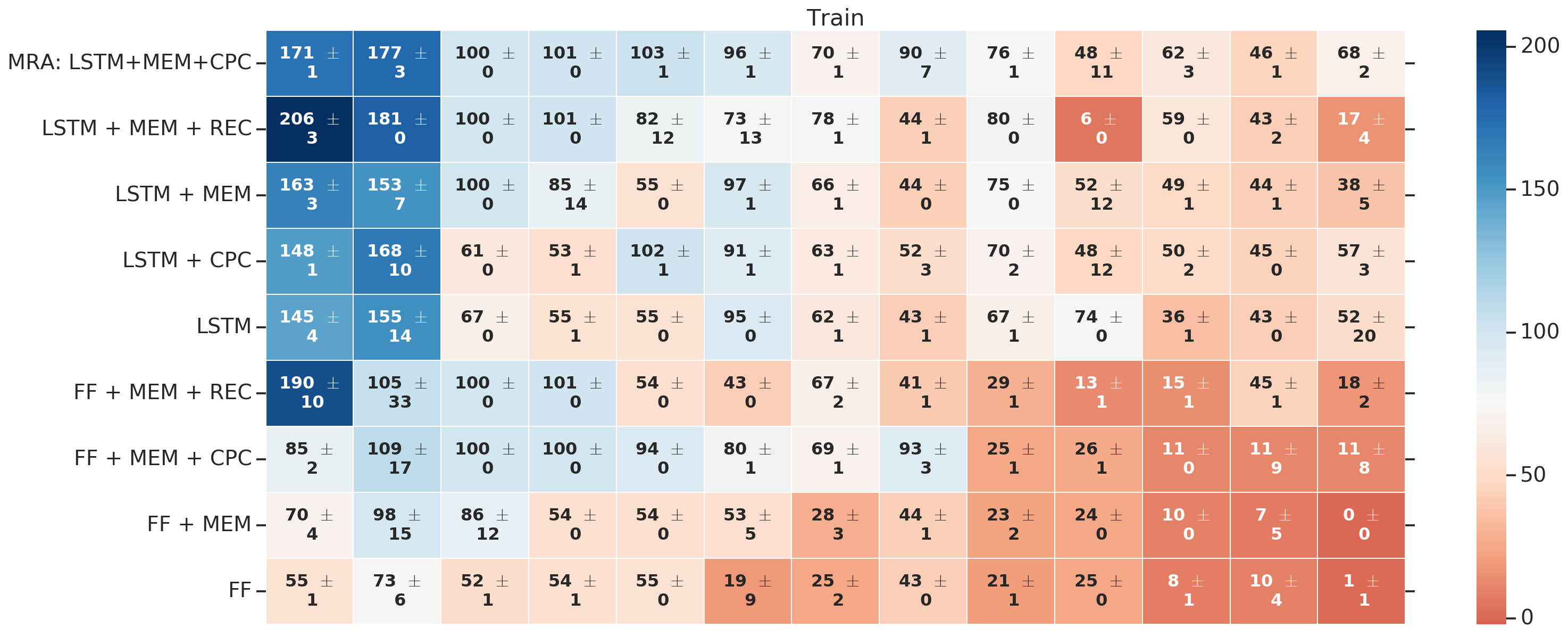}
        }%  
        
    \subfigure{
        \includegraphics[width=0.98\textwidth]{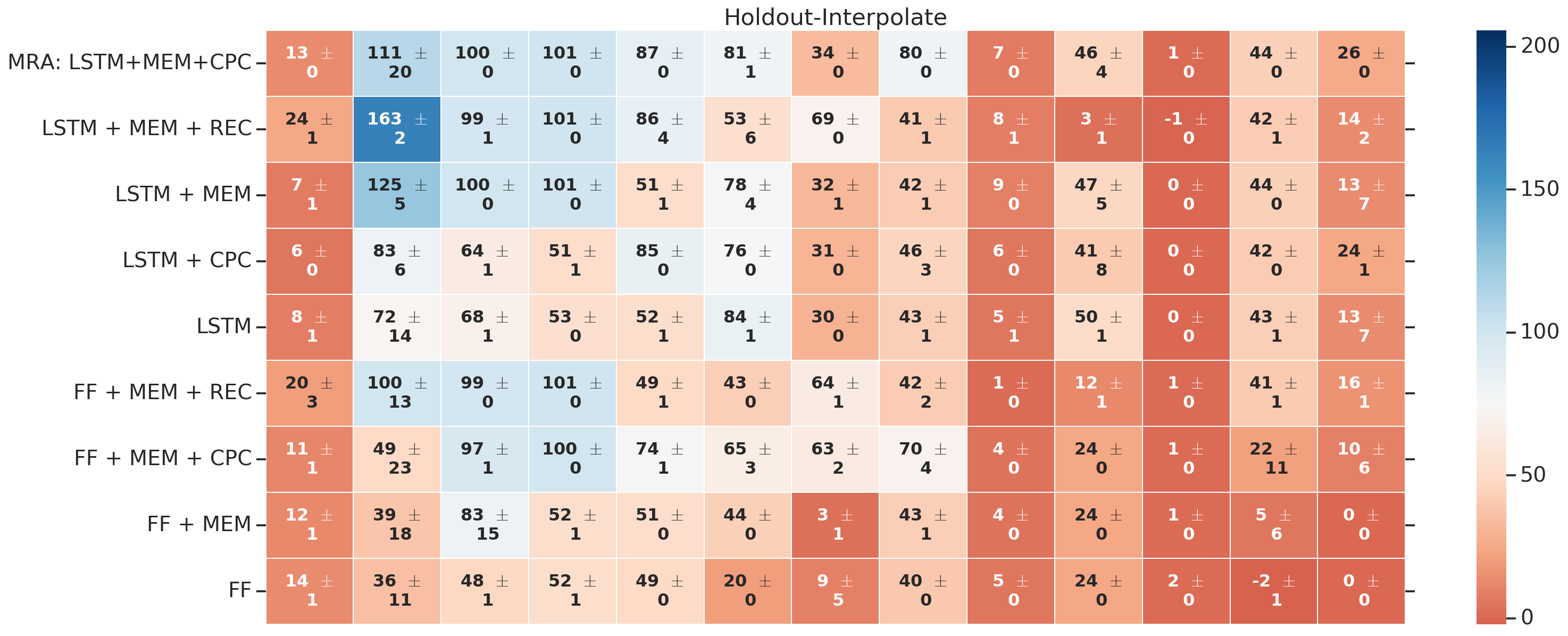}
        }%  
        
    \subfigure{
        \includegraphics[width=0.98\textwidth]{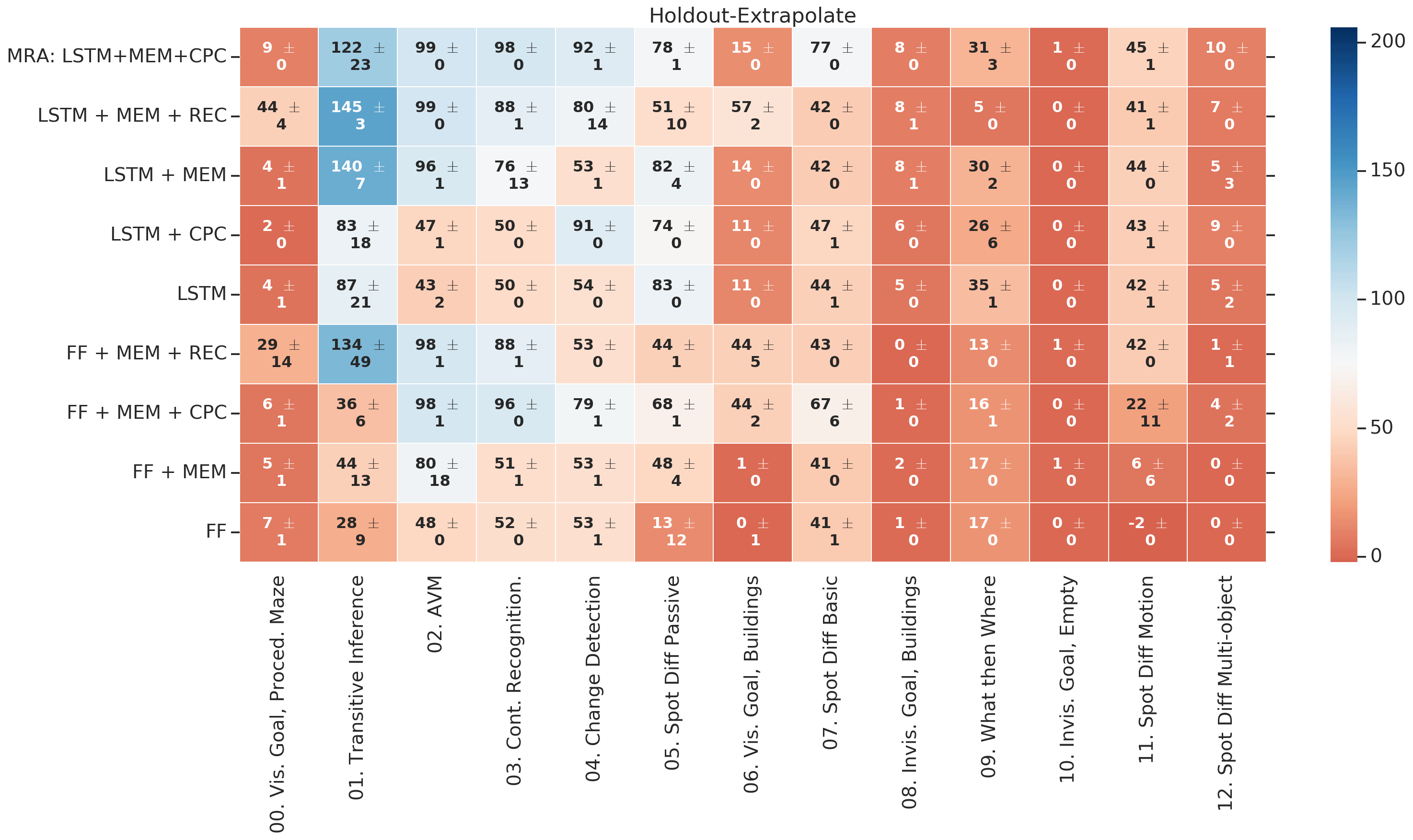}
        }%      
    \caption{Heatmap of ablations per task including standard errors. Tasks are sorted by normalized score across models during training, such that the task with the highest mean scores in training is in the leftmost column, and the model that had the highest mean scores in training is at the top row.}
    \label{appendix:heatmap} 
\end{figure}

%% file: appendix_scores.tex
\section{Human-Normalized Scores and Episode Rewards} 
\label{appendix:results}

We used one action set across all PsychLab tasks, and another across the 3D tasks. 

In PsychLab we used a set of five actions: look left, look right, look up, look down, do nothing. 

For the rest, we used a set of eight actions: move forward, move backward, strafe left, strafe right, look left, look right, look left while moving forward, look right while moving forward. 

In Figure \ref{fig:curve_all} we show the training and test curves for each of our ablation models on all tasks. The curves in bold correspond to the median score across three random seeds, and the corresponding confidence intervals are shown in lighter shades.

\subsection{Human-Normalized Score Computation}

We computed the Human-Normalized Scores used in our heatmap via the following procedure. In our reported results we used three seeds, and took a rolling average as described below.

\begin{enumerate}[nolistsep]
    \item For each seed, apply smoothing in the form of exponential weighted moving average\footnote{For PsychLab tasks and \textit{Visible Goal Procedural Maze}, alpha = 0.05. For the rest, alpha = 0.001.}.
	\item For each seed, take a further rolling average of the episode reward, over a window of 10.
	\item Among these rolling reward windows, find the highest window value over the course of training. The mean over the seeds corresponds to \(R_{Level=Train}\).
	\item For each seed, find the time-step that corresponds to \(R_{Level=Train}\), to use as a snapshot point for comparison against the holdout levels. 
	\item At this snapshot point, record the seed-averaged rolling episode reward for the two holdout levels, \(R_{Level=Holdout-Interpolate}\) and \(R_{Level=Holdout-Extrapolate}\).
	\item Obtain the episode reward of a random agent \(R_{Random}\) and the episode reward achieved by a human, \(R_{Human}\).
	\item For Train, Holdout-Interpolate, and Holdout-Extrapolate, with corresponding standard error: \begin{equation}
	\label{eq:heatmap_score_normalized}
	Human Normalized Score = \frac{R_{Level=\cdot} - R_{Random}}{R_{Human} - R_{Random}} * 100
    \end{equation}
\end{enumerate}

Results are shown ranked (best at top) in Figure \ref{tab:ranked_ablations}.

\begin{table*}[h!]
    \centering
    \caption{Ranking of ablation models, sorted by overall task-averaged human-normalized score.}
    \label{tab:ranked_ablations}
    \begin{tabular}{lcccc}
        \cline{2-5}
         & \multicolumn{4}{c}{Average Human-Normalized Score (percentage points)}\\
        \cline{2-5}
         Model & Train & Holdout-Interpolate & Holdout-Extrapolate \\
        \hline
        \multicolumn{1}{l|}{MRA: LSTM + MEM + CPC}  &  \multicolumn{1}{c|}{92.9 $\pm$ 3.9} & 56.2 $\pm$ 5.8 & 52.6 $\pm$ 6.5\\
        \multicolumn{1}{l|}{LSTM + MEM + REC}  &  \multicolumn{1}{c|}{82.2 $\pm$ 5.2} & 54.2 $\pm$ 2.3 & 51.4 $\pm$ 4.9 \\
        \multicolumn{1}{l|}{LSTM + MEM}  &  \multicolumn{1}{c|}{78.7 $\pm$ 5.8} & 50.0 $\pm$ 3.1 & 45.8 $\pm$ 4.5 \\
        \multicolumn{1}{l|}{LSTM + CPC}  &  \multicolumn{1}{c|}{77.6 $\pm$ 4.6 } & 42.7 $\pm$ 2.8 & 37.7 $\pm$ 5.3 \\
        \multicolumn{1}{l|}{FF + MEM + REC}  &  \multicolumn{1}{c|}{63.1 $\pm$ 9.6} & 45.4 $\pm$ 3.7 & 45.4 $\pm$ 14.2 \\
        \multicolumn{1}{l|}{FF + MEM + CPC}  &  \multicolumn{1}{c|}{62.6 $\pm$ 5.9} & 45.4 $\pm$ 7.3 & 41.3 $\pm$ 4.0 \\
        \multicolumn{1}{l|}{LSTM}  &  \multicolumn{1}{c|}{73.0 $\pm$ 6.9} & 40.2 $\pm$ 4.3& 35.6 $\pm$ 5.9 \\
        \multicolumn{1}{l|}{FF + MEM}  &  \multicolumn{1}{c|}{42.3 $\pm$ 5.8} & 27.8 $\pm$6.7 & 27.0 $\pm$ 6.6 \\
        \multicolumn{1}{l|}{FF}  &  \multicolumn{1}{c|}{33.9 $\pm$ 3.3} & 23.0 $\pm$ 3.5 & 19.7 $\pm$ 4.2 \\
        \hline
    \end{tabular}
\end{table*}

\subsection{Episode Rewards}

Absolute episode rewards per task per level, obtained by trained agent as well as \(R_{Random}\) and \(R_{Human}\), with standard error\footnote{Computed over three seeds for trained agent and for random agent. For human scores, all levels had five trials each except the following: 10 for \textit{Visible Goal with Buildings} and \textit{Invisible Goal Empty Arena}, 19 for the Train level of \textit{Invisible Goal with Buildings} and 20 for the other two levels. The difference was due to time constraints.} bars. See Tables \ref{tab:episode_reward_avm} to \ref{tab:episode_reward_ti}.

\begin{table*}[h!]
    \centering
    \caption{Episode reward: PsychLab - AVM}
    \label{tab:episode_reward_avm}
\resizebox{\textwidth}{!}{%
\begin{tabular}{l|l|l|l}
\textbf{Model} & Train & Holdout-Interpolate & Holdout-Extrapolate \\ \hline
FF & 25.90 $\pm$ 0.32 & 19.37 $\pm$ 0.43 & 35.74 $\pm$ 0.37 \\
FF + MEM & 43.14 $\pm$ 6.12 & 33.34 $\pm$ 5.92 & 60.16 $\pm$ 13.56 \\
FF + MEM + CPC & 49.98 $\pm$ 0.00 & 38.86 $\pm$ 0.32 & 73.66 $\pm$ 0.63 \\
FF + MEM + REC & 50.00 $\pm$ 0.00 & 39.76 $\pm$ 0.14 & 73.17 $\pm$ 0.46 \\
LSTM & 33.35 $\pm$ 0.24 & 27.20 $\pm$ 0.25 & 32.34 $\pm$ 1.20 \\
LSTM + CPC & 30.75 $\pm$ 0.21 & 25.64 $\pm$ 0.26 & 35.50 $\pm$ 1.09 \\
LSTM + MEM & 50.00 $\pm$ 0.00 & 39.99 $\pm$ 0.01 & 72.32 $\pm$ 0.98 \\
LSTM + MEM + REC & 50.00 $\pm$ 0.00 & 39.63 $\pm$ 0.36 & 73.91 $\pm$ 0.28 \\
MRA: LSTM+MEM+CPC & 50.00 $\pm$ 0.00 & 39.99 $\pm$ 0.00 & 74.32 $\pm$ 0.13 \\ \hline
Random & 0.06 $\pm$ 0.00 & 0.06 $\pm$ 0.00 & 0.06 $\pm$ 0.00 \\ \hline
Human & 50.00 $\pm$ 0.00 & 40.00 $\pm$ 0.00 & 75.00 $\pm$ 0.00
\end{tabular}%
}

    \centering
    \caption{Episode reward: PsychLab - Continuous Recognition}
    \label{tab:episode_reward_cont_recog}
\resizebox{\textwidth}{!}{%
\begin{tabular}{l|l|l|l}
\textbf{Model} & Train & Holdout-Interpolate & Holdout-Extrapolate \\ \hline
FF & 26.90 $\pm$ 0.28 & 20.62 $\pm$ 0.28 & 38.45 $\pm$ 0.34 \\
FF + MEM & 26.51 $\pm$ 0.06 & 20.54 $\pm$ 0.49 & 37.96 $\pm$ 0.51 \\
FF + MEM + CPC & 49.60 $\pm$ 0.01 & 39.51 $\pm$ 0.15 & 71.40 $\pm$ 0.15 \\
FF + MEM + REC & 49.78 $\pm$ 0.08 & 39.90 $\pm$ 0.03 & 65.57 $\pm$ 0.56 \\
LSTM & 27.11 $\pm$ 0.29 & 20.92 $\pm$ 0.11 & 37.28 $\pm$ 0.18 \\
LSTM + CPC & 26.25 $\pm$ 0.26 & 20.11 $\pm$ 0.55 & 37.46 $\pm$ 0.36 \\
LSTM + MEM & 42.18 $\pm$ 6.93 & 39.68 $\pm$ 0.06 & 56.59 $\pm$ 9.84 \\
LSTM + MEM + REC & 49.78 $\pm$ 0.08 & 39.90 $\pm$ 0.03 & 65.57 $\pm$ 0.56 \\
MRA: LSTM+MEM+CPC & 49.92 $\pm$ 0.03 & 39.83 $\pm$ 0.00 & 72.52 $\pm$ 0.25 \\ \hline
Random & 0.04 $\pm$ 0.00 & 0.05 $\pm$ 0.00 & 0.05 $\pm$ 0.00 \\ \hline
Human & 49.40 $\pm$ 0.24 & 39.40 $\pm$ 0.40 & 74.20 $\pm$ 0.58
\end{tabular}%
}

    \centering
    \caption{Episode reward: PsychLab - Change Detection}
    \label{tab:episode_reward_change_detection}
\resizebox{\textwidth}{!}{%
\begin{tabular}{l|l|l|l}
\textbf{Model} & Train & Holdout-Interpolate & Holdout-Extrapolate \\ \hline
FF & 26.40 $\pm$ 0.08 & 24.17 $\pm$ 0.10 & 24.73 $\pm$ 0.48 \\
FF + MEM & 25.76 $\pm$ 0.16 & 24.95 $\pm$ 0.12 & 24.97 $\pm$ 0.36 \\
FF + MEM + CPC & 44.76 $\pm$ 0.05 & 36.07 $\pm$ 0.46 & 36.95 $\pm$ 0.40 \\
FF + MEM + REC & 25.82 $\pm$ 0.22 & 23.99 $\pm$ 0.29 & 24.89 $\pm$ 0.23 \\
LSTM & 26.39 $\pm$ 0.21 & 25.24 $\pm$ 0.43 & 25.37 $\pm$ 0.22 \\
LSTM + CPC & 48.37 $\pm$ 0.39 & 41.43 $\pm$ 0.12 & 42.72 $\pm$ 0.11 \\
LSTM + MEM & 26.21 $\pm$ 0.10 & 24.77 $\pm$ 0.28 & 24.63 $\pm$ 0.52 \\
LSTM + MEM + REC & 39.12 $\pm$ 5.88 & 42.12 $\pm$ 1.97 & 37.31 $\pm$ 6.38 \\
MRA: LSTM+MEM+CPC & 49.14 $\pm$ 0.24 & 42.24 $\pm$ 0.07 & 43.00 $\pm$ 0.39 \\ \hline
Random & 0.00 $\pm$ 0.00 & 0.00 $\pm$ 0.00 & 0.00 $\pm$ 0.00 \\ \hline
Human & 47.60 $\pm$ 0.40 & 48.80 $\pm$ 0.58 & 46.80 $\pm$ 1.07
\end{tabular}%
}

    \centering
    \caption{Episode reward: PsychLab - What Then Where}
    \label{tab:episode_reward_wtw}
\resizebox{\textwidth}{!}{%
\begin{tabular}{l|l|l|l}
\textbf{Model} & Train & Holdout-Interpolate & Holdout-Extrapolate \\ \hline
FF & 12.71 $\pm$ 0.06 & 12.19 $\pm$ 0.06 & 8.39 $\pm$ 0.16 \\
FF + MEM & 12.11 $\pm$ 0.14 & 12.05 $\pm$ 0.12 & 8.34 $\pm$ 0.11 \\
FF + MEM + CPC & 12.92 $\pm$ 0.32 & 12.21 $\pm$ 0.20 & 7.73 $\pm$ 0.29 \\
FF + MEM + REC & 6.54 $\pm$ 0.50 & 6.10 $\pm$ 0.35 & 6.30 $\pm$ 0.10 \\
LSTM & 37.18 $\pm$ 0.14 & 25.06 $\pm$ 0.34 & 17.51 $\pm$ 0.38 \\
LSTM + CPC & 24.21 $\pm$ 6.04 & 20.68 $\pm$ 3.89 & 12.99 $\pm$ 2.79 \\
LSTM + MEM & 26.19 $\pm$ 6.23 & 23.74 $\pm$ 2.65 & 14.72 $\pm$ 1.22 \\
LSTM + MEM + REC & 2.96 $\pm$ 0.04 & 1.71 $\pm$ 0.27 & 2.34 $\pm$ 0.22 \\
MRA: LSTM+MEM+CPC & 24.22 $\pm$ 5.45 & 23.10 $\pm$ 1.82 & 15.54 $\pm$ 1.39 \\ \hline
Random & 0.02 $\pm$ 0.00 & 0.03 $\pm$ 0.00 & 0.01 $\pm$ 0.00 \\ \hline
Human & 50.00 $\pm$ 0.00 & 50.00 $\pm$ 0.00 & 49.60 $\pm$ 0.24
\end{tabular}%
}
\end{table*}

\begin{table*}[h!]
    \centering
    \caption{Episode reward: Spot Diff Basic}
    \label{tab:episode_reward_sd_basic}
\resizebox{\textwidth}{!}{%
\begin{tabular}{l|l|l|l}
\textbf{Model} & Train & Holdout-Interpolate & Holdout-Extrapolate \\ \hline
FF & 0.46 $\pm$ 0.00 & 0.43 $\pm$ 0.00 & 0.43 $\pm$ 0.01 \\
FF + MEM & 0.46 $\pm$ 0.01 & 0.45 $\pm$ 0.01 & 0.44 $\pm$ 0.00 \\
FF + MEM + CPC & 0.93 $\pm$ 0.02 & 0.71 $\pm$ 0.04 & 0.69 $\pm$ 0.05 \\
FF + MEM + REC & 0.44 $\pm$ 0.01 & 0.45 $\pm$ 0.02 & 0.45 $\pm$ 0.00 \\
LSTM & 0.46 $\pm$ 0.01 & 0.45 $\pm$ 0.01 & 0.46 $\pm$ 0.01 \\
LSTM + CPC & 0.54 $\pm$ 0.03 & 0.48 $\pm$ 0.03 & 0.49 $\pm$ 0.01 \\
LSTM + MEM & 0.47 $\pm$ 0.00 & 0.45 $\pm$ 0.01 & 0.45 $\pm$ 0.00 \\
LSTM + MEM + REC & 0.46 $\pm$ 0.01 & 0.44 $\pm$ 0.01 & 0.45 $\pm$ 0.00 \\
MRA: LSTM+MEM+CPC & 0.90 $\pm$ 0.07 & 0.81 $\pm$ 0.00 & 0.78 $\pm$ 0.00 \\ \hline
Random & 0.05 $\pm$ 0.00 & 0.04 $\pm$ 0.00 & 0.04 $\pm$ 0.00 \\ \hline
Human & 1.00 $\pm$ 0.00 & 1.00 $\pm$ 0.00 & 1.00 $\pm$ 0.00
\end{tabular}%
}
    \centering
    \caption{Episode reward: Spot Diff Passive}
    \label{tab:episode_reward_sd_passive}
\resizebox{\textwidth}{!}{%
\begin{tabular}{l|l|l|l}
\textbf{Model} & Train & Holdout-Interpolate & Holdout-Extrapolate \\ \hline
FF & 0.22 $\pm$ 0.09 & 0.23 $\pm$ 0.00 & 0.14 $\pm$ 0.11 \\
FF + MEM & 0.54 $\pm$ 0.05 & 0.46 $\pm$ 0.00 & 0.49 $\pm$ 0.04 \\
FF + MEM + CPC & 0.80 $\pm$ 0.01 & 0.66 $\pm$ 0.03 & 0.68 $\pm$ 0.01 \\
FF + MEM + REC & 0.45 $\pm$ 0.00 & 0.45 $\pm$ 0.00 & 0.45 $\pm$ 0.01 \\
LSTM & 0.95 $\pm$ 0.00 & 0.85 $\pm$ 0.01 & 0.84 $\pm$ 0.00 \\
LSTM + CPC & 0.91 $\pm$ 0.01 & 0.77 $\pm$ 0.00 & 0.75 $\pm$ 0.00 \\
LSTM + MEM & 0.97 $\pm$ 0.01 & 0.78 $\pm$ 0.04 & 0.83 $\pm$ 0.03 \\
LSTM + MEM + REC & 0.74 $\pm$ 0.12 & 0.54 $\pm$ 0.06 & 0.52 $\pm$ 0.09 \\
MRA: LSTM+MEM+CPC & 0.96 $\pm$ 0.01 & 0.82 $\pm$ 0.01 & 0.78 $\pm$ 0.01 \\ \hline
Random & 0.03 $\pm$ 0.00 & 0.03 $\pm$ 0.00 & 0.02 $\pm$ 0.00 \\ \hline
Human & 1.00 $\pm$ 0.00 & 1.00 $\pm$ 0.00 & 1.00 $\pm$ 0.00
\end{tabular}%
}
    \centering
    \caption{Episode reward: Spot Diff Multi-Object}
    \label{tab:episode_reward_sd_ring}
\resizebox{\textwidth}{!}{%
\begin{tabular}{l|l|l|l}
\textbf{Model} & Train & Holdout-Interpolate & Holdout-Extrapolate \\ \hline
FF & 0.02 $\pm$ 0.01 & 0.01 $\pm$ 0.00 & 0.00 $\pm$ 0.00 \\
FF + MEM & 0.01 $\pm$ 0.00 & 0.01 $\pm$ 0.00 & 0.00 $\pm$ 0.00 \\
FF + MEM + CPC & 0.12 $\pm$ 0.08 & 0.11 $\pm$ 0.06 & 0.04 $\pm$ 0.02 \\
FF + MEM + REC & 0.18 $\pm$ 0.02 & 0.17 $\pm$ 0.01 & 0.01 $\pm$ 0.01 \\
LSTM & 0.52 $\pm$ 0.20 & 0.14 $\pm$ 0.07 & 0.05 $\pm$ 0.02 \\
LSTM + CPC & 0.58 $\pm$ 0.03 & 0.24 $\pm$ 0.01 & 0.09 $\pm$ 0.00 \\
LSTM + MEM & 0.39 $\pm$ 0.05 & 0.14 $\pm$ 0.07 & 0.05 $\pm$ 0.03 \\
LSTM + MEM + REC & 0.18 $\pm$ 0.04 & 0.15 $\pm$ 0.02 & 0.07 $\pm$ 0.00 \\
MRA: LSTM+MEM+CPC & 0.69 $\pm$ 0.02 & 0.27 $\pm$ 0.00 & 0.10 $\pm$ 0.00 \\ \hline
Random & 0.01 $\pm$ 0.00 & 0.01 $\pm$ 0.00 & 0.00 $\pm$ 0.00 \\ \hline
Human & 1.00 $\pm$ 0.00 & 1.00 $\pm$ 0.00 & 1.00 $\pm$ 0.00
\end{tabular}%
}
    \centering
    \caption{Episode reward: Spot Diff Motion}
    \label{tab:episode_reward_sd_motion}
\resizebox{\textwidth}{!}{%
\begin{tabular}{l|l|l|l}
\textbf{Model} & Train & Holdout-Interpolate & Holdout-Extrapolate \\ \hline
FF & 0.12 $\pm$ 0.04 & 0.00 $\pm$ 0.01 & 0.00 $\pm$ 0.00 \\
FF + MEM & 0.08 $\pm$ 0.05 & 0.07 $\pm$ 0.05 & 0.08 $\pm$ 0.06 \\
FF + MEM + CPC & 0.13 $\pm$ 0.09 & 0.24 $\pm$ 0.10 & 0.23 $\pm$ 0.11 \\
FF + MEM + REC & 0.46 $\pm$ 0.01 & 0.42 $\pm$ 0.01 & 0.43 $\pm$ 0.00 \\
LSTM & 0.45 $\pm$ 0.00 & 0.44 $\pm$ 0.01 & 0.43 $\pm$ 0.01 \\
LSTM + CPC & 0.46 $\pm$ 0.00 & 0.43 $\pm$ 0.00 & 0.44 $\pm$ 0.01 \\
LSTM + MEM & 0.45 $\pm$ 0.01 & 0.46 $\pm$ 0.00 & 0.45 $\pm$ 0.00 \\
LSTM + MEM + REC & 0.44 $\pm$ 0.02 & 0.44 $\pm$ 0.01 & 0.42 $\pm$ 0.01 \\
MRA: LSTM+MEM+CPC & 0.47 $\pm$ 0.01 & 0.45 $\pm$ 0.00 & 0.46 $\pm$ 0.01 \\ \hline
Random & 0.02 $\pm$ 0.00 & 0.02 $\pm$ 0.00 & 0.02 $\pm$ 0.00 \\ \hline
Human & 1.00 $\pm$ 0.00 & 1.00 $\pm$ 0.00 & 1.00 $\pm$ 0.00
\end{tabular}%
}
\end{table*}

\begin{table*}[h!]
    \centering
    \caption{Episode reward: Visible Goal With Buildings}
    \label{tab:episode_reward_mini_city_visible_goal}
\resizebox{\textwidth}{!}{%
\begin{tabular}{l|l|l|l}
\textbf{Model} & Train & Holdout-Interpolate & Holdout-Extrapolate \\ \hline
FF & 12.27 $\pm$ 0.83 & 3.74 $\pm$ 1.70 & 0.14 $\pm$ 0.14 \\
FF + MEM & 13.58 $\pm$ 1.45 & 1.52 $\pm$ 0.28 & 0.47 $\pm$ 0.02 \\
FF + MEM + CPC & 31.87 $\pm$ 0.25 & 22.99 $\pm$ 0.70 & 9.83 $\pm$ 0.40 \\
FF + MEM + REC & 31.01 $\pm$ 0.77 & 23.42 $\pm$ 0.33 & 9.84 $\pm$ 1.09 \\
LSTM & 28.72 $\pm$ 0.34 & 11.37 $\pm$ 0.00 & 2.66 $\pm$ 0.00 \\
LSTM + CPC & 29.46 $\pm$ 0.23 & 11.52 $\pm$ 0.08 & 2.52 $\pm$ 0.02 \\
LSTM + MEM & 30.92 $\pm$ 0.28 & 11.74 $\pm$ 0.22 & 3.29 $\pm$ 0.10 \\
LSTM + MEM + REC & 35.95 $\pm$ 0.28 & 25.16 $\pm$ 0.05 & 12.54 $\pm$ 0.34 \\
MRA: LSTM+MEM+CPC & 32.45 $\pm$ 0.39 & 12.66 $\pm$ 0.06 & 3.38 $\pm$ 0.05 \\ \hline
Random & 1.08 $\pm$ 0.02 & 0.58 $\pm$ 0.01 & 0.22 $\pm$ 0.01 \\ \hline
Human & 23.60 $\pm$ 1.69 & 23.50 $\pm$ 0.75 & 14.30 $\pm$ 0.60
\end{tabular}%
}
    \centering
    \caption{Episode reward: Invisible Goal With Buildings}
    \label{tab:episode_reward_mini_city_watermaze}
\resizebox{\textwidth}{!}{%
\begin{tabular}{l|l|l|l}
\textbf{Model} & Train & Holdout-Interpolate & Holdout-Extrapolate \\ \hline
FF & 9.30 $\pm$ 0.33 & 1.80 $\pm$ 0.01 & 0.28 $\pm$ 0.02 \\
FF + MEM & 9.95 $\pm$ 0.65 & 1.54 $\pm$ 0.02 & 0.48 $\pm$ 0.00 \\
FF + MEM + CPC & 10.65 $\pm$ 0.28 & 1.52 $\pm$ 0.10 & 0.31 $\pm$ 0.01 \\
FF + MEM + REC & 12.29 $\pm$ 0.24 & 0.70 $\pm$ 0.05 & 0.20 $\pm$ 0.00 \\
LSTM & 27.22 $\pm$ 0.48 & 2.01 $\pm$ 0.19 & 0.79 $\pm$ 0.01 \\
LSTM + CPC & 28.46 $\pm$ 0.63 & 2.25 $\pm$ 0.01 & 0.86 $\pm$ 0.01 \\
LSTM + MEM & 30.15 $\pm$ 0.04 & 3.10 $\pm$ 0.08 & 1.17 $\pm$ 0.08 \\
LSTM + MEM + REC & 32.10 $\pm$ 0.05 & 2.75 $\pm$ 0.24 & 1.17 $\pm$ 0.09 \\
MRA: LSTM+MEM+CPC & 30.51 $\pm$ 0.21 & 2.39 $\pm$ 0.09 & 1.09 $\pm$ 0.03 \\ \hline
Random & 0.95 $\pm$ 0.02 & 0.53 $\pm$ 0.01 & 0.20 $\pm$ 0.01 \\ \hline
Human & 17.37 $\pm$ 1.91 & 12.40 $\pm$ 1.45 & 4.90 $\pm$ 0.71
\end{tabular}%
}
    \centering
    \caption{Episode reward: Invisible Goal Empty Arena}
    \label{tab:episode_reward_empty_room_watermaze}
\resizebox{\textwidth}{!}{%
\begin{tabular}{l|l|l|l}
\textbf{Model} & Train & Holdout-Interpolate & Holdout-Extrapolate \\ \hline
FF & 1.78 $\pm$ 0.11 & 0.38 $\pm$ 0.04 & 0.05 $\pm$ 0.00 \\
FF + MEM & 2.21 $\pm$ 0.05 & 0.28 $\pm$ 0.02 & 0.07 $\pm$ 0.01 \\
FF + MEM + CPC & 2.37 $\pm$ 0.07 & 0.22 $\pm$ 0.02 & 0.05 $\pm$ 0.00 \\
FF + MEM + REC & 3.25 $\pm$ 0.25 & 0.27 $\pm$ 0.02 & 0.06 $\pm$ 0.01 \\
LSTM & 7.60 $\pm$ 0.14 & 0.14 $\pm$ 0.01 & 0.05 $\pm$ 0.01 \\
LSTM + CPC & 10.48 $\pm$ 0.35 & 0.12 $\pm$ 0.02 & 0.03 $\pm$ 0.01 \\
LSTM + MEM & 10.32 $\pm$ 0.12 & 0.19 $\pm$ 0.02 & 0.03 $\pm$ 0.01 \\
LSTM + MEM + REC & 12.40 $\pm$ 0.08 & 0.08 $\pm$ 0.01 & 0.04 $\pm$ 0.00 \\
MRA: LSTM+MEM+CPC & 13.04 $\pm$ 0.60 & 0.23 $\pm$ 0.04 & 0.07 $\pm$ 0.01 \\ \hline
Random & 0.15 $\pm$ 0.01 & 0.15 $\pm$ 0.01 & 0.03 $\pm$ 0.00 \\ \hline
Human & 4.90 $\pm$ 1.32 & 1.70 $\pm$ 0.67 & 0.30 $\pm$ 0.30
\end{tabular}%
}
\end{table*}

\begin{table*}[h!]
    \centering
    \caption{Episode reward: Visible Goal Procedural Maze}
    \label{tab:episode_reward_explore_goal_locations}
\resizebox{\textwidth}{!}{%
\begin{tabular}{l|l|l|l}
\textbf{Model} & Train & Holdout-Interpolate & Holdout-Extrapolate \\ \hline
FF & 174.63 $\pm$ 4.27 & 43.55 $\pm$ 3.08 & 11.93 $\pm$ 2.17 \\
FF + MEM & 224.53 $\pm$ 11.31 & 37.80 $\pm$ 2.28 & 8.52 $\pm$ 0.87 \\
FF + MEM + CPC & 272.99 $\pm$ 5.31 & 33.38 $\pm$ 1.51 & 9.99 $\pm$ 1.10 \\
FF + MEM + REC & 607.48 $\pm$ 36.98 & 59.64 $\pm$ 9.41 & 43.07 $\pm$ 19.35 \\
LSTM & 463.43 $\pm$ 12.84 & 27.09 $\pm$ 1.69 & 6.52 $\pm$ 1.86 \\
LSTM + CPC & 473.42 $\pm$ 3.84 & 19.72 $\pm$ 0.98 & 4.90 $\pm$ 0.36 \\
LSTM + MEM & 523.14 $\pm$ 10.52 & 22.39 $\pm$ 3.65 & 7.83 $\pm$ 1.25 \\
LSTM + MEM + REC & 655.10 $\pm$ 11.65 & 49.53 $\pm$ 24.78 & 57.21 $\pm$ 23.92 \\
MRA: LSTM+MEM+CPC & 546.08 $\pm$ 2.26 & 40.64 $\pm$ 0.00 & 14.21 $\pm$ 0.00 \\ \hline
Random & Small: 7.79 $\pm$ 0.14 & 3.89 $\pm$ 0.10 & 1.19 $\pm$ 0.05 \\ 
 & Large: 1.97 $\pm$ 0.06 & & \\  \hline
Human & Small: 364.00 $\pm$ 43.20 & 198.00 $\pm$ 24.98 & 86.00 $\pm$ 20.15 \\
 & Large: 104.00 $\pm$ 23.58 & & 
\end{tabular}%
}
\end{table*}

\begin{table*}[h!]
    \centering
    \caption{Episode reward: Transitive Inference}
    \label{tab:episode_reward_ti}
\resizebox{\textwidth}{!}{%
\begin{tabular}{l|l|l|l}
\textbf{Model} & Train & Holdout-Interpolate & Holdout-Extrapolate \\ \hline
FF & 3.71 $\pm$ 0.07 & 3.52 $\pm$ 0.06 & 3.73 $\pm$ 0.04 \\
FF + MEM & 4.34 $\pm$ 0.51 & 4.09 $\pm$ 0.48 & 5.76 $\pm$ 0.18 \\
FF + MEM + CPC & 4.64 $\pm$ 0.76 & 5.16 $\pm$ 0.01 & 5.62 $\pm$ 0.41 \\
FF + MEM + REC & 0.47 $\pm$ 0.22 & 0.30 $\pm$ 0.01 & 0.69 $\pm$ 0.12 \\
LSTM & 8.67 $\pm$ 0.55 & 5.31 $\pm$ 0.10 & 7.32 $\pm$ 0.38 \\
LSTM + CPC & 9.65 $\pm$ 0.53 & 5.59 $\pm$ 0.02 & 7.77 $\pm$ 0.08 \\
LSTM + MEM & 8.98 $\pm$ 0.42 & 6.76 $\pm$ 0.75 & 8.84 $\pm$ 0.41 \\
LSTM + MEM + REC & 10.86 $\pm$ 0.02 & 8.88 $\pm$ 0.11 & 9.80 $\pm$ 0.16 \\
MRA: LSTM+MEM+CPC & 10.34 $\pm$ 0.10 & 7.21 $\pm$ 0.19 & 9.81 $\pm$ 0.09 \\ \hline
Random & Small: 1.44 $\pm$ 0.02 & 1.42 $\pm$ 0.02 & 1.43 $\pm$ 0.02 \\
 & Large: 1.44 $\pm$ 0.02 &  &  \\ \hline
Human & Small: 5.40 $\pm$ 2.20 & 6.00 $\pm$ 2.45 & 7.20 $\pm$ 2.94 \\
 & Large: 6.60 $\pm$ 2.69 & & 
\end{tabular}%
}
\end{table*}

%% file: hypers.tex
\section{Hyper-parameter Tuning} 
\label{hypers}

All experiments used three seeds, with identical hyper-parameters each. Given the scope of the experiments undertaken, all hyper-parameter tuning was preliminary and not exhaustive. 

Initial hyper-parameters were either inherited from the IMPALA paper or given an arbitrary first-guess value that seemed reasonable. Whatever tuning that was done was performed in a relatively systematic way: Hyper-parameters were shared across all model variations, and tuned with the objective of getting as many model variations as possible to achieve adequate performance on the training tasks.

The PsychLab tasks were the ones with the most tuning. For PsychLab, we performed a manual sweep over arbitrary reasonable-seeming values when train performance wasn't getting off the floor or was too noisy. We had a preference for hypers that fared well across all models (e.g. choosing a bigger hidden size of 1024 rather than 512 for the controller so that FF models would have capacity). 

For the other tasks, very minimal tuning occurred and hyper-parameters were first-guess. With Spot the Difference, we tried two different discount rates and went with the better one. For Goal Navigation and Transitive Inference tasks, we stuck to a standardized discount rate of 0.99.

We did not perform any tuning for REC throughout.

\paragraph{Fixed hyper-parameters} (See Table \ref{tab:hypers_fixed})
For optimizers, whenever we used Adam we standardized the discount rate to 0.98, and whenever we used RMSProp the discount rate was mostly 0.99 except in certain cases where we were able to also try 0.999 and found that it did better. Whenever we used an external episodic memory module (`MEM') we used the fixed hyper-parameters in Table \ref{tab:hypers_fixed}. 

For individual task hyper-parameter configurations see Table \ref{tab:hypers}.

\begin{table*}[htb]
    \caption{Fixed hyper-parameters}
    \label{tab:hypers_fixed}
    \centering
    \begin{tabular}{l|c}
    \hline
    \textbf{Optimizer} & \\ 
    \hline
    Adam: & \\
    Beta1 & 0.9 \\
    Beta2 & 0.999 \\
    Epsilon & 1e-4 \\
    \hline
    RMSProp: & \\
    Epsilon & 0.1 \\
    Momentum (Inherited from IMPALA paper) & 0.0  \\
    Decay & 0.99 \\
    \hline
    \hline
    \textbf{MEM} & \\
    Number of k-nearest neighbors to retrieve from MEM & 10 \\
    MEM key size (and accordingly, query size) & 128 \\
    Capacity (max number of timesteps storable) & 2048 for Unity levels, else 1024 \\ 
    \bottomrule
    \end{tabular}
\end{table*}

\begin{sidewaystable}
\caption{Hyper-parameters} 
\label{tab:hypers}
\centering
\begin{tabular}{lcccccccccc}
\hline
Parameter & Hidden size & Baseline cost\tablefootnote{Inherited from IMPALA paper, except for \textit{What Then Where}.} & Entropy & Batch size & Unroll & Discount & Optimizer & Learning & Num & CPC \\
 & & & & & length & & & rate & CPC steps & weight\\
\hline
\hline
\textbf{PsychLab} & & & & & & & & & & \\
AVM & 512 & 0.5 & 0.00520\tablefootnote{Copied from previous work, not tuned for this paper. 0.01 was slower and noisier.} & 16 & 50 & 0.98 & Adam & 1e-5 & 10 & 10 \\

\hline
Cont. Recognition & 1024 & 0.5 & 0.01 & 16 & 50 & 0.98 & Adam & 1e-5 & 10 & 10 \\

\hline
Change Detection & 512 & 0.5 & 0.01 & 16 & 50 & 0.98 & Adam & 1e-5 & 10 & 10 \\

\hline
What Then Where & 1024 & 2.0 & 0.01 & 32 & 100 & 0.98 & Adam & 1e-5 & 10 & 30 \\
 & & Sweep [0.5, 1.0, 2.0] & & & & & & & & Sweep [10, 30] \\
\hline
\hline
\textbf{Spot Diff} & & & & & & \tablefootnote{Sweep over [0.99, .999] throughout.} & & & & \\
Basic & 1024 & 0.5 & 0.003 & 16 & 200 & 0.99 & RMSProp & 1e-4 & 50 & 20 \\

\hline
Passive & 1024 & 0.5 & 0.003 & 16 & 200 & 0.999 & RMSProp & 1e-4 & 50 & 20 \\

\hline
Multi-object & 1024 & 0.5 & 0.003 & 16 & 200 & 0.99 & RMSProp & 1e-4 & 50 & 20 \\

\hline
Motion & 1024 & 0.5 & 0.003 & 16 & 200 & 0.99 & RMSProp & 1e-4 & 50 & 20 \\

\hline
\hline
\textbf{Goal Navigation} & & & & & & \tablefootnote{Sweep over [0.98, .99] throughout.} & & & & \\
Visible Goal, & 512 & 0.5 & 0.00520\tablefootnote{Copied from AVM} & 16 & 50 & 0.98 & Adam & 1e-5 & 10 & 5 \\
Proced. Maze & & & & & & & & & & \\
\hline
Visible Goal,  & 1024 & 0.5 & 0.003 & 16 & 200 & 0.99 & RMSProp & 1e-4 & 50 & 20 \\
With Buildings & & & & & & & & & & \\
\hline
Invisible Goal & 1024 & 0.5 & 0.003 & 16 & 200 & 0.99 & RMSProp & 1e-4 & 50 & 20 \\
With Buildings & & & & & & & & & & \\
\hline
Invisible Goal & 1024 & 0.5 & 0.003 & 16 & 200 & 0.99 & RMSProp & 1e-4 & 50 & 20 \\

Empty Arena & & & & & & & & & & \\
\hline
\hline
\textbf{Transitive} & 1024 & 0.5 & 0.003 & 16 & 200 & 0.98 & Adam & 1e-4 & 50 & 20 \\
\textbf{Inference} & & & & & & & & & & \\
\bottomrule
\end{tabular}
\end{sidewaystable}

%% file: main_neurips.bbl
\begin{thebibliography}{31}
\providecommand{\natexlab}[1]{#1}
\providecommand{\url}[1]{\texttt{#1}}
\expandafter\ifx\csname urlstyle\endcsname\relax
  \providecommand{\doi}[1]{doi: #1}\else
  \providecommand{\doi}{doi: \begingroup \urlstyle{rm}\Url}\fi

\bibitem[uni()]{unity}
Unity.
\newblock \url{http://unity3d.com/}.

\bibitem[Bahdanau et~al.(2015)Bahdanau, Cho, and Bengio]{neur_mach_trans}
D.~Bahdanau, K.~Cho, and Y.~Bengio.
\newblock Neural machine translation by jointly learning to align and
  translate.
\newblock In \emph{3rd International Conference on Learning Representations,
  {ICLR} 2015, San Diego, CA, USA, May 7-9, 2015, Conference Track
  Proceedings}, 2015.
\newblock URL \url{http://arxiv.org/abs/1409.0473}.

\bibitem[Banino et~al.(2018)Banino, Barry, Uria, Blundell, Lillicrap, Mirowski,
  Pritzel, Chadwick, Degris, Modayil, et~al.]{banino2018vector}
A.~Banino, C.~Barry, B.~Uria, C.~Blundell, T.~Lillicrap, P.~Mirowski,
  A.~Pritzel, M.~J. Chadwick, T.~Degris, J.~Modayil, et~al.
\newblock Vector-based navigation using grid-like representations in artificial
  agents.
\newblock \emph{Nature}, 557\penalty0 (7705):\penalty0 429, 2018.

\bibitem[Beattie et~al.(2016)Beattie, Leibo, Teplyashin, Ward, Wainwright,
  K{\"{u}}ttler, Lefrancq, Green, Vald{\'{e}}s, Sadik, Schrittwieser, Anderson,
  York, Cant, Cain, Bolton, Gaffney, King, Hassabis, Legg, and Petersen]{dmlab}
C.~Beattie, J.~Z. Leibo, D.~Teplyashin, T.~Ward, M.~Wainwright,
  H.~K{\"{u}}ttler, A.~Lefrancq, S.~Green, V.~Vald{\'{e}}s, A.~Sadik,
  J.~Schrittwieser, K.~Anderson, S.~York, M.~Cant, A.~Cain, A.~Bolton,
  S.~Gaffney, H.~King, D.~Hassabis, S.~Legg, and S.~Petersen.
\newblock Deepmind lab.
\newblock \emph{CoRR}, abs/1612.03801, 2016.

\bibitem[Blundell et~al.(2016)Blundell, Uria, Pritzel, Li, Ruderman, Leibo,
  Rae, Wierstra, and Hassabis]{mfec}
C.~Blundell, B.~Uria, A.~Pritzel, Y.~Li, A.~Ruderman, J.~Z. Leibo, J.~Rae,
  D.~Wierstra, and D.~Hassabis.
\newblock Model-free episodic control.
\newblock \emph{arXiv preprint arXiv:1606.04460}, 2016.

\bibitem[Cobbe et~al.(2018)Cobbe, Klimov, Hesse, Kim, and
  Schulman]{cobbe2018quantifying}
K.~Cobbe, O.~Klimov, C.~Hesse, T.~Kim, and J.~Schulman.
\newblock Quantifying generalization in reinforcement learning.
\newblock \emph{arXiv preprint arXiv:1812.02341}, 2018.

\bibitem[Espeholt et~al.(2018)Espeholt, Soyer, Munos, Simonyan, Mnih, Ward,
  Doron, Firoiu, Harley, Dunning, Legg, and Kavukcuoglu]{dmlab_impala}
L.~Espeholt, H.~Soyer, R.~Munos, K.~Simonyan, V.~Mnih, T.~Ward, Y.~Doron,
  V.~Firoiu, T.~Harley, I.~Dunning, S.~Legg, and K.~Kavukcuoglu.
\newblock {IMPALA:} scalable distributed deep-rl with importance weighted
  actor-learner architectures.
\newblock \emph{CoRR}, abs/1802.01561, 2018.
\newblock URL \url{http://arxiv.org/abs/1802.01561}.

\bibitem[Graves et~al.(2016)Graves, Wayne, Reynolds, Harley, Danihelka,
  Grabska{-}Barwinska, Colmenarejo, Grefenstette, Ramalho, Agapiou, Badia,
  Hermann, Zwols, Ostrovski, Cain, King, Summerfield, Blunsom, Kavukcuoglu, and
  Hassabis]{dnc}
A.~Graves, G.~Wayne, M.~Reynolds, T.~Harley, I.~Danihelka,
  A.~Grabska{-}Barwinska, S.~G. Colmenarejo, E.~Grefenstette, T.~Ramalho,
  J.~Agapiou, A.~P. Badia, K.~M. Hermann, Y.~Zwols, G.~Ostrovski, A.~Cain,
  H.~King, C.~Summerfield, P.~Blunsom, K.~Kavukcuoglu, and D.~Hassabis.
\newblock Hybrid computing using a neural network with dynamic external memory.
\newblock \emph{Nature}, 538\penalty0 (7626):\penalty0 471--476, 2016.
\newblock \doi{10.1038/nature20101}.
\newblock URL \url{https://doi.org/10.1038/nature20101}.

\bibitem[Guo et~al.(2018)Guo, Azar, Piot, Pires, Pohlen, and
  Munos]{belief_repres}
Z.~D. Guo, M.~G. Azar, B.~Piot, B.~A. Pires, T.~Pohlen, and R.~Munos.
\newblock Neural predictive belief representations.
\newblock \emph{CoRR}, abs/1811.06407, 2018.
\newblock URL \url{http://arxiv.org/abs/1811.06407}.

\bibitem[Hansen et~al.(2018)Hansen, Pritzel, Sprechmann, Barreto, and
  Blundell]{eva}
S.~Hansen, A.~Pritzel, P.~Sprechmann, A.~Barreto, and C.~Blundell.
\newblock Fast deep reinforcement learning using online adjustments from the
  past.
\newblock In \emph{Advances in Neural Information Processing Systems}, pages
  10567--10577, 2018.

\bibitem[He et~al.(2015)He, Zhang, Ren, and Sun]{resnet}
K.~He, X.~Zhang, S.~Ren, and J.~Sun.
\newblock Deep residual learning for image recognition.
\newblock \emph{CoRR}, abs/1512.03385, 2015.
\newblock URL \url{http://arxiv.org/abs/1512.03385}.

\bibitem[Hochreiter and Schmidhuber(1997)]{lstm}
S.~Hochreiter and J.~Schmidhuber.
\newblock Long short-term memory.
\newblock \emph{Neural Comput.}, 9\penalty0 (8):\penalty0 1735--1780, Nov.
  1997.
\newblock ISSN 0899-7667.
\newblock \doi{10.1162/neco.1997.9.8.1735}.
\newblock URL \url{http://dx.doi.org/10.1162/neco.1997.9.8.1735}.

\bibitem[Jaderberg et~al.(2016)Jaderberg, Mnih, Czarnecki, Schaul, Leibo,
  Silver, and Kavukcuoglu]{unreal}
M.~Jaderberg, V.~Mnih, W.~M. Czarnecki, T.~Schaul, J.~Z. Leibo, D.~Silver, and
  K.~Kavukcuoglu.
\newblock Reinforcement learning with unsupervised auxiliary tasks.
\newblock \emph{arXiv preprint arXiv:1611.05397}, 2016.

\bibitem[Ke et~al.(2018)Ke, Goyal, Bilaniuk, Binas, Mozer, Pal, and
  Bengio]{sab}
N.~R. Ke, A.~Goyal, O.~Bilaniuk, J.~Binas, M.~C. Mozer, C.~Pal, and Y.~Bengio.
\newblock Sparse attentive backtracking: Temporal creditassignment through
  reminding.
\newblock \emph{CoRR}, abs/1809.03702, 2018.
\newblock URL \url{http://arxiv.org/abs/1809.03702}.

\bibitem[Leibo et~al.(2018)Leibo, de~Masson~d'Autume, Zoran, Amos, Beattie,
  Anderson, Casta{\~{n}}eda, Sanchez, Green, Gruslys, Legg, Hassabis, and
  Botvinick]{psychlab}
J.~Z. Leibo, C.~de~Masson~d'Autume, D.~Zoran, D.~Amos, C.~Beattie, K.~Anderson,
  A.~G. Casta{\~{n}}eda, M.~Sanchez, S.~Green, A.~Gruslys, S.~Legg,
  D.~Hassabis, and M.~Botvinick.
\newblock Psychlab: {A} psychology laboratory for deep reinforcement learning
  agents.
\newblock \emph{CoRR}, abs/1801.08116, 2018.

\bibitem[Littman and Sutton(2002)]{littman2002predictive}
M.~L. Littman and R.~S. Sutton.
\newblock Predictive representations of state.
\newblock In \emph{Advances in neural information processing systems}, pages
  1555--1561, 2002.

\bibitem[Miyake and Shah(1999)]{miyake_working_mem}
A.~Miyake and P.~Shah.
\newblock \emph{Models of working memory: Mechanisms of active maintenance and
  executive control}.
\newblock Cambridge University Press, 1999.
\newblock \doi{10.1017/CBO9781139174909}.

\bibitem[Mnih et~al.(2016)Mnih, Badia, Mirza, Graves, Lillicrap, Harley,
  Silver, and Kavukcuoglu]{a3c}
V.~Mnih, A.~P. Badia, M.~Mirza, A.~Graves, T.~P. Lillicrap, T.~Harley,
  D.~Silver, and K.~Kavukcuoglu.
\newblock Asynchronous methods for deep reinforcement learning.
\newblock \emph{CoRR}, abs/1602.01783, 2016.

\bibitem[Pineau(2018)]{pineau2018repro}
J.~Pineau.
\newblock Oreproducible, reusable, and robust reinforcement learning (invited
  talk).
\newblock \emph{Advances in Neural Information Processing Systems, 2018}, 2018.

\bibitem[Pritzel et~al.(2017)Pritzel, Uria, Srinivasan, Badia, Vinyals,
  Hassabis, Wierstra, and Blundell]{nec}
A.~Pritzel, B.~Uria, S.~Srinivasan, A.~P. Badia, O.~Vinyals, D.~Hassabis,
  D.~Wierstra, and C.~Blundell.
\newblock Neural episodic control.
\newblock In \emph{Proceedings of the 34th International Conference on Machine
  Learning-Volume 70}, pages 2827--2836. JMLR.org, 2017.

\bibitem[Racani{\`e}re et~al.(2017)Racani{\`e}re, Weber, Reichert, Buesing,
  Guez, Rezende, Badia, Vinyals, Heess, Li, et~al.]{racaniere2017imagination}
S.~Racani{\`e}re, T.~Weber, D.~Reichert, L.~Buesing, A.~Guez, D.~J. Rezende,
  A.~P. Badia, O.~Vinyals, N.~Heess, Y.~Li, et~al.
\newblock Imagination-augmented agents for deep reinforcement learning.
\newblock In \emph{Advances in neural information processing systems}, pages
  5690--5701, 2017.

\bibitem[Ritter et~al.(2018)Ritter, Wang, Kurth-Nelson, Jayakumar, Blundell,
  Pascanu, and Botvinick]{ritter2018been}
S.~Ritter, J.~X. Wang, Z.~Kurth-Nelson, S.~M. Jayakumar, C.~Blundell,
  R.~Pascanu, and M.~Botvinick.
\newblock Been there, done that: Meta-learning with episodic recall.
\newblock \emph{arXiv preprint arXiv:1805.09692}, 2018.

\bibitem[Santoro et~al.(2018)Santoro, Faulkner, Raposo, Rae, Chrzanowski,
  Weber, Wierstra, Vinyals, Pascanu, and Lillicrap]{rmc}
A.~Santoro, R.~Faulkner, D.~Raposo, J.~W. Rae, M.~Chrzanowski, T.~Weber,
  D.~Wierstra, O.~Vinyals, R.~Pascanu, and T.~P. Lillicrap.
\newblock Relational recurrent neural networks.
\newblock \emph{CoRR}, abs/1806.01822, 2018.
\newblock URL \url{http://arxiv.org/abs/1806.01822}.

\bibitem[Smith and Squire(2005)]{Smith10138}
C.~Smith and L.~R. Squire.
\newblock Declarative memory, awareness, and transitive inference.
\newblock \emph{Journal of Neuroscience}, 25\penalty0 (44):\penalty0
  10138--10146, 2005.
\newblock ISSN 0270-6474.
\newblock \doi{10.1523/JNEUROSCI.2731-05.2005}.
\newblock URL \url{http://www.jneurosci.org/content/25/44/10138}.

\bibitem[Sukhbaatar et~al.(2015)Sukhbaatar, Szlam, Weston, and Fergus]{memnets}
S.~Sukhbaatar, A.~Szlam, J.~Weston, and R.~Fergus.
\newblock Weakly supervised memory networks.
\newblock \emph{CoRR}, abs/1503.08895, 2015.
\newblock URL \url{http://arxiv.org/abs/1503.08895}.

\bibitem[Tulving(2002)]{epmem}
E.~Tulving.
\newblock Episodic memory: From mind to brain.
\newblock \emph{Annual Review of Psychology}, 53\penalty0 (1):\penalty0 1--25,
  2002.
\newblock \doi{10.1146/annurev.psych.53.100901.135114}.
\newblock URL \url{https://doi.org/10.1146/annurev.psych.53.100901.135114}.
\newblock PMID: 11752477.

\bibitem[Tulving and Murray(1985)]{tulving1985elements}
E.~Tulving and D.~Murray.
\newblock Elements of episodic memory.
\newblock \emph{Canadian Psychology}, 26\penalty0 (3):\penalty0 235--238, 1985.

\bibitem[van~den Oord et~al.(2018)van~den Oord, Li, and Vinyals]{cpc}
A.~van~den Oord, Y.~Li, and O.~Vinyals.
\newblock Representation learning with contrastive predictive coding.
\newblock \emph{CoRR}, abs/1807.03748, 2018.
\newblock URL \url{http://arxiv.org/abs/1807.03748}.

\bibitem[Vaswani et~al.(2017)Vaswani, Shazeer, Parmar, Uszkoreit, Jones, Gomez,
  Kaiser, and Polosukhin]{transformer}
A.~Vaswani, N.~Shazeer, N.~Parmar, J.~Uszkoreit, L.~Jones, A.~N. Gomez,
  L.~Kaiser, and I.~Polosukhin.
\newblock Attention is all you need.
\newblock \emph{CoRR}, abs/1706.03762, 2017.
\newblock URL \url{http://arxiv.org/abs/1706.03762}.

\bibitem[Wayne et~al.(2018)Wayne, Hung, Amos, Mirza, Ahuja,
  Grabska{-}Barwinska, Rae, Mirowski, Leibo, Santoro, Gemici, Reynolds, Harley,
  Abramson, Mohamed, Rezende, Saxton, Cain, Hillier, Silver, Kavukcuoglu,
  Botvinick, Hassabis, and Lillicrap]{merlin}
G.~Wayne, C.~Hung, D.~Amos, M.~Mirza, A.~Ahuja, A.~Grabska{-}Barwinska, J.~W.
  Rae, P.~Mirowski, J.~Z. Leibo, A.~Santoro, M.~Gemici, M.~Reynolds, T.~Harley,
  J.~Abramson, S.~Mohamed, D.~J. Rezende, D.~Saxton, A.~Cain, C.~Hillier,
  D.~Silver, K.~Kavukcuoglu, M.~Botvinick, D.~Hassabis, and T.~P. Lillicrap.
\newblock Unsupervised predictive memory in a goal-directed agent.
\newblock \emph{CoRR}, abs/1803.10760, 2018.

\bibitem[Zambaldi et~al.(2018)Zambaldi, Raposo, Santoro, Bapst, Li, Babuschkin,
  Tuyls, Reichert, Lillicrap, Lockhart, Shanahan, Langston, Pascanu, Botvinick,
  Vinyals, and Battaglia]{relrl}
V.~F. Zambaldi, D.~Raposo, A.~Santoro, V.~Bapst, Y.~Li, I.~Babuschkin,
  K.~Tuyls, D.~P. Reichert, T.~P. Lillicrap, E.~Lockhart, M.~Shanahan,
  V.~Langston, R.~Pascanu, M.~Botvinick, O.~Vinyals, and P.~Battaglia.
\newblock Relational deep reinforcement learning.
\newblock \emph{CoRR}, abs/1806.01830, 2018.
\newblock URL \url{http://arxiv.org/abs/1806.01830}.

\end{thebibliography}
